arXiv:1412.8697v2 [stat.ML] 15 Oct 2015

# On Semiparametric Exponential Family Graphical Models

Zhuoran Yang* Yang Ning† Han Liu‡

**Abstract**

We propose a new class of semiparametric exponential family graphical models for the analysis of high dimensional mixed data. Different from the existing mixed graphical models, we allow the nodewise conditional distributions to be semiparametric generalized linear models with unspecified base measure functions. Thus, one advantage of our method is that it is unnecessary to specify the type of each node and the method is more convenient to apply in practice. Under the proposed model, we consider both problems of parameter estimation and hypothesis testing in high dimensions. In particular, we propose a symmetric pairwise score test for the presence of a single edge in the graph. Compared to the existing methods for hypothesis tests, our approach takes into account of the symmetry of the parameters, such that the inferential results are invariant with respect to the different parametrizations of the same edge. Thorough numerical simulations and a real data example are provided to back up our results.

## 1 Introduction

Given a $d$-dimensional random vector $\boldsymbol{X} = (X_1, \ldots, X_d)^T$, inferring the conditional independence among $\boldsymbol{X}$ and quantifying its uncertainty are important tasks in statistics. This paper proposes a unified framework for modeling, estimation and uncertainty assessment for a new type of graphical model, named as semiparametric exponential family graphical model. Let $G = (V, E)$ be an undirected graph with node set $V = \{1, 2, \ldots, d\}$ and edge set $E \subset \{(j,k) : 1 \le j < k \le d\}$. The semiparametric exponential family graphical model specifies the joint distribution of $\boldsymbol{X} = (X_1, \ldots, X_d)^T$ such that for each $j \in V$, the conditional distribution of $X_j$ given $\boldsymbol{X}_{\backslash j} := (X_1, \ldots, X_{j-1}, X_{j+1}, \ldots, X_d)^T$ is of the form

$$p(x_j \mid \boldsymbol{x}_{\backslash j}) = \exp\big[\eta_j(\boldsymbol{x}_{\backslash j}) \cdot x_j + f_j(x_j) - b_j(\eta_j, f_j)\big], \tag{1.1}$$

*Department of Operations Research and Financial Engineering, Princeton University, Princeton, NJ 08544. This work is carried out during an internship in the SMiLe (Statistical Machine Learning) lab at Princeton University. e-mail: zy6@princeton.edu.

†Department of Operations Research and Financial Engineering, Princeton University, Princeton, NJ 08544, USA; e-mail: yning@princeton.edu.

‡Department of Operations Research and Financial Engineering, Princeton University, Princeton, NJ 08544, USA; e-mail: hanliu@princeton.edu.

where $\boldsymbol{x}_{\setminus j} = (x_1, \ldots, x_{j-1}, x_{j+1}, \ldots, x_d)$, $\eta_j(\boldsymbol{x}_{\setminus j}) = \alpha_j + \sum_{k \neq j} \beta_{jk} x_k$ is the canonical parameter, $f_j(\cdot)$ is an unknown base measure function and $b_j(\cdot, \cdot)$ is the log-partition function. The unknown parameter contains $\{(\alpha_j, \beta_{jk}, f_j); k \neq j\}$. To make the model identifiable, we set $\alpha_j = 0$ and absorb the term $\alpha_j x_j$ into $f_j(x_j)$. By the Hammersley-Clifford theorem, it is easily seen that $\beta_{jk} \neq 0$ if and only if $X_j$ and $X_k$ are conditionally independent given $\{X_\ell \colon \ell \neq j, k\}$. Therefore, we set $(j, k) \in E$ if and only if $\beta_{jk} \neq 0$. The graph $G$ thus characterizes the conditional independence relationship among the high dimensional distribution of $\boldsymbol{X}$. The key feature of the proposed model is that (1) it is a general semiparametric model and (2) it can be used to handle mixed data, which means that $\boldsymbol{X}$ may contain both continuous and discrete random variables. Compared to the existing mixed graphical models, we allow the nodewise conditional distributions to be semiparametric generalized linear models with unspecified base measure functions. Thus, our method does not need to specify the type of each node and is more convenient to apply in practice. In addition to the proposed new model, our paper has the following two novel contributions.

First, for purpose of estimation of $\beta_{jk}$, we extend the multistage relaxation algorithm (Zhang, 2010) and conduct a localized analysis for a more sophisticated loss function obtained by a statistical chromatography method (Ning and Liu, 2014). The gradient and Hessian matrix of the loss function are nonlinear U-statistics with unbounded kernel functions. This makes our technical analysis more challenging than Zhang (2010). Under the assumption that the sparse eigenvalue condition holds locally, we prove the same optimal statistical rates for parameter estimation as in high-dimensional linear models.

Second, we propose a symmetric pairwise score test for the null hypothesis $H_0 \colon \beta_{jk} = 0$. This is equivalent to testing whether $X_j$ and $X_k$ are independent given $\{X_\ell \colon \ell \neq j, k\}$. Compared to Ning and Liu (2014), the novelty of our method is that we consider a more sophisticated cross type inference which incorporates the symmetry of the parameter, i.e., $\beta_{jk} = \beta_{kj}$. By considering this unique structure of the graphical model, our proposed method achieves the invariance property of the inferential results. That means the same p-values are obtained for testing $\beta_{jk} = 0$ and $\beta_{kj} = 0$. In contrast, the method in Ning and Liu (2014) may lead to different conclusions for testing these two equivalent null hypotheses.

## 1.1 Related Works

There is a huge literature on estimating undirected graphical models (Lauritzen, 1996; Edwards, 2000; Whittaker, 2009). For modeling continuous data, the most commonly used methods are Gaussian graphical models (Yuan and Lin, 2007; Banerjee et al., 2008; Friedman et al., 2008; Ravikumar et al., 2011; Rothman et al., 2008; Lam and Fan, 2009; Shen et al., 2012; Yuan, 2010; Cai et al., 2011; Sun and Zhang, 2013; Guo et al., 2011; Danaher et al., 2014; Mohan et al., 2014; Meinshausen and Bühlmann, 2006; Peng et al., 2009; Friedman et al., 2010). To relax the Gaussian assumption, Liu et al. (2009); Xue et al. (2012b); Liu et al. (2012) propose the Gaussian copula model and Voorman et al. (2014) study the joint additive models for graph estimation. For modeling binary data, the Ising graphical model is considered by Lee et al. (2006); Höfling and Tibshirani (2009); Ravikumar et al. (2010); Xue et al. (2012a); Cheng et al. (2012). In addition to binary data, Allen and Liu (2012) and Yang et al. (2013b) consider the Possion data and Guo et al. (2015) consider the ordinal data. Moreover, Yang et al. (2013a) propose exponential family graphical models, and Tan et al. (2014) propose a general framework for graphical models with hubs.

Recently, modeling the mixed data attracts increasing interests (Lee and Hastie, 2015; Felling-hauer et al., 2013; Cheng et al., 2013; Chen et al., 2015; Fan et al., 2014a; Yang et al., 2014). Compared to Lee and Hastie (2015); Cheng et al. (2013); Chen et al. (2015); Yang et al. (2014), our model has the following two main advantages. First, it is a semiparametric model, which does not need to specify the parametric conditional distribution for each node. Therefore, it provides a more flexible modeling framework than the existing ones. Second, under our proposed model, the estimation and inference methods are easier to implement. Unlike these existing methods, we propose a unified estimation and inference procedure, which does not need to distinguish whether the node satisfies the Gaussian distribution or the Bernoulli distribution. In addition, our estimation and inference methods are more efficient than the nonparametric approach in Fellinghauer et al. (2013). Finally, our method is more convenient for modeling the count data than the latent Gaussian copula approach in Fan et al. (2014a).

Though significant progress has been made towards developing new graph estimation procedures, the research on uncertainty assessment of the estimated graph lags behind. In low dimensions, Drton et al. (2007); Drton and Perlman (2008) establish confidence subgraph of Gaussian graphical models. In high dimensions, Ren et al. (2015); Janková and van de Geer (2015) study the confidence interval for a single edge under Gaussian graphical models and Liu et al. (2013) studies the false discovery rate control. However, all these methods rely on the Gaussian or sub-Gaussian assumption and cannot be easily applied to the discrete data and more generally the mixed data in high dimensions.

## 1.2 Notation

We adopt the following notation throughout this paper. For any vector $\mathbf{v} = (v_1, \ldots, v_d)^T \in \mathbb{R}^d$, define its support as $\text{supp}(\mathbf{v}) = \{t\colon v_t \neq 0\}$. We define its $\ell_0$-norm, $\ell_p$-norm and $\ell_\infty$-norm as $\|\mathbf{v}\|_0 = \big|\text{supp}(\mathbf{v})\big|$, $\|\mathbf{v}\|_p = \big(\sum_{i\in[d]} |v_j|^p\big)^{1/p}$ and $\|\mathbf{v}\|_\infty = \max_{i\in[d]} |v_i|$. Let $\mathbf{v}^{\otimes 2} = \mathbf{v}\mathbf{v}^T$ be the Kronecker product of a vector $\mathbf{v}$ and itself and $\mathbf{v} \circ \mathbf{u} = (v_1u_1, \ldots, v_du_d)^T$ be the Hadamard product of two vectors $\mathbf{u}, \mathbf{v} \in \mathbb{R}^d$. In addition, we use $|\mathbf{v}| = (|v_1|, \ldots, |v_d|)^T$ to denote the elementwise absolute value of vector $\mathbf{v}$ and let $\|\mathbf{v}\|_{\min} = \min_{j\in[d]} |v_i|$. For any matrix $\mathbf{A} = [a_{ij}] \in \mathbb{R}^{d_1\times d_2}$, let $\mathbf{A}_{S_1S_2} = [a_{ij}]_{i\in S_1, j\in S_2}$ be the submatrix of $\mathbf{A}$ with indices in $S_1 \times S_2$; let $\mathbf{A}_{j\setminus j} = [a_{jk}]_{k\neq j}$. Let $\|\mathbf{A}\|_2$, $\|\mathbf{A}\|_1$, $\|\mathbf{A}\|_\infty$, $\|\mathbf{A}\|_{\ell_p}$ be the spectral norm, elementwise supreme norm, elementwise $\ell_1$-norm and operator $\ell_p$-norm respectively. For two matrices $\mathbf{A}_1$ and $\mathbf{A}_2$, we denote $\mathbf{A}_1 \prec \mathbf{A}_2$ if $\mathbf{A}_2 - \mathbf{A}_1$ is positive semidefinite and denote $\mathbf{A}_1 \leq \mathbf{A}_2$ if every entry of $\mathbf{A}_2 - \mathbf{A}_1$ is nonnegative. For a function $f(\boldsymbol{x})\colon \mathbb{R}^d \to \mathbb{R}$, let $\nabla f(\boldsymbol{x})$, $\nabla_S f(\boldsymbol{x})$, $\nabla^2 f(\boldsymbol{x})$ and $\partial f(\boldsymbol{x})$ be the gradient of $f(\boldsymbol{x})$, gradient of $f(\boldsymbol{x})$ with respect to $\boldsymbol{x}_S$, the Hessian of $f(\boldsymbol{x})$ and the subgradient of $f(\boldsymbol{x})$. Let $[d] = \{1, 2, \ldots, d\}$ be the first $d$ positive integers. For a sequence of $d$-dimensional random vectors $\{\boldsymbol{Y}_i\}_{i\geq 1}$, we denote $\boldsymbol{Y}_i \rightsquigarrow \boldsymbol{Y}$ when $\boldsymbol{Y}_i$ converges to a random vector $\boldsymbol{Y}$ in distribution. Finally, for functions $f(n)$ and $g(n)$, we write $f(n) \lesssim g(n)$ to denote that $f(n) \leq cg(n)$ for a universal constant $c \in (0, +\infty)$ and we write $f(n) \asymp g(n)$ when $f(n) \lesssim g(n)$ and $g(n) \lesssim f(n)$ hold simultaneously.

## 1.3 Paper Organization

The rest of this paper is organized as follows. In §2 we introduce the semiparametric exponential family graphical models. In §3 we present our methods for graph estimation and uncertainty assessment. In §4 we lay out the assumptions and main theoretical results. We study the finite-sample performance of our method on both simulated and real-world datasets in §5 and conclude the paper in §6 with some more discussions.

# 2 Semiparametric Exponential Family Graphical Models

The semiparametric exponential family graphical models are defined by specifying the conditional distribution of each variable $X_j$ on the rest of the variables $\{X_k; k \neq j\}$.

**Definition 2.1** (Semiparametric exponential family graphical model)**.** A $d$-dimensional random vector $\boldsymbol{X} = (X_1, \ldots, X_d)^T \in \mathbb{R}^d$ follows a semiparametric exponential graphical model $G = (V, E)$ if for any node $j \in V$, the conditional density of $X_j$ given $\boldsymbol{X}_{\setminus j}$ satisfies

$$p(x_j \,|\, \boldsymbol{x}_{\setminus j}) = \exp\big[x_j(\boldsymbol{\beta}_j^T \boldsymbol{x}_{\setminus j}) + f_j(x_j) - b_j(\boldsymbol{\beta}_j, f_j)\big], \tag{2.1}$$

where $f_j(\cdot)$ is a base measure that does not depend on $\boldsymbol{X}_{\setminus j}$ and $b_j(\cdot,\cdot)$ is the log-partition function. In particular, $(j,k) \in E$ if and only if $\beta_{jk} \neq 0$.

This model is semiparametric since we treat both $\boldsymbol{\beta}_j = (\beta_{j1}, \ldots, \beta_{jj-1}, \beta_{jj+1}, \ldots, \beta_{jd})^T \in \mathbb{R}^{d-1}$ and the univariate function $f_j(\cdot)$ as parameters. Because the model in Definition 2.1 is only specified by the conditional distribution of each variable, it is important to understand the conditions under which a valid joint distribution of $\boldsymbol{X}$ exists. This problem has been addressed by Chen et al. (2015). As shown in Proposition 1 of their paper, one sufficient condition for the existence of joint distribution of $\boldsymbol{X}$ is that, (i) $\beta_{jk} = \beta_{kj}$ for $1 \leq j,k \leq d$ and (ii) $g(\boldsymbol{x}) := \exp\big\{\sum_{j<k}\beta_{jk}x_jx_k + \sum_{j=1}^d f_j(x_j)\big\}$ is integrable.

This paper assumes the above conditions (i) and (ii) hold, so that there exists a joint probability distribution for the model defined in (2.1), whose density has the form of

$$p(\boldsymbol{x}) = \exp\Big[\sum_{k<\ell}\beta_{k\ell}x_kx_\ell + \sum_{j=1}^d f_j(x_j) - A\big(\{\boldsymbol{\beta}_i, f_i\}_{i\in[d]}\big)\Big], \tag{2.2}$$

where $\beta_{k\ell} \neq 0$ if and only if $(k,\ell) \in E$. The log-partition function $A(\cdot)$ is defined as

$$A\big(\{\boldsymbol{\beta}_i, f_i\}_{i\in[d]}\big) := \log\bigg\{\int_{\mathbb{R}^d} \exp\Big[\sum_{k<\ell}\beta_{k\ell}x_kx_\ell + \sum_{j=1}^d f_j(x_j)\Big]\nu(\mathrm{d}\boldsymbol{x})\bigg\}, \tag{2.3}$$

where $\nu(\cdot)$ is the corresponding probability measure. Since $\beta_{k\ell} = \beta_{\ell k}$ for all pairs of nodes $k, \ell$, we will use $\beta_{k\ell}$ and $\beta_{\ell k}$ interchangeably for notational simplicity.

## 2.1 Examples

We provide some widely used parametric examples in the family of semiparametric exponential family graphical models.

**Gaussian Graphical Models:** The Gaussian graphical models assume that $\boldsymbol{X} \in \mathbb{R}^d$ follows a multivariate Gaussian distribution $N(\mathbf{0}, \boldsymbol{\Theta}^{-1})$ where $\boldsymbol{\Theta} \in \mathbb{R}^{d\times d}$ is the precision matrix satisfying $\boldsymbol{\Theta}_{jj} = 1$ for $j \in [d]$. The conditional distribution of $X_j$ given $\boldsymbol{X}_{\setminus j}$ satisfies

$$X_j \mid \boldsymbol{X}_{\setminus j} = \boldsymbol{\alpha}_j^T \boldsymbol{X}_{\setminus j} + \epsilon_j \quad \text{with} \quad \epsilon_j \sim N(0,1),$$

where $\boldsymbol{\alpha}_j = \boldsymbol{\Theta}_{\setminus j, j}$. The conditional density is given by

$$p(x_j \,|\, \boldsymbol{x}_{\setminus j}) = \sqrt{1/(2\pi)}\exp\big[-x_j(\boldsymbol{\Theta}_{\setminus j,j}^T \boldsymbol{x}_{\setminus j}) - 1/2 \cdot x_j^2 - 1/2 \cdot (\boldsymbol{\Theta}_{\setminus j,j}^T \boldsymbol{x}_{\setminus j})^2\big].$$

Compared with (2.1), we obtain $\boldsymbol{\beta}_j = -\boldsymbol{\Theta}_{\backslash j,j}$, $f_j(x) = -x^2/2$ and $b_j(\boldsymbol{\beta}_j, f_j) = (\boldsymbol{\Theta}_{\backslash j,j}^T \boldsymbol{x}_{\backslash j})^2/2 + \log(2\pi)/2$.

**Ising Models:** In an Ising model with no external field, $\boldsymbol{X}$ takes value in $\{0,1\}^d$ and the joint probability mass function $p(\boldsymbol{x}) \propto \exp\{\sum_{j<k} \theta_{jk} x_j x_k\}$. Let $\boldsymbol{\theta}_j = (\theta_{j1}, \ldots, \theta_{j,j-1}, \theta_{j,j+1}, \ldots, \theta_{jd})^T$. The conditional distribution of $X_j$ given $\boldsymbol{X}_{\backslash j}$ is of the form

$$p(x_j \,|\, \boldsymbol{x}_{\backslash j}) = \frac{\exp\{\sum_{k<\ell} \theta_{k\ell} x_k x_\ell\}}{\sum_{x_j \in \{0,1\}} \exp\{\sum_{k<\ell} \theta_{k\ell} x_k x_\ell\}} = \exp\Big\{x_j\big(\boldsymbol{\theta}_j^T \boldsymbol{x}_{\backslash j}\big) - \log\big[1 + \exp(\boldsymbol{\theta}_j^T \boldsymbol{x}_{\backslash j})\big]\Big\}.$$

Therefore in this case we have $\boldsymbol{\beta}_j = \boldsymbol{\theta}_j$, $f_j(x) = 0$ and $b_j(\boldsymbol{\beta}_j, f) = \log[1 + \exp(\boldsymbol{\beta}_j^T \boldsymbol{x}_{\backslash j})]$.

**Exponential Graphical Models:** For exponential graphical models, $\boldsymbol{X}$ takes values in $[0, +\infty)^d$ and the joint probability density satisfies $p(\boldsymbol{x}) \propto \exp\{-\sum_{i=1}^d \phi_i x_i - \sum_{k<\ell} \theta_{k\ell} x_k x_\ell\}$. In order to ensure this probability distribution is normalizable, we require that $\phi_j > 0, \theta_{jk} \geq 0$ for all $j, k \in [d]$. Then we obtain the following conditional probability density of $X_j$ given $\boldsymbol{X}_{\backslash j}$:

$$\begin{aligned} p(x_j \,|\, \boldsymbol{x}_{\backslash j}) &= \exp\Big\{-\sum_{i=1}^d \phi_i x_i - \sum_{k<\ell} \theta_{k\ell} x_k x_\ell\Big\} \Big/ \int_{x_j \geq 0} \exp\Big\{-\sum_{i=1}^d \phi_i x_i - \sum_{k<\ell} \theta_{k\ell} x_k x_\ell\Big\} \mathrm{d}x_j \\ &= \exp\Big\{-x_j\big(\phi_j + \boldsymbol{\theta}_j^T \boldsymbol{x}_{\backslash j}\big) - \log\big(\phi_j + \boldsymbol{\theta}_j^T \boldsymbol{x}_{\backslash j}\big)\Big\}. \end{aligned}$$

Thus we have $\boldsymbol{\beta}_j = -\boldsymbol{\theta}_j$, $f_j(x) = -\phi_j x$ and $b_j(\boldsymbol{\beta}_j, f_j) = \log(\boldsymbol{\beta}_j^T \boldsymbol{x}_{\backslash j} + \phi_j)$.

**Poisson Graphical Models:** In a Poisson graphical model, every node $X_j$ is a discrete random variable taking values in $\mathbb{N} = \{0, 1, 2, \ldots\}$. The joint probability mass function is given by

$$p(\boldsymbol{x}) \propto \exp\Big\{\sum_{j=1}^d \phi_j x_j - \sum_{j=1}^d \log(x_j!) + \sum_{k<\ell} \theta_{k\ell} x_k x_\ell\Big\}.$$

Similar to exponential graphical models, we also need to impose some restrictions on the parameters so that the probability mass function is normalizable. Here we require that $\theta_{jk} \leq 0$ for all $j, k \in [d]$. Then we obtain the following conditional probability mass function of $X_j$ given $\boldsymbol{X}_{\backslash j}$:

$$p(x_j \,|\, \boldsymbol{x}_{\backslash j}) = \exp\Big\{x_j\big(\boldsymbol{\theta}_j^T \boldsymbol{x}_{\backslash j}\big) + \phi_j x_j - \log(x_j!) - b_j(\boldsymbol{\theta}_j, f_j)\Big\}.$$

Thus, $\boldsymbol{\beta}_j = \boldsymbol{\theta}_j$, $f_j(x) = \phi_j x - \log(x!)$ and $b_j(\boldsymbol{\beta}_j, f_j) = \log\{\sum_{y=0}^\infty \exp\big[y(\boldsymbol{\beta}_j^T \boldsymbol{x}_{\backslash j}) + f_j(y)\big]\}$.

# 3 Graph Estimation and Uncertainty Assessment

In this section, we lay out the procedures for graph estimation and uncertainty assessment. Throughout our analysis, we use $\{\boldsymbol{\beta}_i^*, f_i^*\}_{i \in [d]}$ to denote the true parameters, and $\mathbb{E}(\cdot)$ to denote the expectation with respect to the density (2.2) under the true parameters. We first introduce a pseudo-likelihood loss function for $\{\boldsymbol{\beta}_j\}_{j=1}^d$ that is invariant to the nuisance parameters $f_1, \ldots, f_d$ in (2.1). Based on this loss function, we present an Adaptive Multi-stage Convex Relaxation algorithm to estimate each $\boldsymbol{\beta}_j^*$ by minimizing the loss function regularized by a nonconvex penalty function. We then proceed to introduce the respective inferential procedures for accessing the uncertainty of a given edge in the graph.

### 3.1 A Nuisance-Free Loss Function

For graph estimation, we treat $\boldsymbol{\beta}_j$ as the parameter of interest and the base measures $f_j(\cdot)$ as nuisance parameters. Let $\boldsymbol{X}_1, \ldots, \boldsymbol{X}_n$ be $n$ i.i.d. copies of $\boldsymbol{X}$. Due to the presence of $f_j(\cdot)$, finding the conditional maximum likelihood estimator of $\boldsymbol{\beta}_j$ is intractable. To solve this problem, we exploit a pseudo-likelihood loss function proposed in Ning and Liu (2014) that is invariant to the nuisance parameters. This pseudo-likelihood loss is based on pairwise local order statistics. More details are presented as follows.

Let $\boldsymbol{x}_1, \boldsymbol{x}_2, \ldots, \boldsymbol{x}_n$ be $n$ data points that are realizations of $\boldsymbol{X}_1, \boldsymbol{X}_2, \ldots, \boldsymbol{X}_n$. For any $1 \le i < i' \le n$, let $\mathcal{A}^j_{ii'} := \big\{(X_{ij}, X_{i'j}) = (x_{ij}, x_{i'j}), \boldsymbol{X}_{i\setminus j} = \boldsymbol{x}_{i\setminus j}, \boldsymbol{X}_{i'\setminus j} = \boldsymbol{x}_{i'\setminus j}\big\}$ be the event where we observe $\boldsymbol{X}_{i\setminus j}$ and $\boldsymbol{X}_{i'\setminus j}$ and the order statistics of $X_{ij}$ and $X_{i'j}$ (but not the relative ranks of $X_{ij}$ and $X_{i'j}$). In another word, we know the values of a pair of observations $\boldsymbol{x}_i$ and $\boldsymbol{x}_{i'}$ except that we are not aware of $x_{ij} > x_{i'j}$ or $x_{ij} < x_{i'j}$. Ning and Liu (2014) show that

$$\mathbb{P}\big(\boldsymbol{X}_i = \boldsymbol{x}_i, \boldsymbol{X}_{i'} = \boldsymbol{x}_{i'} \big| \mathcal{A}^j_{ii'}\big) = \Big\{1 + \exp\big[-(x_{ij} - x_{i'j})\boldsymbol{\beta}_j{}^T(\boldsymbol{x}_{i\setminus j} - \boldsymbol{x}_{i'\setminus j})\big]\Big\}^{-1},$$

which is free of the nuisance function $f_j(\cdot)$. If we denote $R^j_{ii'}(\boldsymbol{\beta}_j) := \exp[-(x_{ij} - x_{i'j})\boldsymbol{\beta}_j{}^T(\boldsymbol{x}_{i\setminus j} - \boldsymbol{x}_{i'\setminus j})]$, we construct the following pseudo-likelihood loss function for $\boldsymbol{\beta}_j$

$$L_j(\boldsymbol{\beta}_j) := \frac{2}{n(n-1)} \sum_{1 \le i < i' \le n} \log\big[1 + R^j_{ii'}(\boldsymbol{\beta}_j)\big]. \tag{3.1}$$

As shown in (3.1), $L_j(\cdot)$ only involves $\boldsymbol{\beta}_j$ and its form resembles a logistic loss function.

### 3.2 Adaptive Multi-stage Convex Relaxation Algorithm

Now we present the algorithm for parameter estimation. For high dimensional sparse estimation, we minimize the sum of the loss functions $L_j(\boldsymbol{\beta}_j)$ and some penalty function. Two of the most prevalent methods are the LASSO ($\ell_1$-penalization) (Tibshirani, 1996) and the folded concave penalization (Fan et al., 2014b). Although the $\ell_1$-penalization enjoys good computational properties as a convex optimization problem, it is known to incur significant estimation bias for parameters with large absolute values (Zhang and Huang, 2008). In contrast, nonconvex penalties such as smoothly clipped absolute deviation (SCAD) penalty, minimax concave penalty (MCP) and capped-$\ell_1$ penalty can eliminate such bias and attain improved rates of convergence. Therefore, we consider the nonconvex optimization problem

$$\widehat{\boldsymbol{\beta}}_j = \arg\min_{\mathbb{R}^{d-1}} L_j(\boldsymbol{\beta}_j) + \sum_{k \ne j} p_\lambda(|\beta_{jk}|), \tag{3.2}$$

where $\lambda$ is a tuning parameter and $p_\lambda(\cdot) : [0, +\infty) \to [0, +\infty)$ is a penalty function satisfying :

(C.1) The penalty function $p_\lambda(u)$ is continuously nondecreasing and concave with $p_\lambda(0) = 0$.

(C.2) The right-hand derivative at $u = 0$ satisfies $p'_\lambda(0) = p'_\lambda(0+) = \lambda$.

(C.3) There exists $c_1 \in [0, 1]$ and $c_2 \in (0, +\infty)$ such that $p'_\lambda(u+) \ge c_1\lambda$ for $u \in [0, c_2\lambda]$.

Note that we only require the penalty function to be right-differentiable. In what follows, we will denote $p'_\lambda(u)$ as the right-hand derivative. Since $p_\lambda(u)$ is continuously nondecreasing and concave, $p'_\lambda(u)$ is nonincreasing and nonnegative in $[0,\infty)$. It is easy to verify that SCAD, MCP and capped-$\ell_1$ penalty satisfy the conditions (C.1)–(C.3).

Optimization problem (3.2) is nonconvex and may have multiple local solutions. To overcome such difficulty, we exploit the local linear approximation algorithm (Zou and Li, 2008; Fan et al., 2014b) or equivalently, the multi-stage convex relaxation (Zhang, 2010; Zhang et al., 2013) to solve (3.2). Compared with previous works that mainly focus on the problem of sparse linear regression, our loss function $L_j(\boldsymbol{\beta}_j)$ is a $U$-statistic based logistic loss, which requires nontrivial extensions of the existing theoretical analysis.

We present the proposed adaptive multi-stage convex relaxation method in Algorithm 1 . Our algorithm solves a sequence of convex optimization problems corresponding to finer and finer convex relaxations of the original nonconvex optimization problem. More specifically, for each $j=1,\ldots,d$, in the first iteration, step 4 of Algorithm 1 is equivalent to an $\ell_1$-regularized optimization problem and we obtain the first-step solution $\widehat{\boldsymbol{\beta}}_j^{(1)}$. Then in each subsequent iteration, we solve an adaptive $\ell_1$-regularized optimization problem where the weights of the penalty depend on the solution of the previous step. For example, in the $\ell$-th iteration, the regularization parameter $\lambda_{jk}^{(\ell-1)}$ in (3.3) is updated using the $(\ell-1)$-th step estimator $\widehat{\boldsymbol{\beta}}_j^{(\ell-1)}$. Note that $p'_\lambda\big(|\beta_{jk}^{(\ell)}|\big)$ is the right-hand derivative of $p_\lambda(u)$ evaluated at $u=\beta_{jk}^{(\ell)}$.

Since the optimization problem in step 4 is convex, our method is computationally efficient. Moreover, the solution of (3.3) for $\ell=1$ is the solution to the $\ell_1$-regularized problem. As we will show in §4.1, the estimator $\widehat{\boldsymbol{\beta}}_j$ of $\boldsymbol{\beta}_j^*$ obtained by Algorithm 1 attains the optimal rates of convergence for parameter estimation.

**Algorithm 1** Adaptive Multi-stage Convex Relaxation algorithm for parameter estimation

1: Initialize $\lambda_{jk}^{(0)}=\lambda$ for $1\le j,k\le d$.
2: **for** j= 1,2,. . . ,d **do**
3: **for** $\ell=1,2,\ldots,$ until convergence **do**
4: Solve the convex optimization problem

$$\widehat{\boldsymbol{\beta}}_j^{(\ell)}=\arg\min_{\mathbb{R}^{d-1}}\Big\{L_j(\boldsymbol{\beta}_j)+\sum_{k\neq j}\lambda_{jk}^{(\ell-1)}|\beta_{jk}|\Big\}. \tag{3.3}$$

5: Update $\lambda_{jk}^{(\ell)}$ by $\lambda_{jk}^{(\ell)}=p'_\lambda(|\widehat{\beta}_{jk}^{(\ell)}|)$ for $1\le k\le d, k\neq j$.
6: **end for**
7: **Output** $\widehat{\boldsymbol{\beta}}_j=\widehat{\boldsymbol{\beta}}_j^{(\ell)}$, where $\ell$ is the number of iterations until convergence appears.
8: **end for**

### 3.3 Graph Inference: Composite Pairwise Score Test

We consider the hypothesis testing problem $H_0:\beta_{jk}^*=0$ versus $H_1:\beta_{jk}^*\neq 0$ for any given $1\le j<k\le d$. Letting $\boldsymbol{\beta}_{j\setminus k}=(\beta_{j1},\ldots,\beta_{jj-1},\beta_{jj+1},\ldots,\beta_{jk-1},\beta_{jk+1},\ldots,\beta_{jd})^T\in\mathbb{R}^{d-2}$, we denote the parameters associated with node $j$ and node $k$ as $\boldsymbol{\beta}_{j\vee k}:=\big(\beta_{jk};\boldsymbol{\beta}_{j\setminus k}^T,\boldsymbol{\beta}_{k\setminus j}^T\big)^T\in\mathbb{R}^{2d-3}$. Let $\mathbf{H}^j := \mathbb{E}\big[\nabla^2 L_j(\boldsymbol{\beta}_j^*)\big]$ be the expectation of the Hessian of $L_j(\boldsymbol{\beta}_j)$ evaluated at the true parameter $\boldsymbol{\beta}_j^*$. We define two submatrices $\mathbf{H}^j_{jk,j\setminus k}$ and $\mathbf{H}^j_{j\setminus k,j\setminus k}$ of $\mathbf{H}^j$ as

$$\mathbf{H}^j_{jk,j\setminus k} := \left[\mathbb{E}\frac{\partial^2 L_j(\boldsymbol{\beta}_j^*)}{\partial\beta_{jk}\partial\beta_{jv}}\right]_{v\neq k} \in \mathbb{R}^{d-2} \quad \text{and} \quad \mathbf{H}^j_{j\setminus k,j\setminus k} := \left[\mathbb{E}\frac{\partial^2 L_j(\boldsymbol{\beta}_j^*)}{\partial\beta_{ju}\partial\beta_{jv}}\right]_{u,v\neq k} \in \mathbb{R}^{(d-2)\times(d-2)}$$

and define $\mathbf{H}^k_{jk,k\setminus j}$ and $\mathbf{H}^k_{k\setminus j,k\setminus j}$ similarly. Let $\mathbf{w}^*_{j,k}=\mathbf{H}^j_{jk,j\setminus k}\big[\mathbf{H}^j_{j\setminus k,j\setminus k}\big]^{-1}$ and $\mathbf{w}^*_{k,j}=\mathbf{H}^k_{jk,k\setminus j}\big[\mathbf{H}^k_{k\setminus j,k\setminus j}\big]^{-1}$. The composite pairwise score function for parameter $\beta_{jk}$ is defined as

$$S_{jk}(\boldsymbol{\beta}_{j\vee k}) = \nabla_{jk}L_j(\boldsymbol{\beta}_j) + \nabla_{jk}L_k(\boldsymbol{\beta}_k) - \mathbf{w}^{*T}_{j,k}\nabla_{j\setminus k}L_j(\boldsymbol{\beta}_j) - \mathbf{w}^{*T}_{k,j}\nabla_{k\setminus j}L_k(\boldsymbol{\beta}_k). \tag{3.4}$$

where $\nabla_{jk}L_j(\boldsymbol{\beta}_j) = \partial L_j(\boldsymbol{\beta}_j)/\partial\beta_{jk}$ and $\nabla_{j\setminus k}L_j(\boldsymbol{\beta}_j) = \partial L_j(\boldsymbol{\beta}_j)/\partial\boldsymbol{\beta}_{j\setminus k}$. The last two terms are constructed to reduce the effect of nuisance parameters $\boldsymbol{\beta}_{j\setminus k}$ and $\boldsymbol{\beta}_{k\setminus j}$ on accessing the uncertainty of $\beta^*_{jk}$, which is the parameter of interest. A key feature of $S_{jk}(\boldsymbol{\beta}_{j\vee k})$ is that the symmetry of $\beta_{jk}$ and $\beta_{kj}$ (i.e., $\beta_{jk} = \beta_{kj}$) is taken into account, which is distinct from the existing works by Ren et al. (2015); Janková and van de Geer (2015); Liu et al. (2013) for Gaussian graphical models and by Ning and Liu (2014) in the regression setup.

Note that $\mathbf{w}^*_{j,k}$ and $\mathbf{w}^*_{k,j}$ involve unknown quantities, we estimate them using the Dantzig-type estimators. We define the empirical versions of $\mathbf{H}^j_{jk,j\setminus k}$ and $\mathbf{H}^j_{j\setminus k,j\setminus k}$ as

$$\nabla^2_{jk,j\setminus k}L_j(\boldsymbol{\beta}_j) = \left[\frac{\partial^2 L_j(\boldsymbol{\beta}_j)}{\partial\beta_{jk}\partial\beta_{jv}}\right]_{v\neq k} \quad \text{and} \quad \nabla^2_{j\setminus k,j\setminus k}L_j(\boldsymbol{\beta}_j) = \left[\frac{\partial^2 L_j(\boldsymbol{\beta}_j)}{\partial\beta_{ju}\partial\beta_{jv}}\right]_{u,v\neq k}.$$

We also define $\nabla^2_{jk,k\setminus j}L_k(\boldsymbol{\beta}_k)$ and $\nabla^2_{k\setminus j,k\setminus j}L_k(\boldsymbol{\beta}_k)$ similarly. Then we estimate $\mathbf{w}^*_{j,k}$ by solving the optimization problem:

$$\widehat{\mathbf{w}}_{j,k} = \operatorname{argmin} \|\mathbf{w}\|_1 \quad \text{such that} \quad \big\|\nabla^2_{jk,j\setminus k}L_j(0,\widehat{\boldsymbol{\beta}}_{j\setminus k}) - \mathbf{w}^T\nabla^2_{j\setminus k,j\setminus k}L_j(0,\widehat{\boldsymbol{\beta}}_{j\setminus k})\big\|_\infty \leq \lambda_D, \tag{3.5}$$

where $\widehat{\boldsymbol{\beta}}_j$ is the estimator of $\boldsymbol{\beta}^*_j$ obtained from Algorithm 1 and $\lambda_D$ is a tuning parameter. An estimator $\widehat{\mathbf{w}}_{k,j}$ of $\mathbf{w}^*_{k,j}$ can be similarly obtained. With $\widehat{\mathbf{w}}_{j,k}$ and $\widehat{\mathbf{w}}_{k,j}$, we construct a composite pairwise score statistic for $\beta_{jk}$ as

$$\widehat{S}_{jk} = \nabla_{jk}L_j(0,\widehat{\boldsymbol{\beta}}_{j\setminus k}) + \nabla_{jk}L_k\big(0,\widehat{\boldsymbol{\beta}}_{k\setminus j}\big) - \widehat{\mathbf{w}}^T_{j,k}\nabla_{j\setminus k}L_j\big(0,\widehat{\boldsymbol{\beta}}_{j\setminus k}\big) - \widehat{\mathbf{w}}^T_{k,j}\nabla_{k\setminus j}L_k\big(0,\widehat{\boldsymbol{\beta}}_{k\setminus j}\big). \tag{3.6}$$

Comparing (3.4) and (3.6), we see that the pairwise score statistic is computed by replacing $\boldsymbol{\beta}_j$ and $\boldsymbol{\beta}_k$ with $(0,\widehat{\boldsymbol{\beta}}_{j\setminus k})$ and $(0,\widehat{\boldsymbol{\beta}}_{k\setminus j})$ respectively and replacing $\mathbf{w}^*_{j,k}$ and $\mathbf{w}^*_{k,j}$ with $\widehat{\mathbf{w}}_{j,k}$ and $\widehat{\mathbf{w}}_{k,j}$.

To obtain a valid test, we need to derive the limiting distribution of $\widehat{S}_{jk}$ under the null hypothesis. Note that $\widehat{S}_{jk}$ is a linear combination of entries of $\nabla L_j(\boldsymbol{\beta}_j)$ and $\nabla L_k(\boldsymbol{\beta}_k)$, both of which are $U$-statistics. In the next section, we prove the asymptotic normality of $\widehat{S}_{jk}$. More specifically, under the null hypothesis, $\sqrt{n}\widehat{S}_{jk}/2 \rightsquigarrow N(0,\sigma^2_{jk})$ where the limiting variance can be estimated consistently by $\widehat{\sigma}^2_{jk}$ (More details will be explained in the next section). With significance level $\alpha \in (0,1)$, our proposed pairwise score test function $\psi_{jk}(\alpha)$ is defined by

$$\psi_{jk}(\alpha) = \begin{cases} 1 & \text{if} \quad \big|\sqrt{n}\widehat{S}_{jk}/(2\widehat{\sigma}_{jk})\big| > \Phi^{-1}(1-\alpha/2) \\ 0 & \text{if} \quad \big|\sqrt{n}\widehat{S}_{jk}/(2\widehat{\sigma}_{jk})\big| \leq \Phi^{-1}(1-\alpha/2) \end{cases}, \tag{3.7}$$

where $\Phi(t)$ is the distribution function of a standard normal random variable.

The composite pairwise score test for the null hypothesis $H_0\colon \beta^*_{jk} = 0$ is summarized as follows. (i) Calculate $\widehat{\boldsymbol{\beta}}_j$ and $\widehat{\boldsymbol{\beta}}_k$ from Algorithm 1 ; (ii) Obtain $\widehat{\mathbf{w}}_{j,k}$ and $\widehat{\mathbf{w}}_{k,j}$ by solving two Dantzig-type problems (3.5); (iii) Compute the limiting variance $\widehat{\sigma}^2_{jk}$; (iv) Obtain the outcome of pairwise score test by (3.7).

# 4 Theoretical Properties

We have two main results. We first prove that the proposed procedure attains the optimal rate of convergence for parameter estimation. Then, we provide theory for the composite pairwise score test.

## 4.1 Theoretical Results for Parameter Estimation

We first establish theoretical results on the rates of convergence of the adaptive multi-stage convex relaxation estimator. We begin by listing several required assumptions. The first is about moment conditions of $\{X_j\}$ and local smoothness of the log-partition function $A(\cdot)$ defined in (2.3). This assumption also appears in Yang et al. (2013a) and Chen et al. (2015) as a pivotal technical condition for theoretical analysis.

**Assumption 4.1.** For all $j \in [d]$, we assume that the first two moments of $X_j$ are bounded. That is, there exist two constants $\kappa_m$ and $\kappa_v$ such that $\big|\mathbb{E}(X_j)\big| \le \kappa_m$ and $\mathbb{E}(X_j^2) \le \kappa_v$. Denote the true parameters as $\big\{\boldsymbol{\beta}^*_j, f^*_j\big\}_{j\in[d]}$ and define $d$ univariate functions $\bar{A}_j(\cdot)\colon \mathbb{R} \to \mathbb{R}$ as

$$\bar{A}_j(u) := \log\bigg\{\int_{\mathbb{R}^d} \exp\Big[u x_j + \sum_{k<\ell} \beta^*_{k\ell} x_k x_\ell + \sum_{i=1}^d f^*_i(x_i)\Big] \mathrm{d}\nu(\boldsymbol{x})\bigg\}, \quad j \in [d].$$

We assume that there exists a constant $\kappa_h$ such that $\max_{u\colon |u|\le 1} \bar{A}''_j(u) \le \kappa_h$ for all $j \in [d]$.

Unlike the Ising graphical models, $\{X_j\}_{j\in[d]}$ are not bounded in general for semiparametric exponential family graphical models. Instead, we impose mild conditions as in Assumption 4.1 to obtain a loose control of the tail behaviors of the distribution of $\boldsymbol{X}$. As shown in Yang et al. (2013a), Assumption 4.1 implies that for all $j \in [d]$, $\max\big\{\log\mathbb{E}\big[\exp(X_j)\big], \log\mathbb{E}\big[\exp(-X_j)\big]\big\} \le \kappa_m + \kappa_h/2$. Markov inequality implies that, for any $x > 0$, we have

$$\mathbb{P}\big(|X_j| \ge x\big) \le 2\exp(\kappa_m + \kappa_h/2)\exp(-x). \tag{4.1}$$

By setting $x = C\log d$, for some sufficiently large constant $C$, (4.1) implies $\|\boldsymbol{X}\|_\infty \le C\log d$ with high probability.

In addition to Assumption 4.1, we also impose conditions to control the curvature of function $L_j(\cdot)$. In high-dimensional settings, although the loss function is generally not strongly convex, it is sometimes strongly convex in some directions. To characterize this phenomenon, we define the sparse eigenvalue condition as follows.

**Definition 4.2** (Sparse eigenvalue condition)**.** For any $j, s \in [d]$, we define the $s$-sparse eigenvalues of $\mathbb{E}\big[\nabla^2 L_j(\boldsymbol{\beta}_j^*)\big]$ as

$$\rho_{j+}^*(s) := \sup_{\mathbf{v}\in\mathbb{R}^{d-1}} \big\{\mathbf{v}^T\mathbb{E}\big[\nabla^2 L_j(\boldsymbol{\beta}_j^*)\big]\mathbf{v} \colon \|\mathbf{v}\|_0 \le s, \|\mathbf{v}\|_2 = 1\big\};$$
$$\rho_{j-}^*(s) := \inf_{\mathbf{v}\in\mathbb{R}^{d-1}} \big\{\mathbf{v}^T\mathbb{E}\big[\nabla^2 L_j(\boldsymbol{\beta}_j^*)\big]\mathbf{v} \colon \|\mathbf{v}\|_0 \le s, \|\mathbf{v}\|_2 = 1\big\}.$$

**Assumption 4.3.** Let $s^* = \max_{j\in[d]}\big\|\boldsymbol{\beta}_j^*\big\|_0$. We assume that for any $j \in [d]$, there exist an integer $k^* \ge 2s^*$ and a positive number $\rho_*$ satisfying $\lim_{n\to\infty} k^*(\log^9 d/n)^{1/2} = 0$ such that the sparse eigenvalues of $\mathbb{E}\big[\nabla^2 L_j(\boldsymbol{\beta}_j^*)\big]$ satisfy

$$0 < \rho_* \le \rho_{j-}^*(2s^*{+}2k^*) < \rho_{j+}^*(k^*) < +\infty \quad \text{and}$$
$$\rho_{j+}^*(k^*)/\rho_{j-}^*(2s^*{+}2k^*) \le 1 + 0.2k^*/s^* \quad \text{for any } j \in [d].$$

The condition $\rho_{j+}^*(k^*)/\rho_{j-}^*(2s^*{+}2k^*) \le 1{+}0.2k^*/s^*$ requires the eigenvalue ratio $\rho_{j+}^*(k)/\rho_{j-}^*(2k{+}2s^*)$ to grow sub-linearly in $k$. Assumption 4.3 is commonly referred to as sparse eigenvalue condition, which is standard for sparse estimation problems and has been studied by Bickel et al. (2009); Raskutti et al. (2010); Zhang (2010); Negahban et al. (2012); Xiao and Zhang (2013); Loh and Wainwright (2015) and Wang et al. (2014). Our assumption is similar to that in Zhang (2010) and is weaker than the restricted isometry property (RIP) proposed by Candés and Tao (2005). We claim that this assumption is true in general and will be verified for Gaussian graphical models in the supplementary material.

Now we are ready to present the main theorem of this section. Recall that the penalty function $p_\lambda(u)$ satisfies conditions (C.1), (C.2) and (C.3) in §3.2. We use $p'_\lambda(u)$ to denote its right-hand derivative. For convenience, we will set $p'_\lambda(u){=}1$ when $u{<}0$.

**Theorem 4.4** ($\ell_2$- and $\ell_1$-rates of convergence)**.** For all $j \in [d]$, we define the support of $\boldsymbol{\beta}_j^*$ as $S_j := \big\{(j,k)\colon \beta_{jk}^* {\neq} 0, k {\in} [d]\big\}$ and let $s^* {=} \max_{j\in[d]} \|\boldsymbol{\beta}_j^*\|_0$. Let $\rho_* > 0$ be defined in Assumption 4.3. Under Assumptions 4.1 and 4.3, there exists an absolute constant $K > 0$ such that $\|\nabla L_j(\boldsymbol{\beta}_j^*)\|_\infty \le K\sqrt{\log d/n}, \forall j \in [d]$ with probability at least $1 - (2d)^{-1}$. Moreover, the penalty function $p_\lambda(u)\colon [0,+\infty) \to [0,+\infty)$ in (3.2) satisfies conditions (C.1), (C.2) and (C.3) listed in §3.2 with $c_1 = 0.91$ and $c_2 \ge 24/\rho_*$ for condition (C.3). We set the regulization parameter $\lambda {=} C\sqrt{\log d/n}$ with $C {\ge} 25K$. We denote constants $\varrho = c_2(c_2\rho_* {-} 11)^{-1}$, $A_1 {=} 22\varrho$, $A_2 {=} 2.2c_2$, $B_1 {=} 32\varrho$, $B_2 {=} 3.2c_2$, $\gamma {=} 11c_2^{-1}\rho_*^{-1} {<} 1$ and define $\Upsilon_j {:=} \big[\sum_{(j,k)\in S_j} p'_\lambda(|\beta_{jk}^*| {-} c_2\lambda)^2\big]^{1/2}$. Then with probability at least $1 {-} d^{-1}$, we have the following statistical rates of convergence:

$$\big\|\widehat{\boldsymbol{\beta}}_j^{(\ell)} - \boldsymbol{\beta}_j^*\big\|_2 \le A_1\Big[\big\|\nabla_{S_j} L_j(\boldsymbol{\beta}_j^*)\big\|_2 + \Upsilon_j\Big] + A_2\sqrt{s^*}\lambda\gamma^\ell \quad \text{and} \tag{4.2}$$
$$\big\|\widehat{\boldsymbol{\beta}}_j^{(\ell)} - \boldsymbol{\beta}_j^*\big\|_1 \le B_1\sqrt{s^*}\Big[\big\|\nabla_{S_j} L_j(\boldsymbol{\beta}_j^*)\big\|_2 + \Upsilon_j\Big] + B_2 s^*\lambda\gamma^\ell, \forall j \in [d]. \tag{4.3}$$

From Theorem 4.4 we see that the statistical rates are dominated by the second term if $p'_\lambda(|\beta_{jk}^*| {-} c_2\lambda)$ is not negligible. If the signal strength is large enough such that $p'_\lambda(\beta - c_2\lambda) = 0$ where $\beta {=} \min_{(j,k)\in S_j} |\beta_{jk}^*|$, after sufficient number of iterations, the statistical rates will be of order

$$\big\|\widehat{\boldsymbol{\beta}}_j^{(\ell)} - \boldsymbol{\beta}_j^*\big\|_2 = \mathcal{O}_{\mathbb{P}}\Big(\big\|\nabla_{S_j} L_j(\boldsymbol{\beta}_j^*)\big\|_2\Big) \quad \text{and} \quad \big\|\widehat{\boldsymbol{\beta}}_j^{(\ell)} - \boldsymbol{\beta}_j^*\big\|_1 = \mathcal{O}_{\mathbb{P}}\Big(\sqrt{s^*}\big\|\nabla_{S_j} L_j(\boldsymbol{\beta}_j^*)\big\|_2\Big).$$

However, if the signals are uniformly small such that $p'_\lambda\big(|\beta^*_{jk}|-c_2\lambda\big)>0$ for all $(j,k)\in S_j$, the rates of convergence will be of order

$$\big\|\widehat{\boldsymbol{\beta}}_j^{(\ell)}-\boldsymbol{\beta}_j^*\big\|_2=\mathcal{O}_{\mathbb{P}}\big(\sqrt{s^*}\lambda\big)\quad\text{and}\quad\big\|\widehat{\boldsymbol{\beta}}_j^{(\ell)}-\boldsymbol{\beta}_j^*\big\|_1=\mathcal{O}_{\mathbb{P}}\big(s^*\lambda\big),$$

which are identical to the $\ell_2$- and $\ell_1$-rates of convergence of LASSO estimator, respectively (Ning and Liu, 2014). Thus $c_2\lambda$ can be viewed as the threshold of signal strength. Therefore, after sufficient numbers of iterations, the final estimator $\widehat{\boldsymbol{\beta}}_j$ obtained from the multi-stage convex relaxation algorithms attains the following more refined rates of convergence:

$$\big\|\widehat{\boldsymbol{\beta}}_j-\boldsymbol{\beta}_j^*\big\|_2=\mathcal{O}_{\mathbb{P}}\Big(\big\|\nabla_{S_j}L_j(\boldsymbol{\beta}_j^*)\big\|_2+\Upsilon_j\Big)\ \text{ and }\ \big\|\widehat{\boldsymbol{\beta}}_j-\boldsymbol{\beta}_j^*\big\|_1=\mathcal{O}_{\mathbb{P}}\Big(\sqrt{s^*}\big[\big\|\nabla_{S_j}L_j(\boldsymbol{\beta}_j^*)\big\|_2+\Upsilon_j\big]\Big).$$

These statistical rates of convergence are optimal in the sense that they cannot be improved in terms of order.

Finally, we comment that the sparsity level $s^*$ in (4.2) and (4.3) can be replaced by the sparsity level of each $\boldsymbol{\beta}_j^*$. Let $s_j^*=\big\|\boldsymbol{\beta}_j^*\big\|_0$ be the sparsity level of $\boldsymbol{\beta}_j^*$ and $\lambda_j$ be the regularization parameter for optimization problem (3.2) such that $\lambda_j\asymp\big\|\nabla L_j(\boldsymbol{\beta}_j^*)\big\|_\infty$. The statistical rates of convergence for each $\widehat{\boldsymbol{\beta}}_j^{(\ell)}$ can be improved to

$$\big\|\widehat{\boldsymbol{\beta}}_j^{(\ell)}-\boldsymbol{\beta}_j^*\big\|_2=\mathcal{O}_{\mathbb{P}}\Big(\sqrt{s_j^*}\lambda_j\Big)\quad\text{and}\quad\big\|\widehat{\boldsymbol{\beta}}_j^{(\ell)}-\boldsymbol{\beta}_j^*\big\|_1=\mathcal{O}_{\mathbb{P}}\big(s_j^*\lambda_j\big).$$

## 4.2 Theoretical Results for Composite Pairwise Score Test

In the composite pairwise score test for the null hypothesis $H_0:\beta^*_{jk}=0$, to utilize the structure of the graphical model, we construct the test statistic by combining loss functions $L_j(\cdot)$ and $L_k(\cdot)$ together because $\beta_{jk}$ appears in both $L_j(\boldsymbol{\beta}_j)$ and $L_k(\boldsymbol{\beta}_k)$ (recall that we use $\beta_{jk}$ and $\beta_{kj}$ interchangeably). In what follows, we present the theoretical results that guarantee the validity of the proposed method.

Recall that we define the pairwise score function $S_{jk}(\boldsymbol{\beta}_{j\vee k})$ and the pairwise score statistic $\widehat{S}_{jk}$ in (3.4) and (3.6) respectively. According to a pair of nodes $j,k\in[d]$, entries in $\boldsymbol{\beta}_j$ and $\boldsymbol{\beta}_k$ can be categorized into three types: (i) $\beta_{jk}$, (ii) $\boldsymbol{\beta}_{j\backslash k}=(\beta_{j\ell};\ell\neq k)^T$ and (iii) $\boldsymbol{\beta}_{k\backslash j}=(\beta_{k\ell};\ell\neq j)^T$. Recall that we denote $\boldsymbol{\beta}_{j\vee k}:=\big(\beta_{jk},\boldsymbol{\beta}_{j\backslash k}^T,\boldsymbol{\beta}_{k\backslash j}^T\big)^T$ for notational simplicity. If we let $L_{jk}\big(\boldsymbol{\beta}_{j\vee k}\big):=L_j(\boldsymbol{\beta}_j)+L_k(\boldsymbol{\beta}_j)$, then the components of $\nabla L_{jk}\big(\boldsymbol{\beta}_{j\vee k}\big)$ are given by

$$\begin{aligned}&\nabla_{jk}L_{jk}(\boldsymbol{\beta}_{j\vee k})=\nabla_{jk}L_j(\boldsymbol{\beta}_j)+\nabla_{kj}L_k(\boldsymbol{\beta}_k);\quad\nabla_{j\backslash k}L_{jk}(\boldsymbol{\beta}_{j\vee k})=\nabla_{j\backslash k}L_j(\boldsymbol{\beta}_j)\ \text{ and}\\&\nabla_{k\backslash j}L_{jk}(\boldsymbol{\beta}_{j\vee k})=\nabla_{k\backslash j}L_k(\boldsymbol{\beta}_k).\end{aligned}$$

Let $\widehat{\boldsymbol{\beta}}_j$ and $\widehat{\boldsymbol{\beta}}_k$ be the estimators of $\boldsymbol{\beta}_j^*$ and $\boldsymbol{\beta}_k^*$ obtained from Algorithm 1. Assume that $\mathbf{H}^j=\mathbb{E}\big[\nabla^2L_j(\boldsymbol{\beta}_j^*)\big]$ is invertible and denote $\mathbf{w}^*_{j,k}=\mathbf{H}^j_{jk,j\backslash k}\big(\mathbf{H}^j_{j\backslash k,j\backslash k}\big)^{-1}$. Note that we can write the pairwise score function $S_{jk}(\cdot)$ and the test statistic $\widehat{S}_{jk}$ as

$$S_{jk}(\boldsymbol{\beta}_{j\vee k})=\nabla_{jk}L_{jk}\big(\boldsymbol{\beta}_{j\vee k}\big)-\mathbf{w}^{*T}_{j,k}\nabla_{j\backslash k}L_{jk}\big(\boldsymbol{\beta}_{j\vee k}\big)-\mathbf{w}^{*\,T}_{k,j}\nabla_{k\backslash j}L_{jk}\big(\boldsymbol{\beta}_{j\vee k}\big)\ \text{ and}\tag{4.4}$$

$$\widehat{S}_{jk}=\nabla_{jk}L_{jk}\big(\widehat{\boldsymbol{\beta}}'_{j\vee k}\big)-\widehat{\mathbf{w}}^T_{j,k}\nabla_{j\backslash k}L_{jk}\big(\widehat{\boldsymbol{\beta}}'_{j\vee k}\big)-\widehat{\mathbf{w}}^T_{k,j}\nabla_{k\backslash j}L_{jk}\big(\widehat{\boldsymbol{\beta}}'_{j\vee k}\big).\tag{4.5}$$

Here $\widehat{\boldsymbol{\beta}}'_{j\vee k}:=\big(0,\widehat{\boldsymbol{\beta}}_{j\backslash k}^T,\widehat{\boldsymbol{\beta}}_{k\backslash j}^T\big)^T$, $\widehat{\mathbf{w}}_{j,k}$ is obtained from the Dantzig-type problem (3.5), and $\widehat{\mathbf{w}}_{k,j}$ can be obtained similarly. To derive the asymptotic distribution of $\widehat{S}_{jk}$, we first show that $\sqrt{n}\big[\widehat{S}_{jk}-$ $S_{jk}(\boldsymbol{\beta}^*_{j\vee k})\big] = o_{\mathbb{P}}(1)$. Then the problem is reduced to finding the limiting distribution of $U$-statistic $S_{jk}(\boldsymbol{\beta}^*_{j\vee k})$. Thanks to the $U$-statistic structure, we can characterize the limiting distribution of $S_{jk}(\boldsymbol{\beta}^*_{j\vee k})$ using the method of Hájek projection (Van der Vaart, 2000), which approximates a $U$-statistic with a sum of i.i.d. random variables.

To begin with, we denote the kernel functions of $\nabla L_j(\boldsymbol{\beta}_j)$, $\nabla L_k(\boldsymbol{\beta}_k)$ and $\nabla L_{jk}\big(\boldsymbol{\beta}_{j\vee k}\big)$ as $\mathbf{h}^j_{ii'}(\boldsymbol{\beta}_j)$, $\mathbf{h}^k_{ii'}(\boldsymbol{\beta}_k)$ and $\mathbf{h}^{jk}_{ii'}\big(\boldsymbol{\beta}_{j\vee k}\big)$ respectively. It can be shown that $\mathbb{E}\big[\mathbf{h}^j_{ii'}(\boldsymbol{\beta}^*_j)\big] = \mathbb{E}\big[\mathbf{h}^k_{ii'}(\boldsymbol{\beta}^*_k)\big] = 0$; hence $\mathbf{h}^{jk}_{ii'}\big(\boldsymbol{\beta}^*_{j\vee k}\big)$ is also centered. We define

$$\mathbf{g}_{jk}(\boldsymbol{X}_i) := n/2 \cdot \mathbb{E}\big[\nabla L_{jk}\big(\boldsymbol{\beta}^*_{j\vee k}\big)\big|\boldsymbol{X}_i\big] = \mathbb{E}\big[\mathbf{h}^{jk}_{ii'}\big(\boldsymbol{\beta}^*_{j\vee k}\big)\big|\boldsymbol{X}_i\big] \quad \text{and} \tag{4.6}$$

$$\mathbf{U}_{jk} := 2/n \cdot \sum_{i=1}^{n} \mathbf{g}_{jk}(\boldsymbol{X}_i) = \sum_{i=1}^{n} \mathbb{E}\big[\nabla L_{jk}\big(\boldsymbol{\beta}^*_{j\vee k}\big)\big|\boldsymbol{X}_i\big]. \tag{4.7}$$

Thus $2/n \cdot \mathbf{g}_{jk}\big(\boldsymbol{X}_i\big)$ is the projection of $\nabla L_{jk}\big(\boldsymbol{\beta}^*_{j\vee k}\big)$ onto the $\sigma$-filed generated by $\boldsymbol{X}_i$ and $\mathbf{U}_{jk}$ is the Hájek projection of $\nabla L_{jk}\big(\boldsymbol{\beta}^*_{j\vee k}\big)$. Under mild conditions, $\mathbf{U}_{jk}$ is a good approximation of $\nabla L_{jk}\big(\boldsymbol{\beta}^*_{j\vee k}\big)$, which enables us to characterize the limiting distribution of $S_{jk}(\boldsymbol{\beta}^*_{j\vee k})$. We present the following assumption that guarantees that $\mathbf{g}_{jk}\big(\boldsymbol{X}_i\big)$ is not degenerate.

**Assumption 4.5.** Under Assumption 4.1, for $\mathbf{g}_{jk}(\boldsymbol{X}_i)$ defined in (4.6), we denote the covariance matrix of $\mathbf{g}_{jk}(\boldsymbol{X}_i)$ as $\boldsymbol{\Sigma}^{jk} := \mathbb{E}\big[\mathbf{g}_{jk}(\boldsymbol{X}_i)\mathbf{g}_{jk}(\boldsymbol{X}_i)^T\big]$. We assume that there exist a constant $c_\Sigma > 0$ such that $\lambda_{\min}(\boldsymbol{\Sigma}^{jk}) \geq c_\Sigma$ for all $1 \leq j < k \leq d$.

Assumption 4.5 requires the minimum eigenvalue of $\boldsymbol{\Sigma}^{jk}$ to be bounded away from 0; which implies that for all $\mathbf{v} \in \mathbb{R}^{2d-3}$, $\|\mathbf{v}\|_2 = 1$ $\mathrm{Var}(\mathbf{v}^T\mathbf{U}_{jk}) \geq 4c_\Sigma$. Thus this assumption guarantees the asymptotic variance of $\sqrt{n}S_{jk}(\boldsymbol{\beta}^*_{j\vee k})$ is bounded away from 0. We also present the following assumption that specifies the scaling of the Dantzig selector problem (3.5).

**Assumption 4.6.** We assume that for all $j \in [d]$, $\mathbf{H}^j$ is invertible. In addition, we assume that there exist an integer $s^\star_0$ and a positive number $w_0$ such that $\|\mathbf{w}^*_{j,k}\|_0 \leq s^\star_0 - 1$ and $\|\mathbf{w}^*_{j,k}\|_1 \leq w_0$. The regularization parameter $\lambda_D$ in the Dantzig selector problem (3.5) satisfies $\lambda_D \asymp \max\{1, w_0\}s^*\lambda \log^2 d$. Moreover, we assume that

$$\lim_{n\to\infty} (1+w_0+w_0^2)s^*\lambda\log^2 d = 0, \quad \lim_{n\to\infty} (1+w_0)s^\star_0\lambda_D = 0 \quad \text{and} \quad \lim_{n\to\infty} \sqrt{n}(s^*+s^\star_0)\lambda\lambda_D = 0. \tag{4.8}$$

In addition, similar to Assumption 4.3, we denote the $s$-sparse eigenvalues of $\mathbb{E}\big[\nabla^2 L_j(\boldsymbol{\beta}^*_j)\big]$ as $\rho^*_{j-}(s)$ and $\rho^*_{j+}(s)$ respectively. We further assume that there exist an integer $k^\star_0 \geq s^\star_0$ and a positive number $\nu_*$ such that $\lim_{n\to\infty} k^\star_0\big(\log^9 d/n\big)^{1/2} = 0$ and

$$0 < \nu_* \leq \rho^*_{j-}(s^\star_0 + k^\star_0) < \rho^*_{j+}(k^\star_0) \leq \big(1 + 0.5k^\star_0/s^\star_0\big)\nu_*, \quad 1 \leq j \leq d. \tag{4.9}$$

If we can treat $w_0$ as a constant, and $k^*$ and $k^\star_0$ is of the same order of $s^*$ and $s^\star_0$ respectively, Assumption 4.6 is reduced to $\lambda_D \asymp s^*\lambda\log^2 d$, $s^\star_0\lambda_D = o(1)$, $s^*\lambda\log^2 d = o(1)$ and $(s^*+s^\star)\lambda\lambda_D = o(n^{1/2})$. Since $\lambda \asymp \sqrt{\log d/n}$, we can choose $\lambda_D = Cs^*(\log^5 d/n)^{1/2}$ with a sufficiently large $C$, provided $(s^*+s^\star_0)(\log^9 d/n)^{1/2} = o(1)$, $s^\star_0 s^*(\log^5 d/n)^{1/2} = o(1)$ and $(s^*+s^\star_0)s^*\log^3 d/n = o(n^{-1/2})$. Hence the condition is fulfilled if

$$\log d = o\Big(\min\big\{(\sqrt{n}/s^*)^{2/9}, (\sqrt{n}/s^\star_0)^{2/9}, (\sqrt{n}/s^{*2})^{1/3}, (\sqrt{n}/s^*s^\star)^{1/3}\big\}\Big).$$

Now we are ready to present the main theorem of composite pairwise score test.

**Theorem 4.7.** Under the Assumptions 4.1, 4.3, 4.5 and 4.6, for $j,k\in[d]$ and $j\neq k$, we denote $\boldsymbol{\beta}_{j\setminus k}=(\beta_{j\ell},\ell\neq k)^T$, $\boldsymbol{\beta}_{k\setminus j}=(\beta_{k\ell},\ell\neq j)^T$ and $\boldsymbol{\beta}_{j\vee k}=(\beta_{jk},\boldsymbol{\beta}_{j\setminus k}^T,\boldsymbol{\beta}_{k\setminus j}^T)^T$. The pairwise score statistic and pairwise score function $\widehat{S}_{jk}$ and $S_{jk}(\boldsymbol{\beta}_{j\vee k})$ are defined in (3.6) and (3.4) respectively. Then it holds uniformly for all $j\neq k$ and $j,k\in[d]$ that $\sqrt{n}\widehat{S}_{jk}=\sqrt{n}S_{jk}(\boldsymbol{\beta}^*_{j\vee k})+o_{\mathbb{P}}(1)$. Furthermore, we let $\widehat{\boldsymbol{\beta}}'_{j\vee k}=(0,\widehat{\boldsymbol{\beta}}_{j\setminus k}^T,\widehat{\boldsymbol{\beta}}_{k\setminus j}^T)^T$ and define $\widehat{\boldsymbol{\Sigma}}^{jk}:=n^{-1}\sum_{i=1}^n\{(n-1)^{-1}\sum_{i'\neq i}\mathbf{h}_{ii'}^{jk}(\widehat{\boldsymbol{\beta}}'_{j\vee k})\}^{\otimes 2}$, where $\mathbf{h}_{ii'}^{jk}(\boldsymbol{\beta}_{j\vee k})$ is the kernel function of the second-order $U$-statistic $\nabla L_{jk}(\boldsymbol{\beta}^*_{j\vee k})$. And we define $\widehat{\sigma}_{jk}$ by $\widehat{\sigma}^2_{jk}:=\widehat{\boldsymbol{\Sigma}}^{jk}_{jk,jk}-2\widehat{\boldsymbol{\Sigma}}^{jk}_{jk,j\setminus k}\widehat{\mathbf{w}}_{j,k}-2\widehat{\boldsymbol{\Sigma}}^{jk}_{jk,k\setminus j}\widehat{\mathbf{w}}_{k,j}+\widehat{\mathbf{w}}_{j,k}^T\widehat{\boldsymbol{\Sigma}}^{jk}_{j\setminus k,j\setminus k}\widehat{\mathbf{w}}_{j,k}+\widehat{\mathbf{w}}_{k,j}^T\widehat{\boldsymbol{\Sigma}}^{jk}_{k\setminus j,k\setminus j}\widehat{\mathbf{w}}_{k,j}$. Under the null hypothesis $H_0\colon\beta^*_{jk}=0$, we have $\sqrt{n}\widehat{S}_{jk}/(2\widehat{\sigma}_{jk})\rightsquigarrow N(0,1)$.

By Theorem 4.7, to test the null hypothesis $H_0\colon\beta^*_{jk}=0$ against the alternative hypothesis $H_1\colon\beta^*_{jk}\neq 0$, we reject $H_0$ if the studentized test statistic $\sqrt{n}\widehat{S}_{jk}/(2\widehat{\sigma}_{jk})$ is too extreme. Recall that the test function of the composite pairwise score test with significance level $\alpha$ is given by $\psi_{jk}(\alpha)$ in (3.7). The associated p-value is defined as $p_\psi^{jk}:=2\big[1-\Phi\big(\big|\sqrt{n}\widehat{S}_{jk}/(2\widehat{\sigma}_{jk})\big|\big)\big]$. By Theorem 4.7, under $H_0$, we have

$$\lim_{n\to\infty}\mathbb{P}\big(\psi_{jk}(\alpha)=1\mid H_0\big)=\alpha\quad\text{and}\quad p_\psi^{jk}\rightsquigarrow\text{Unif}[0,1]\ \text{ under } H_0,$$

where Unif$[0,1]$ is the uniform distribution on $[0,1]$.

**Remark 4.8.** There are a number of recent works on the uncertainty assessment for high dimensional linear models or generalized linear models with $\ell_1$-penalty; see Lee et al. (2013); Lockhart et al. (2014); Belloni et al. (2012, 2013); Zhang and Zhang (2014); Javanmard and Montanari (2014); van de Geer et al. (2014). These works utilize the convexity and the Karush-Kuhn-Tuker conditions of the LASSO problem. Compared with these works, our pairwise score test is constructed using a nonconvex penalty function and is applicable to a larger model class. Ning and Liu (2014) consider the score test for $\ell_1$-penalized semiparametric generalized linear models in the regression setting. Compared with this work, we adopt a composite score test with a nonconvex penalty and relax many technical assumptions including the bounded covariate assumption. For nonconvex penalizations, Fan and Lv (2011); Bradic et al. (2011) establish the asymptotic normality for low dimensional nonzero parameters in high-dimensional models based on the oracle properties. However, their approach depends on the minimal signal strength assumption. This assumption is not needed in our approach.

# 5 Numerical Results

In this section we study the finite-sample performance of the proposed graph inference methods on both simulated and real-world datasets. More specifically, for numerical simulations, we examine the validity of the proposed methods on Gaussian, Ising, and mixed graphical models. In addition, we analyze the music annotation dataset as a real data application.

## 5.1 Simulation Studies

We first examine the numerical performance of the graph inference using the proposed pairwise score tests for the null hypothesis $H_0\colon\beta^*_{jk}=0$. We simulate data from the following three settings:

(i) Gaussian graphical model. We set $n = 100$ and $d = 200$. The graph structure is a 4-nearest-neighbor graph, that is, for $j, k \in [d]$, $j \neq k$, node $j$ is connected with node $k$ if $|j-k| = 1, 2, d-2, d-1$. More specifically, we sample $\boldsymbol{X}_1, \ldots, \boldsymbol{X}_n$ from a Gaussian distribution $N_d(\mathbf{0}, \boldsymbol{\Sigma})$. For the precision matrix $\boldsymbol{\Theta} = \boldsymbol{\Sigma}^{-1}$, we set $\boldsymbol{\Theta}_{jj} = 1$, $\big|\boldsymbol{\Theta}_{jk}\big| = \mu \in [0, 0.25)$ for $|j-k| = 1, 2, d-2, d-1$ and $\boldsymbol{\Theta}_{jk} = 0$ for $2 \leq |j-k| \leq d-2$. Note that $\mu$ denotes the signal strength of the graph inference problem and $\mu \leq 0.25$ ensures that $\boldsymbol{\Theta}$ is diagonal dominant and invertible.

(ii) Ising graphical model. We set $n = 100$ and $d = 200$. The graph structure is a $10 \times 20$ grid with the sparsity level $s^* = 4$. We use Markov Chain Monte Carlo method (MCMC) to simulate $n$ data from an Ising model with joint distribution $p(\boldsymbol{x}) \propto \exp\big(\sum_{j \neq k} \beta^*_{jk} x_j x_k\big)$ (using the package `IsingSampler`). We set $|\beta^*_{jk}| = \mu \in [0, 1]$ if there exists an edge connecting node $j$ and node $k$, and $\beta^*_{jk} = 0$ otherwise.

(iii) Mixed graphical model. We set $n = 100$ and $d = 200$. The graph structure is a $10 \times 10 \times 2$ grid with the sparsity level $s^* = 5$. We set the nodes in the first layer to be binomial and nodes in the second layer to be Gaussian. We set $|\beta^*_{jk}| = \mu \in [0, 1]$ if there exists an edge connecting node $j$ and node $k$, and $\beta^*_{jk} = 0$ otherwise. We refer to Lee and Hastie (2015) for details.

We denote the true parameters of the graphical models as $\{\beta^*_{jk}, j \neq k\}$. We also denote $\boldsymbol{\beta}^*_j = (\beta^*_{j1}, \ldots, \beta^*_{jd})^T$. For the Gaussian graphical model, we have $\beta^*_{jk} = \boldsymbol{\Theta}_{jk}$. We first obtain a point estimate of $\boldsymbol{\beta}^*_j$ by solving (3.2) using Algorithm 1 with the capped-$\ell_1$ penalty $p_\lambda(u) = \lambda \min\{u, \lambda\}$. The parameter $\lambda$ is chosen by 10-fold cross validation as suggested by Ning and Liu (2014).

Recall that the form of the loss function $L_j(\boldsymbol{\beta}_j)$ is exactly the loss function for logistic regression, where we use Rademacher random variables $y_{ii'}$ as response and $y_{ii'}(x_{ij} - x_{i'j})\boldsymbol{\beta}_j^T(\boldsymbol{x}_{i\setminus j} - \boldsymbol{x}_{i'\setminus j})$ as covariates, Algorithm 1 can be easily implemented by using the $\ell_1$-regularized logistic regression such as the `glmnet` package. In particular, the algorithm converges quickly after a few iterations, indicating that it attains a good balance between computational efficiency and statistical accuracy.

Once $\widehat{\boldsymbol{\beta}}_j$ is obtained, we solve the Dantzig-type problem (3.5) using $\widehat{\boldsymbol{\beta}}_j$ as input. We set the regularization parameter $\lambda_D$ to be 0.2. We find that the final performance of the proposed method is not quite sensitive to $\lambda_D$.

We compare the method of pairwise score test with the desparsity method proposed in van de Geer et al. (2014). Although the desparsity method is only proposed for hypothesis tests in generalized linear models (GLMs), it can be used for graphical models by performing nodewise regression. Their method constructs the confidence interval for each parameter of the GLMs, which is equivalent to a hypothesis test by rejecting the null hypothesis $H_0\colon \beta^*_{jk} = 0$ if the confidence interval fails to cover 0. To examine the validity of our method, we test the null hypothesis $H_0\colon \beta^*_{jk} = 0$ when the data are generated by $\beta^*_{jk} = \mu$. Here, we let $\mu$ increase from 0 to a sufficiently large number. In Figure1, we report the powers of the hypothesis tests at the 0.05 significance level. To obtain the type I error and power, we repeat the whole procedure for 1000 times and use the empirical rejection rate to evaluate type I error and power. In particular, the value corresponding to $\mu = 0$ is the type I error. As revealed by the figures, our method achieves accurate type I errors, which is comparable to the desparsity method. In terms of the power of the test, our method is as powerful as the desparsity method. In addition, we emphasize that for mixed graphical models the desparsity method needs to know the type of each nodes as a priori. Such phenomenon suggests that we sacrifice little testing efficiency for model generality.

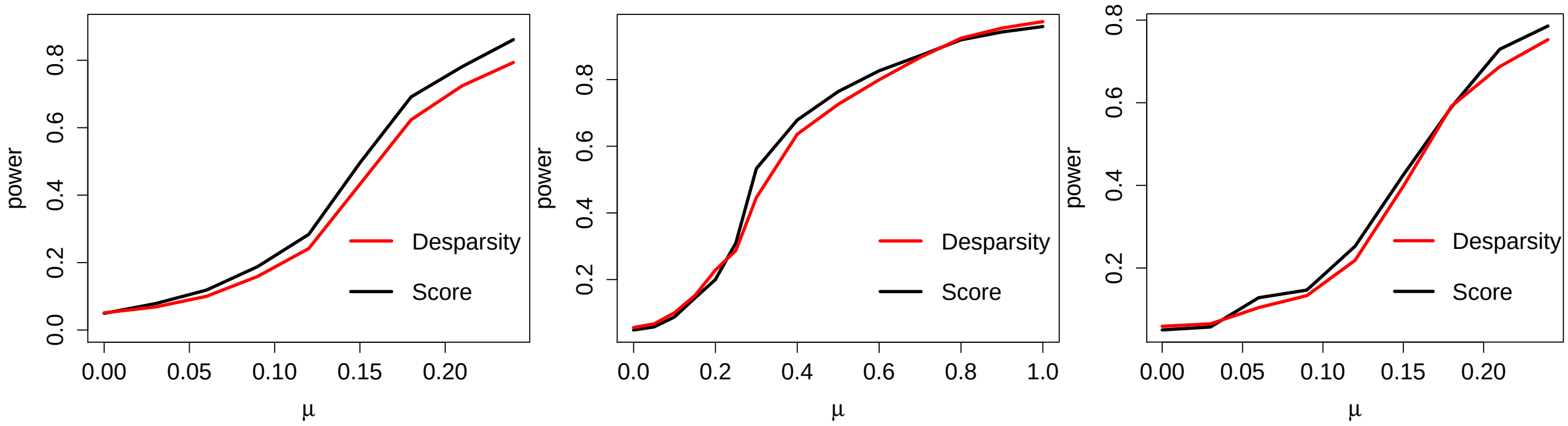

(a). Gaussian graphical model. (b). Ising model. (c). Mixed graphical model.

Figure 1: Power of the composite pairwise score test and the desparsity method in three models.

### 5.2 Real Data Analysis

We then apply the proposed methods to analyze a publicly available dataset named `Computer Audition Lab 500-Song (CAL500)` dataset (Turnbull et al., 2008). The data can be obtained from the `Mulan` database (Tsoumakas et al., 2011). The `CAL500` dataset consists of 502 popular music tracks each of which is annotated by at least three listeners. The attributes of this dataset include two subsets: (i) continuous numerical features extracted from the time series of the audio signal and (ii) discrete binary labels assigned by human listeners to give semantic descriptions of the song. For each music track, short time Fourier transform is implemented for a sequence of half-overlapping 23ms time windows over the song's digital audio file. This procedure generates four types of continuous features: *spectral centroids, spectral flux, zero crossings* and a time series of Mel-frequency cepstral coefficient (MFCC). For the MFCC vectors, every consecutive 502 short time windows are grouped together as a block window to produce the following four types of features: (i) overall mean of MFCC vectors in each block window, (ii) mean of standard deviations of MFCC vectors in each block window, (iii) standard deviation of the means of MFCC vectors in each block window, and (iv) standard deviation of the standard deviations of MFCC vectors in each block window. More details on feature extraction can be found in Tzanetakis and Cook (2002). In addition to these continuous variables, binary variables in the `CAL500` dataset are represented by a 174-dimensional array indicating the existence of each annotation. These 174 annotations can be grouped into six categories: emotions (36 variables), instruments (33), usages (15), genres (47), song characteristics (27) and vocal types (16). Our goal is to infer the association relationships between these different types of variables using semiparametric exponential family graphical models. This dataset has been analyzed in Cheng et al. (2013) where they exploit a nodewise group-LASSO regression to estimate the graph structure. In what follows, we use the proposed pairwise score test to examine the connection of each pair of nodes.

We model the `CAL500` dataset using the semiparametric exponential family graphical model. Similar to Turnbull et al. (2008) and Cheng et al. (2013), we only keep the MFCC features because they can be interpreted as the amplitude of the audio signal and the other continuous features are not readily interpretable. Unlike Cheng et al. (2013), we keep all the binary labels. Thus the processed dataset has $n = 502$ data points of dimension $d = 226$ with 52 continuous variables and 174 binary variables. We apply the pairwise score test to each pair of variables to determine the presence of an edge between them. The p-values for the null hypothesis that these two variables are conditionally independence given the rest of variables are kept. We then use the Bonferroni correction to control the familywise error rate at 0.05. We set the nonconvex penalty function in optimization problem (3.2) to be capped-$\ell_1$ penalty $p_\lambda(u) = \lambda \min\{u, \lambda\}$ with the regularization parameter $\lambda$ selected by 10-fold cross-validation as in the previous section. For numerical stability, in our analysis, we perform the above testing method 100 times on randomly drawn subsamples of size $n/2$ and report the edges selected at least 90 times. We present the fitted graph in Figure 2, where we plot the connected components and omit the singletons. To better display the graphical structure, we use a square to represent each type of 13 MFCC features respectively. If a node connects to any node within the group of variables in a MFCC node, we connect it with this MFCC node with an edge. We use circles to represent the binary variables and use different colors to indicate their categories. The obtained graph has some interesting properties. First of all, the continuous features are densely connected within themselves, which is similar to the results in Cheng et al. (2013). For connections between continuous and binary variables, we find that the noisiness of the music (square 4) is connected with not cheerful emotions (circle 16), rock music (circle 70), the instrument electric guitar (circle 79), songs are very catchy (circle 96, negatively correlated) and very tonic songs (circle 119, negatively correlated). We also find that both the average amplitude (square 1) and periodic amplitude variation (square 2) are connected with unpleasant songs (circle 30). Moreover, edges connecting two binary variables also display interesting patterns. For instance, we find that passionate emotions (circle 17) are connected with not passionate emotions (circle 18), soft rock music (circle 143), and emotional vocals (circle 58); very danceable songs (circle 122) are connected with the usage "at a party" (circle 123) and songs with fast tempo (circle 100); the songs with heavy beats (circle 103) are connected with folk songs (circle 65), and not tender emotions (circle 38). In addition, we find edges between songs with positive feelings (circle 108) and the carefree (circle 13), cheerful and festive (circle 15), happy (circle 21) and optimistic (circle 31) emotions, which has intuitive explanations.

In summary, the application of the proposed method to the `CAL500` dataset reveals some interesting associations between these variables and can be used as a useful complement for analyzing high dimensional datasets with more complex distributions.

# 6 Conclusion

We propose an integrated framework for uncertainty assessment of a new semiparametric exponential family graphical model. The novelty of our model is that the base measures of each nodewise conditional distribution are treated as unknown nuisance functions. Towards the goal of uncertainty assessment, we first adopt the adaptive multi-stage relaxation algorithm to perform the parameter estimation. Then we propose a composite pairwise score test for the presence of an edge in the graph. Our method provides a rigorous justification for the uncertainty assessment, and is further supported by extensive numerical results.

# Acknowledgement

The authors are grateful for the support of NSF CAREER Award DMS1454377, NSF IIS1408910, NSF IIS1332109, NIH R01MH102339, NIH R01GM083084, and NIH R01HG06841.

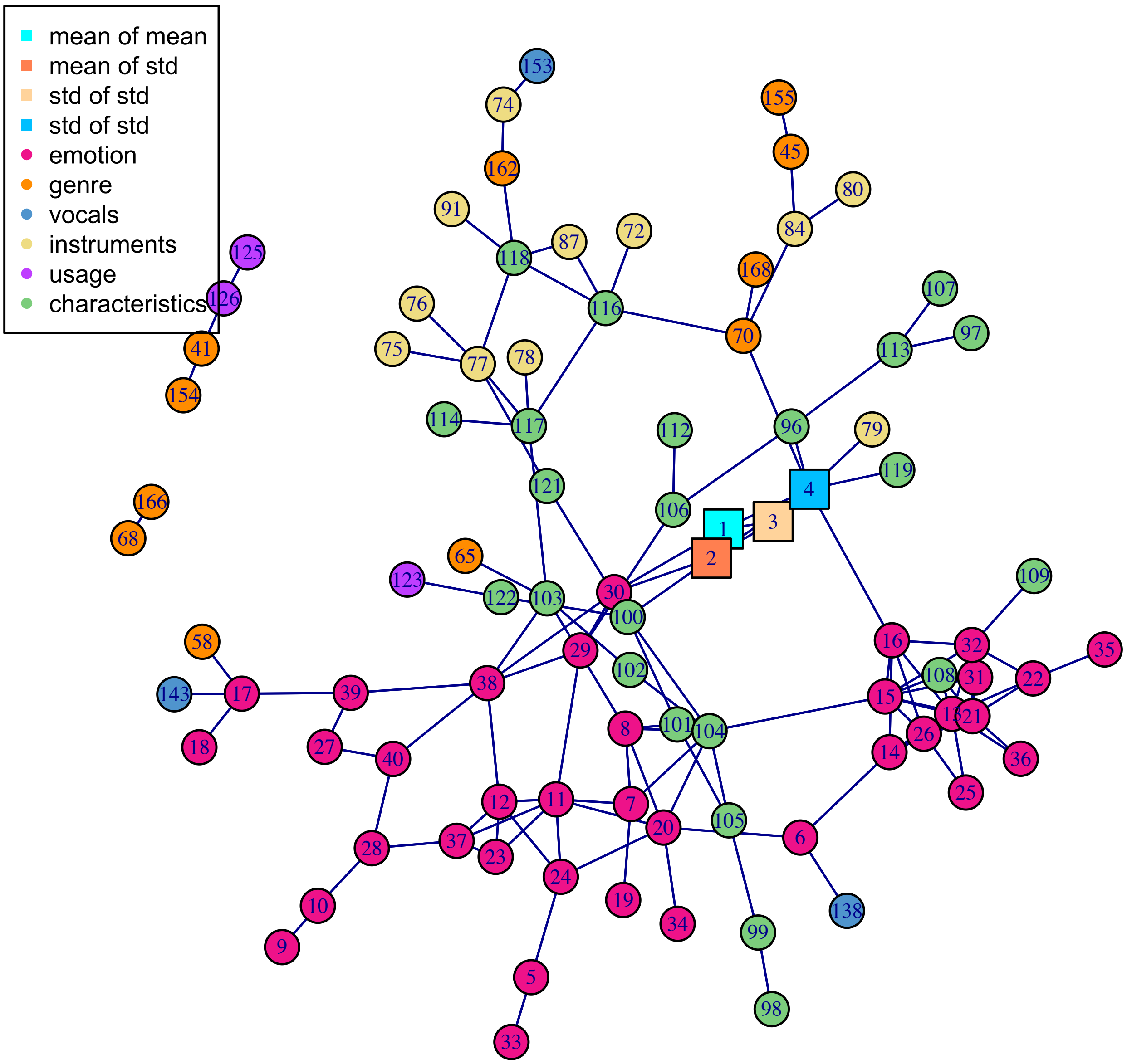

Figure 2: Estimated graph of the `CAL500` dataset inferred by pairwise score test. We apply the pairwise score test to the processed dataset with $n = 502$ instances and $d = 226$ attributes which includes 52 continuous variables of 4 types and 174 binary variables of 6 types. We plot the connected components of the estimated graph. We combine each type of continuous variables and display them as a square and draw each binary variable as a circle. The edges of the estimated graph shows the association of between these continuous and binary variables.

# A Proof of the Main Results

In this appendix we lay out the proof of the main results. In §A.1 we prove the result of parameter estimation. The proof is based an induction argument that Algorithm 1 keeps penalizing most of the irrelevant features and gradually reduces the bias in relevant features.

## A.1 Proof of Theorem 4.4

*Proof of Theorem 4.4.* We only need to prove the theorem for one node $j \in [d]$, the proof is identical for the rest. To begin with, we first define a few index sets that play a significant role in our analysis. For all $j \in [d]$, we let $S_j := \{(j,k) \colon \beta^*_{jk} \neq 0, k \in [d]\}$ be the support of $\boldsymbol{\beta}^*_j$. For the number of iterations $\ell = 1, 2, \ldots$, let $G^\ell_j := \big\{(j,k) \notin S_j \colon \lambda^{(\ell-1)}_{jk} \geq p'_\lambda(c_2\lambda), k \in [d]\big\}$. By condition (C.3) of the penalty function $p_\lambda(u)$ (see §3.2), we have $p'_\lambda(c_2\lambda) \geq 0.91\lambda$. In addition, we let $J^\ell_j$ be the largest $k^*$ components of $\big[\widehat{\boldsymbol{\beta}}^{(\ell)}_j\big]_{G^\ell_j}$ in absolute value where $k^*$ is defined in Assumption 4.3. In addition, we let $I^\ell_j = (G^\ell_j)^c \cup J^\ell_j$. Moreover, for notational simplicity, we denote $\big[\boldsymbol{\beta}_j\big]_{G^\ell_j}$, $\big[\boldsymbol{\beta}_j\big]_{G^\ell_j}$ and $\big[\boldsymbol{\beta}_j\big]_{I^\ell_j}$ as $\boldsymbol{\beta}_{G^\ell_j}, \boldsymbol{\beta}_{J^\ell_j}$ and $\boldsymbol{\beta}_{I^\ell_j}$ respectively when no ambiguity arises.

The key point of the proof is to show that the complement of $G^\ell_j$ is not too large. To be more specific, we show that $\big|(G^\ell_j)^c\big| \leq 2s^*$ for $\ell \geq 1$. Since $S_j \subset (G^\ell_j)^c$, $(G^\ell_j)^c \leq 2s^*$ implies $\big|(G^\ell_j)^c - S_j\big| \leq s^*$. Note that $G^\ell_j$ is the set of irrelevant features that are heavily penalized in the $\ell$-th iteration of the algorithm, $(G^\ell_j)^c - S$ being a small set indicates that the most of the irrelevant features are heavily penalized in each step. We show that $\big|(G^\ell)^c\big| \leq 2s^*$ for each $\ell \geq 1$ by induction.

For $\ell = 1$, we have $G^1_j = S^c_j$ because $\lambda^{(0)}_{jk} = \lambda$ for all $j,k \in [d]$. Hence $\big|(G^1_j)^c\big| \leq s^*$. Now we assume that $\big|\big(G^\ell_j\big)^c\big| \leq 2s^*$ for some integer $\ell$ and our goal is to prove that $\big|\big(G^{\ell+1}_j\big)^c\big| \leq 2s^*$. Our proof is based on three technical lemmas. The first lemma shows that the regularization parameter $\lambda$ in (3.2) is of the same order as $\|\nabla L_j(\boldsymbol{\beta}^*_j)\|_\infty$.

**Lemma A.1.** Under Assumptions 4.1 and 4.3, there exists a positive constants $K$ such that, it holds with probability at least $1 - (2d)^{-1}$ that

$$\big\|\nabla L_j(\boldsymbol{\beta}^*_j)\big\|_\infty \leq K\sqrt{\log d/n}, \quad \forall j \in [d]. \tag{A.1}$$

*Proof of Lemma A.1.* See §C.1 for a proof. □

By this lemma, we conclude that the regularization parameter $\lambda \geq 25\big\|\nabla L_j(\boldsymbol{\beta}^*_j)\big\|_\infty$ with high probability. The following lemma bounds the $\ell_1$- and $\ell_2$-norms of $\widehat{\boldsymbol{\beta}}^{(\ell)}_j - \boldsymbol{\beta}^*_j$ by the norms of its subvector under the induction assumption that $\big|(G^\ell_j)^c\big| \leq 2s^*$.

**Lemma A.2.** Letting the index sets $S_j, G^\ell_j, J^\ell_j$ and $I^\ell_j$ be defined as above, we denote $\widetilde{G}^\ell_j := \big(G^\ell_j\big)^c$. Under the assumption that $|G^\ell_j| \leq 2s^*$, we have

$$\big\|\widehat{\boldsymbol{\beta}}^{(\ell)}_j - \boldsymbol{\beta}^*_j\big\|_2 \leq 2.2\big\|\widehat{\boldsymbol{\beta}}^{(\ell)}_{I^\ell_j} - \boldsymbol{\beta}^*_{I^\ell_j}\big\|_2 \quad \text{and} \quad \big\|\widehat{\boldsymbol{\beta}}^{(\ell)}_j - \boldsymbol{\beta}^*_j\big\|_1 \leq 2.2\big\|\widehat{\boldsymbol{\beta}}^{(\ell)}_{\widetilde{G}^\ell_j} - \boldsymbol{\beta}^*_{\widetilde{G}^\ell_j}\big\|_1. \tag{A.2}$$

*Proof of Lemma A.2.* See §C.2 for a detailed proof.

□

The next lemma guarantees that $\widehat{\boldsymbol{\beta}}^{(\ell)}_j$ stayes in the $\ell_1$-ball centered at $\boldsymbol{\beta}^*_j$ with radius $r$ for $\ell \geq 1$ where $r$ appears in Assumption 4.3. Moreover, by showing this property of Algorithm 1 , we obtain a crude rate for parameter estimation. We summarized this result in the next lemma.

**Lemma A.3.** For $\ell \geq 1$ and $j \in [d]$, we denote $\boldsymbol{\lambda}_{S_j}^{(\ell)} := (\lambda_{jk}^{(\ell)}, (j,k) \in S_j)^T$. Assuming that $\big|(G_j^\ell)^c\big| \leq 2s^*$, it holds with probability at least $1 - d^{-1}$ that, for all $j \in [d]$, the estimators $\widehat{\boldsymbol{\beta}}_j^{(\ell)}$ obtained in each iteration of Algorithm 1 satisfy

$$\big\|\widehat{\boldsymbol{\beta}}_{I_j^\ell}^{(\ell)} - \boldsymbol{\beta}_{I_j^\ell}^*\big\|_2 \leq 10\rho_*^{-1}\Big[\big\|\nabla_{\widetilde{G}_j^\ell} L_j(\boldsymbol{\beta}_j^*)\big\|_2 + \big\|\boldsymbol{\lambda}_{S_j}^{(\ell-1)}\big\|_2\Big], \quad \widetilde{G}_j^\ell := (G_j^\ell)^c. \tag{A.3}$$

This implies the following crude rates of convergence for $\widehat{\boldsymbol{\beta}}_j^{(\ell)}$:

$$\big\|\widehat{\boldsymbol{\beta}}_j^{(\ell)} - \boldsymbol{\beta}_j^*\big\|_2 \leq 24\rho_*^{-1}\sqrt{s^*}\lambda \;\text{ and }\; \big\|\widehat{\boldsymbol{\beta}}_j^{(\ell)} - \boldsymbol{\beta}_j^*\big\|_1 \leq 33\rho_*^{-1}s^*\lambda. \tag{A.4}$$

*Proof of Lemma A.3.* See §C.3 for a detailed proof. □

Now we show that $\widetilde{G}_j^{\ell+1} = (G_j^{\ell+1})^c$ satisfies $|\widetilde{G}_j^{\ell+1}| \leq 2s^*$, which concludes our induction. Letting $A := (G_j^{\ell+1})^c - S_j$, by the definition of $G_j^{\ell+1}$, $(j,k) \in A$ implies that $(j,k) \notin S_j$ and $p'_\lambda\big(\big|\widehat{\beta}_{jk}^{(\ell)}\big|\big) \leq p'_\lambda(c_2\lambda)$. Hence by the concavity of $p_\lambda(\cdot)$, for any $(j,k) \in A$, $\big|\widehat{\beta}_{jk}^{(\ell)}\big| \geq c_2\lambda$. Therefore we have

$$\sqrt{|A|} \leq \big\|\widehat{\boldsymbol{\beta}}_A^{(\ell)}\big\|_2/(c_2\lambda) = \big\|\widehat{\boldsymbol{\beta}}_A^{(\ell)} - \boldsymbol{\beta}_A^*\big\|_2/(c_2\lambda) \leq 24\rho_*^{-1}\sqrt{s^*}/c_2 \leq \sqrt{s^*}, \tag{A.5}$$

where the first inequality follows from $|A| \leq \sum_{(j,k)\in A}\big|\widehat{\beta}_{jk}^{(\ell)}\big|^2/(c_2\lambda)^2$. Note that (A.5) implies that $\big|(G_j^{\ell+1})^c\big| \leq 2s^*$. Therefore by induction, $\big|(G_j^\ell)^c\big| \leq 2s^*$ for any $\ell \geq 1$.

Now we have shown that for $\ell \geq 1$ and $j \in [d]$, $\big|(G_j^\ell)^c\big| \leq 2s^*$ and the crude statistical rates (A.4) hold. In what follows, we derive the more refined rates (4.2) and (4.3).

**A refined bound for $\|\widehat{\boldsymbol{\beta}}_j^{(\ell)} - \boldsymbol{\beta}_j^*\|_2$ and $\|\widehat{\boldsymbol{\beta}}_j^{(\ell)} - \boldsymbol{\beta}_j^*\|_1$:** For notional simplicity, we let $\boldsymbol{\delta}^{(\ell)} = \widehat{\boldsymbol{\beta}}_j^{(\ell)} - \boldsymbol{\beta}_j^*$ and omit subscript $j$ in $S_j, G_j^\ell, J_j^\ell$ and $I_j^\ell$. We also denote $\widetilde{G}^\ell := (G^\ell)^c$. We first derive a recursive bound that links $\|\boldsymbol{\delta}_{I^\ell}^{(\ell)}\|_2$ to $\|\boldsymbol{\delta}_{I^{\ell-1}}^{(\ell-1)}\|_2$. Note that by (A.2), $\|\boldsymbol{\delta}^{(\ell)}\|_1 \leq 2.2\|\boldsymbol{\delta}_{\widetilde{G}^\ell}^{(\ell)}\|_1 \leq 2.2\sqrt{2s^*}\|\boldsymbol{\delta}_{\widetilde{G}^\ell}^{(\ell)}\|_2$. Hence we only need to control $\|\boldsymbol{\delta}_{I^\ell}^{(\ell)}\|_2$ to obtain the statistical rates of convergence for $\widehat{\boldsymbol{\beta}}_j^{(\ell)}$. By triangle inequality,

$$\big\|\nabla_{\widetilde{G}^\ell} L_j(\boldsymbol{\beta}_j^*)\big\|_2 \leq \big\|\nabla_S L_j(\boldsymbol{\beta}_j^*)\big\|_2 + \sqrt{|\widetilde{G}^\ell - S|}\big\|\nabla L_j(\boldsymbol{\beta}_j^*)\big\|_\infty.$$

Since $\lambda > 25\big\|\nabla L_j(\boldsymbol{\beta}_j^*)\big\|_\infty$, (A.5) implies that

$$\big\|\nabla_{\widetilde{G}^\ell} L_j(\boldsymbol{\beta}_j^*)\big\|_2 \leq \big\|\nabla_S L_j(\boldsymbol{\beta}_j^*)\big\|_2 + \big\|\boldsymbol{\delta}_A^{(\ell-1)}\big\|_2/(25c_2), \tag{A.6}$$

where $A := (G^\ell)^c - S \subset I^\ell$. Thus (A.6) can be written as

$$\big\|\nabla_{\widetilde{G}^\ell} L_j(\boldsymbol{\beta}_j^*)\big\|_2 \leq \big\|\nabla_S L_j(\boldsymbol{\beta}_j^*)\big\|_2 + \big\|\boldsymbol{\delta}_{I^\ell}^{(\ell-1)}\big\|_2/(25c_2). \tag{A.7}$$

Also notice that $\forall \beta_{jk} \in \mathbb{R}$, if $|\beta_{jk} - \beta_{jk}^*| \geq c_2\lambda$,

$$p'_\lambda(|\beta_{jk}|) \leq \lambda \leq |\beta_{jk} - \beta_{jk}^*|/c_2;$$

otherwise we have $|\beta^*_{jk}| - |\beta_{jk}| \le |\beta_{jk} - \beta^*_{jk}| < c_2\lambda$ and thus $p'_\lambda(|\beta_{jk}|) \le p'_\lambda\big(|\beta^*_{jk}| - c_2\lambda\big)$ by the concavity of $p_\lambda(\cdot)$. Hence the following inequality always holds:

$$p'_\lambda(|\beta_{jk}|) \le p'_\lambda\big(|\beta^*_{jk}| - c_2\lambda\big) + |\beta_{jk} - \beta^*_{jk}|/c_2. \tag{A.8}$$

Applying (A.8) to $\widehat{\boldsymbol{\beta}}_j^{(\ell-1)}$ we have

$$\|\boldsymbol{\lambda}_S^{(\ell-1)}\|_2 \le \Big[\sum_{(j,k)\in S} p'_\lambda\big(|\beta^*_{jk}| - c_2\lambda\big)^2\Big]^{1/2} + \Big[\sum_{(j,k)\in S} |\widehat{\beta}_{jk}^{(\ell-1)} - \beta^*_{jk}|^2\Big]^{1/2} \Big/ c_2,$$

which leads to

$$\|\boldsymbol{\lambda}_S^{(\ell-1)}\|_2 \le \Big[\sum_{(j,k)\in S} p'_\lambda\big(|\beta^*_{jk}| - c_2\lambda\big)^2\Big]^{1/2} + \|\boldsymbol{\delta}_{I^{\ell-1}}^{(\ell-1)}\|_2/c_2. \tag{A.9}$$

By (A.3), (A.7) and (A.9) we obtain

$$\|\boldsymbol{\delta}_{I^\ell}^{(\ell)}\|_2 \le 10\rho_*^{-1}\big[\|\nabla_S L_j(\boldsymbol{\beta}_j^*)\|_2 + \Upsilon_j\big] + \gamma\|\boldsymbol{\delta}_{I^{\ell-1}}^{(\ell-1)}\|_2,$$

where $\gamma := 11(c_2\rho_*)^{-1}$ and we define $\Upsilon_j := \big[\sum_{(j,k)\in S} p'_\lambda\big(|\beta^*_{jk}| - c_2\lambda\big)^2\big]^{1/2}$ for notational simplicity. Note that since $c_2 \ge 24\rho_*^{-1}$, we have $\gamma < 1$. By recursion we obtain

$$\|\boldsymbol{\delta}_{I^\ell}^{(\ell)}\|_2 \le 10\varrho\big[\|\nabla_S L_j(\boldsymbol{\beta}_j^*)\|_2 + \Upsilon_j\big] + \gamma^{\ell-1}\|\boldsymbol{\delta}_{I^1}^{(1)}\|_2, \tag{A.10}$$

where $\varrho := \rho_*^{-1}\cdot(1-\gamma)^{-1} = c_2(c_2\rho_* - 11)^{-1}$. Using $\|\widehat{\boldsymbol{\beta}}_j^{(\ell)} - \boldsymbol{\beta}_j^*\|_2 \le 2.2\|\widehat{\boldsymbol{\beta}}_{I_j^\ell}^{(\ell)} - \boldsymbol{\beta}_{I_j^\ell}^*\|_2$, we can bound $\|\widehat{\boldsymbol{\beta}}_j^{(\ell)} - \boldsymbol{\beta}_j^*\|_2$ by

$$\|\widehat{\boldsymbol{\beta}}_j^{(\ell)} - \boldsymbol{\beta}_j^*\|_2 \le 22\varrho\big[\|\nabla_{S_j} L_j(\boldsymbol{\beta}_j^*)\|_2 + \Upsilon_j\big] + 2.2\gamma^{\ell-1}\|\boldsymbol{\delta}_{I_j^1}^{(1)}\|_2.$$

Note that for $\ell = 1$, by (A.3) we have

$$\|\boldsymbol{\delta}_{I_j^1}^{(1)}\|_2 \le 10\rho_*^{-1}\sqrt{s^*}\big[\lambda + \sqrt{2}\|\nabla L_j(\boldsymbol{\beta}_j^*)\|_\infty\big] \le 11\rho_*^{-1}\sqrt{s^*}\lambda = c_2\gamma\sqrt{s^*}\lambda. \tag{A.11}$$

then we establish the following bound for $\|\widehat{\boldsymbol{\beta}}_j^{(\ell)} - \boldsymbol{\beta}_j^*\|_2$:

$$\|\widehat{\boldsymbol{\beta}}_j^{(\ell)} - \boldsymbol{\beta}_j^*\|_2 \le 22\varrho\big[\|\nabla_{S_j} L_j(\boldsymbol{\beta}_j^*)\|_2 + \Upsilon_j\big] + 2.2c_2\sqrt{s^*}\lambda\gamma^\ell. \tag{A.12}$$

Similarly, by $\|\widehat{\boldsymbol{\beta}}_j^{(\ell)} - \boldsymbol{\beta}_j^*\|_1 \le 2.2\sqrt{2s^*}\|\widehat{\boldsymbol{\beta}}_{I_j^\ell}^{(\ell)} - \boldsymbol{\beta}_{I_j^\ell}^*\|_2$, we obtain a bound on $\|\widehat{\boldsymbol{\beta}}_j^{(\ell)} - \boldsymbol{\beta}_j^*\|_1$:

$$\|\widehat{\boldsymbol{\beta}}_j^{(\ell)} - \boldsymbol{\beta}_j^*\|_1 \le 32\sqrt{s^*}\varrho\big[\|\nabla_{S_j} L_j(\boldsymbol{\beta}_j^*)\|_2 + \Upsilon_j\big] + 2.2\gamma^{\ell-1}\sqrt{2s^*}\|\boldsymbol{\delta}_{I_j^1}^{(1)}\|_2. \tag{A.13}$$

By (A.11) we have $2.2\sqrt{2s^*}\|\boldsymbol{\delta}_{I_j^1}^{(1)}\|_2 \le 3.2c_2\gamma s^*\lambda$, then the right-hand side of (A.13) can be bounded by

$$\|\widehat{\boldsymbol{\beta}}_j^{(\ell)} - \boldsymbol{\beta}_j^*\|_1 \le 32\sqrt{s^*}\varrho\big[\|\nabla_{S_j} L_j(\boldsymbol{\beta}_j^*)\|_2 + \Upsilon_j\big] + 3.2c_2 s^*\lambda\gamma^\ell. \tag{A.14}$$

Therefore (4.2) and (4.3) can be implied by (A.12) and (A.14) respectively. Moreover, by Lemma A.3, we conclude that the statistical rates (A.12) and (A.14) hold for all $j \in [d]$ with probability at least $1 - d^{-1}$.

□

## A.2 Proof of Theorem 4.7

*Proof of Theorem 4.7.* We first remind the reader that for $1 \le j \ne k \le d$, we denote $\boldsymbol{\beta}_j = (\beta_{j1}, \ldots, \beta_{jj-1}, \beta_{jj+1}, \ldots, \beta_{jd})^T \in \mathbb{R}^{d-1}$, $\boldsymbol{\beta}_{j\setminus k} = (\beta_{j1}, \ldots, \beta_{jj-1}, \beta_{jj+1}, \ldots, \beta_{jk-1}, \beta_{jk+1}, \ldots, \beta_{jd})^T \in \mathbb{R}^{d-2}$, $\boldsymbol{\beta}_{j\vee k} = (\beta_{jk}, \boldsymbol{\beta}_{j\setminus k}, \boldsymbol{\beta}_{k\setminus j})^T \in \mathbb{R}^{2d-3}$ and $\widehat{\boldsymbol{\beta}}'_{j\vee k} = (0, \widehat{\boldsymbol{\beta}}_{j\setminus k}, \widehat{\boldsymbol{\beta}}_{k\setminus j})^T$. In addition, we define $\sigma_{jk}^2 = \boldsymbol{\Sigma}^{jk}_{jk,jk} - 2\boldsymbol{\Sigma}^{jk}_{jk,j\setminus k}\mathbf{w}^*_{j,k} - 2\boldsymbol{\Sigma}^{jk}_{jk,k\setminus j}\mathbf{w}^*_{k,j} + \mathbf{w}^{*T}_{j,k}\boldsymbol{\Sigma}^{jk}_{j\setminus k,j\setminus k}\mathbf{w}^*_{j,k} + \mathbf{w}^{*T}_{k,j}\boldsymbol{\Sigma}^{jk}_{k\setminus j,k\setminus j}\mathbf{w}^*_{k,j}$. To prove the theorem our goal is to prove the following two arguments:

$$\lim_{n\to\infty} \max_{j<k} \sqrt{n}\big|\widehat{S}_{jk} - S_{jk}(\boldsymbol{\beta}^*_{j\vee k})\big| = 0 \ \text{ and } \ \lim_{n\to\infty} \max_{j<k} |\widehat{\sigma}_{jk} - \sigma_{jk}| = 0. \tag{A.15}$$

Note that by Lemma A.5, $\sigma_{jk}^2$ is the asymptotic variance of $\sqrt{n}/2 \cdot S_{jk}(\boldsymbol{\beta}^*_{j\vee k})$. Thus combining (A.15) and Slutsky's theorem yields the theorem. By the the expression of $S_{jk}(\boldsymbol{\beta}^*_{j\vee k})$ and $\widehat{S}_{jk}$ in (4.4) and (4.5), under null hypothesis, for a fixed pair of nodes $j$ and $k$, we have $\widehat{S}_{jk} - S_{jk}(\boldsymbol{\beta}^*_{j\vee k}) = I_{1j} + I_{2j} + I_{1k} + I_{2k}$ where $I_{1j}$ and $I_{2j}$ are defined as

$$\begin{aligned} I_{1j} &:= \big[\nabla_{jk} L_j(\widehat{\boldsymbol{\beta}}'_j) - \nabla_{jk} L_j(\boldsymbol{\beta}^*_j)\big] - \widehat{\mathbf{w}}^T_{j,k}\big[\nabla_{j\setminus k} L_j(\widehat{\boldsymbol{\beta}}'_j) - \nabla_{j\setminus k} L_j(\boldsymbol{\beta}^*_j)\big] \ \text{ and} \\ I_{2j} &:= (\mathbf{w}^*_{j,k} - \widehat{\mathbf{w}}_{j,k})^T \nabla_{j\setminus k} L_j(\boldsymbol{\beta}^*_j); \end{aligned}$$

whereas $I_{1k}$ and $I_{2k}$ are defined by interchanging $j$ and $k$ in $I_{1j}$ and $I_{2j}$:

$$\begin{aligned} I_{1k} &:= \big[\nabla_{kj} L_k(\widehat{\boldsymbol{\beta}}'_k) - \nabla_{jk} L_k(\boldsymbol{\beta}^*_k)\big] - \widehat{\mathbf{w}}^T_{k,j}\big[\nabla_{k\setminus j} L_k(\widehat{\boldsymbol{\beta}}'_k) - \nabla_{k\setminus j} L_k(\boldsymbol{\beta}^*_k)\big] \ \text{ and} \\ I_{2k} &:= (\mathbf{w}^*_{k,j} - \widehat{\mathbf{w}}_{k,j})^T \nabla_{k\setminus j} L_j(\boldsymbol{\beta}^*_k). \end{aligned}$$

We first bound $I_{1j}$. Recall that $\widehat{\boldsymbol{\beta}}'_j = (0, \widehat{\boldsymbol{\beta}}_{j\setminus k})^T$. Note that under the null hypothesis, $\beta^*_{jk} = 0$, by the Mean-Value Theorem, there exists a $\widetilde{\boldsymbol{\beta}}_{j\setminus k} \in \mathbb{R}^{d-2}$ in the line segment between $\widehat{\boldsymbol{\beta}}_{j\setminus k}$ and $\boldsymbol{\beta}^*_{j\setminus k}$ such that

$$I_{1j} = \big[\widetilde{\boldsymbol{\Lambda}}_{jk,j\setminus k} - \widehat{\mathbf{w}}^T_{j,k}\widetilde{\boldsymbol{\Lambda}}_{j\setminus k,j\setminus k}\big](\widehat{\boldsymbol{\beta}}_{j\setminus k} - \boldsymbol{\beta}^*_{j\setminus k}),$$

where $\widetilde{\boldsymbol{\Lambda}} := \nabla^2 L_j(0, \widetilde{\boldsymbol{\beta}}_{j\setminus k})$. We let $\boldsymbol{\delta} := \widehat{\boldsymbol{\beta}}'_j - \boldsymbol{\beta}^*_j$ and denote $\nabla^2 L_j(\widehat{\boldsymbol{\beta}}'_j)$ and $\nabla^2(\boldsymbol{\beta}^*_j)$ as $\boldsymbol{\Lambda}$ and $\boldsymbol{\Lambda}^*$ respectively. From the definition of Dantzig selector we obtain

$$\begin{aligned} |I_{1j}| \le\ & \underbrace{\|\boldsymbol{\Lambda}_{jk,j\setminus k} - \widehat{\mathbf{w}}^T\boldsymbol{\Lambda}_{j\setminus k,j\setminus k}\|_\infty \|\boldsymbol{\delta}_{j\setminus k}\|_1}_{I_{11}} + \underbrace{\|\boldsymbol{\Lambda}_{jk,j\setminus k} - \widetilde{\boldsymbol{\Lambda}}_{jk,j\setminus k}\|_\infty \|\boldsymbol{\delta}_{j\setminus k}\|_1}_{I_{12}} \\ & + \underbrace{\|\widehat{\mathbf{w}}^T(\boldsymbol{\Lambda}_{j\setminus k,j\setminus k} - \widetilde{\boldsymbol{\Lambda}}_{j\setminus k,j\setminus k})\boldsymbol{\delta}_{j\setminus k}\|_\infty}_{I_{13}}. \end{aligned}$$

Theorem 4.4 implies that $\|\boldsymbol{\delta}\|_1 \le Cs^*\lambda$ with probability tending to 1 for some constant $C > 0$. Then by the definition of Dantzig selector, $I_{11} \le Cs^*\lambda\lambda_D$. with high probability. Moreover, the constant $C$ is the same for all $(j, k)$. By assumption 4.6, $I_{11} = o(n^{-1/2})$ with probability tending to one.

For term $I_{12}$, Hölder's inequality implies that

$$I_{12} \le \|\boldsymbol{\Lambda}_{jk,j\setminus k} - \widetilde{\boldsymbol{\Lambda}}_{jk,j\setminus k}\|_\infty \|\boldsymbol{\delta}_{j\setminus k}\|_1. \tag{A.16}$$

By Lemma E.1 we obtain

$$\|\boldsymbol{\Lambda} - \widetilde{\boldsymbol{\Lambda}}\|_\infty \le \|\boldsymbol{\Lambda} - \boldsymbol{\Lambda}^*\|_\infty + \|\boldsymbol{\Lambda}^* - \widetilde{\boldsymbol{\Lambda}}\|_\infty \le 2Cs^*\lambda \log^2 d. \tag{A.17}$$

Therefore combining (A.16) and (A.17) we have

$$I_{12} \le 2Cs^{*2}\lambda^2 \log^2 d \lesssim s^*\lambda\lambda_D \quad \text{uniformly for } 1 \le j < k \le d.$$

Similarly by Hölder's inequality, we have

$$I_{13} \le \|\widehat{\mathbf{w}}_{j,k}\|_1 \|\boldsymbol{\Lambda} - \widetilde{\boldsymbol{\Lambda}}\|_\infty \|\boldsymbol{\delta}\|_1. \tag{A.18}$$

Notice that by the optimality of $\widehat{\mathbf{w}}_{j,k}$, $\|\widehat{\mathbf{w}}_{j,k}\|_1 \le \|\mathbf{w}^*_{j,k}\|_1 \le w_0$. Combining (A.18) and (A.17) we have

$$I_{13} \le Cw_0 s^{*2}\lambda^2 \log^2 d \lesssim s^*\lambda\lambda_D \quad \text{uniformly for } 1 \le j < k \le d.$$

where we use the fact that $\lambda_D \gtrsim \max\{1, w_0\} s^*\lambda \log^2 d$. Therefore we conclude that for all $j \in [d]$, $|I_{1j}| \lesssim s^*\lambda\lambda_D = o_{\mathbb{P}}(n^{-1/2})$. For $I_{2j}$, Hölder's inequality implies that $|I_{2j}| \le \|\mathbf{w}^*_{j,k} - \widehat{\mathbf{w}}_{j,k}\|_1 \big\|\nabla L_j(\boldsymbol{\beta}^*_j)\big\|_\infty$. To control $\|\mathbf{w}^*_{j,k} - \widehat{\mathbf{w}}_{j,k}\|_1$, we need to the following lemma to obtain the estimation error of the Dantzig selector $\widehat{\mathbf{w}}_{j,k}$.

**Lemma A.4.** For $1 \le j \ne k \le d$, let $\widehat{\mathbf{w}}_{j,k}$ be the solution of the Dantzig-type optimization problem (3.5) and let $\mathbf{w}^*_{j,k} = \mathbf{H}^j_{jk,j\setminus k}(\mathbf{H}^j_{j\setminus k,j\setminus k})^{-1}$. Under Assumptions 4.1, 4.3, 4.5 and 4.6, with probability tending to one, we have

$$\|\widehat{\mathbf{w}}_{j,k} - \mathbf{w}^*_{j,k}\|_1 \le 37\nu_*^{-1} s_0^\star \lambda_D \quad \text{for all} \quad 1 \le j \ne k \le d.$$

*Proof of Lemma A.4.* See §D.2 for a detailed proof. □

Now combining Lemma A.4 and Theorem A.1 we obtain that

$$|I_{2j}| \le 37\nu_*^{-1} K_1 s_0^\star \lambda_D \sqrt{\log d/n} \asymp s_0^\star \lambda\lambda_D = o(n^{-1/2}).$$

Therefore we have shown that $I_{1j} + I_{2j} = o(n^{-1/2})$ with high probability. Similarly, we also have $I_{1k} + I_{2k} = o(n^{-1/2})$ with high probability. Moreover, since the bounds for $|I_{1j}|$ and $|I_{2j}|$ is independent of the choice of $(j,k) \in \{(j,k) \colon 1 \le j \ne k \le d\}$, we conclude that

$$\sqrt{n}\big[\widehat{S}_{jk} - S_{jk}\big(\boldsymbol{\beta}^*_{j\vee k}\big)\big] = o_{\mathbb{P}}(1) \quad \text{uniformly for } 1 \le j < k \le d.$$

Our next lemma characterizes the limiting distribution of $\nabla L_{jk}\big(\boldsymbol{\beta}^*_{j\vee k}\big)$ and is pivotal for establishing the validity of the composite pairwise score test.

**Lemma A.5.** For any $\mathbf{b} \in \mathbb{R}^{2d-3}$ with $\|\mathbf{b}\|_2 = 1$ and $\|\mathbf{b}\|_0 \le \widetilde{s}$, if $\lim_{n\to\infty} \widetilde{s}/n = 0$, we have

$$\sqrt{n}/2 \cdot \mathbf{b}^T \nabla L_{jk}\big(\boldsymbol{\beta}^*_{j\vee k}\big) \rightsquigarrow N\big(0, \mathbf{b}^T \boldsymbol{\Sigma}^{jk}\mathbf{b}\big). \tag{A.19}$$

By Lemma A.5 we obtain

$$\sqrt{n}/2 \cdot S(\boldsymbol{\beta}^*_{j\vee k}) = \nabla_{jk} L_{jk}(\boldsymbol{\beta}^*_{j\vee k}) - \mathbf{w}^{*T}_{j,k} \nabla_{j\backslash k} L_{jk}(\boldsymbol{\beta}^*_{j\vee k}) - \mathbf{w}^{*T}_{k,j} \nabla_{j\backslash k} L_{jk}(\boldsymbol{\beta}^*_{j\vee k}) \rightsquigarrow N(0, \sigma^2_{jk}),$$

where the asymptotic variance $\sigma^2_{jk}$ is given by

$$\sigma^2_{jk} = \boldsymbol{\Sigma}^{jk}_{jk,jk} - 2\boldsymbol{\Sigma}^{jk}_{jk,j\backslash k}\mathbf{w}^*_{j,k} - 2\boldsymbol{\Sigma}_{jk,k\backslash j}\mathbf{w}^*_{k,j} + \mathbf{w}^{*T}_{j,k}\boldsymbol{\Sigma}^{jk}_{j\backslash k,j\backslash k}\mathbf{w}^*_{j,k} + \mathbf{w}^{*T}_{k,j}\boldsymbol{\Sigma}^{jk}_{k\backslash j,k\backslash j}\mathbf{w}^*_{k,j}.$$

For a more accurate estimation of $\widehat{S}_{jk} - S_{jk}(\boldsymbol{\beta}^*_{j\vee k})$, we have

$$\sqrt{n}\big|\widehat{S}_{jk} - S_{jk}(\boldsymbol{\beta}^*_{j\vee k})\big| \le \sqrt{n}\big(|I_1| + |I_2|\big) \lesssim \sqrt{n}(s^* + s^\star_0)\lambda\lambda_D. \tag{A.20}$$

Finally, the following lemma, whose proof is deferred to the supplementary material, shows that $\widehat{\sigma}_{jk}$ is a consistent estimator of $\sigma_{jk}$.

**Lemma A.6.** For $1 \le j \ne k \le d$, we denote the asymptotic variance of $\sqrt{n}/2 \cdot S_{jk}(\boldsymbol{\beta}^*_{j\vee k})$ as $\sigma^2_{jk}$. Under Assumptions 4.1, 4.3, 4.5 and 4.6, the estimator $\widehat{\sigma}_{jk}$ satisfies $\lim_{n\to\infty} \max_{j<k} |\widehat{\sigma}_{jk} - \sigma_{jk}| = 0$.

*Proof of Lemma A.6.* See §D.3 for a proof. □

Since $\widehat{\sigma}_{jk}$ is consistent for $\sigma_{jk}$ by Lemma A.6 and $\sigma_{jk}$ is bounded away from zero by Assumption 4.5, Slutsky's theorem implies that $\sqrt{n}\widehat{S}_{jk}/(2\widehat{\sigma}_{jk}) \rightsquigarrow N(0,1)$. □

# B Additional Estimation Results

We present the additional results of parameter estimation. In §B.1 we verify the sparse eigenvalue condition for Gaussian graphical models, which justifies Assumption 4.3 in our paper. In §B.2 we derive a more refined statistical rates of convergence for the iterates of Algorithm 1.

## B.1 Verify the Sparse Eigenvalue Condition for Gaussian Graphical Models

In this subsection, we verify the sparse eigenvalue condition for Gaussian graphical models. Moreover, we show that such condition holds uniformly over a $\ell_1$-ball centered at the true parameter $\boldsymbol{\beta}^*_j$.

**Proposition B.1** (Sparse eigenvalue condition for Gaussian graphical models)**.** Suppose $\boldsymbol{X} \sim N(\mathbf{0}, \boldsymbol{\Sigma})$ is a Gaussian graphical model and let $\boldsymbol{\Theta} = \boldsymbol{\Sigma}^{-1}$ be the precision matrix. For all $j \in [d]$, the conditional distribution of $X_j$ given $\boldsymbol{X}_{\backslash j}$ is a normal distribution with mean $\boldsymbol{\beta}^{*T}_j \boldsymbol{X}_{\backslash j}$ and variance $\boldsymbol{\Theta}^{-1}_{jj}$, where $\boldsymbol{\beta}^*_j = \boldsymbol{\Theta}_{j\backslash j}$. Let $L_j(\cdot)$ be the loss function defined in (3.1). We assume that there exist positive constants $D$, $c_\lambda$ and $C_\lambda$ such that $\|\boldsymbol{\Sigma}\|_\infty \le D$ and $c_\lambda \le \lambda_{\min}(\boldsymbol{\Sigma}) \le \lambda_{\max}(\boldsymbol{\Sigma}) \le C_\lambda$. We let $s^* = \max_{j\in[d]} \|\boldsymbol{\beta}^*_j\|_0$ and also assume that there exists a constant $C_\beta > 0$ such that $\|\boldsymbol{\beta}^*_j\|_2 \le C_\beta$ for all $j \in [d]$. Suppose $r > 0$ is a real number such that $r = \mathcal{O}(1/\sqrt{s^*})$. Then, there exist $\rho_*, \rho^* > 0$ such that for all $j \in [d]$, and $s = 1, \ldots, d-1$,

$$\rho_* \le \rho_-\big(\mathbb{E}[\nabla^2 L_j], \boldsymbol{\beta}^*_j; s, r\big) \le \rho_+\big(\mathbb{E}[\nabla^2 L_j], \boldsymbol{\beta}^*_j; s, r\big) \le \rho^*.$$

*Proof of Proposition B.1.* We prove this lemma in two steps. For any $\boldsymbol{\beta}_j \in \mathbb{R}^{d-1}$ such that $\|\boldsymbol{\beta}_j - \boldsymbol{\beta}_j^*\|_1 \leq r$ and any $\mathbf{v} \in \mathbb{R}^{d-1}$ such that $\|\mathbf{v}\|_2 = 1$, we first give a lower bound for $\mathbf{v}^T\mathbb{E}\big[\nabla^2 L_j(\boldsymbol{\beta}_j)\big]\mathbf{v}$ by truncation. Then we give an upper bound in the second step.

**Step (i): Lower Bound of $\mathbf{v}^T\mathbb{E}\big[\nabla^2 L_j(\boldsymbol{\beta}_j)\big]\mathbf{v}$.** We denote $\mathbb{B}_j(r) := \big\{\boldsymbol{\beta} \in \mathbb{R}^{d-1} \colon \|\boldsymbol{\beta} - \boldsymbol{\beta}_j^*\|_1 \leq r\big\}$. For two truncation levels $\tau > 0$ and $R > 0$, we denote $\mathcal{A}_{ii'} := \big\{|X_{ij}| \leq \tau\big\} \cap \big\{|X_{i'j}| \leq \tau\big\}$, $\mathcal{B}_i := \big\{\big|\boldsymbol{X}_{i\backslash j}^T\boldsymbol{\beta}_j\big| \leq R, \forall \boldsymbol{\beta}_j \in \mathbb{B}_j(r)\big\}$ and $\mathcal{B}_{i'} := \big\{\big|\boldsymbol{X}_{i'\backslash j}^T\boldsymbol{\beta}_j^*\big| \leq R, \forall \boldsymbol{\beta}_j \in \mathbb{B}_j(r)\big\}$. The values of $R$ and $\tau$ will be determined later. By the definition of $L_j(\cdot)$, for any $\boldsymbol{\beta}_j \in \mathbb{B}_j(r)$ and any $\mathbf{v} \in \mathbb{R}^{d-1}$ with $\|\mathbf{v}\|_2 = 1$, we have

$$\mathbf{v}^T\nabla^2 L_j(\boldsymbol{\beta}_j)\mathbf{v} \geq \frac{2C_1(R,\tau)}{n(n-1)}\sum_{i<i'}\big(X_{ij} - X_{i'j}\big)^2\big[\big(\boldsymbol{X}_{i\backslash j} - \boldsymbol{X}_{i'\backslash j}\big)^T\mathbf{v}\big]^2 I(\mathcal{B}_i)T(\mathcal{B}_{i'})I(\mathcal{A}_{ii'}), \tag{B.1}$$

where $C_1(R,\tau) := \exp(-4R\tau)\big[1 + \exp(-4R\tau)\big]^{-2}$. For notational simplicity, we denote the right-hand side of (B.1) as $C_1(R,\tau)\mathbf{v}^T\boldsymbol{\Delta}\mathbf{v}$. By the properties of Gaussian graphical models, the conditional density of $X_{ij}$ given $\mathcal{I}_i := \big\{\boldsymbol{X}_{i\backslash j} = \boldsymbol{x}_{i\backslash j}\big\} \cap \mathcal{B}_i$ is

$$p\big(x_{ij}|\mathcal{I}_i\big) = p(\boldsymbol{x}_i|\mathcal{B}_i)\Big/\int_{\mathbb{R}} p(\boldsymbol{x}_i|\mathcal{B}_i)\mathrm{d}x_{ij} = p(x_{ij}|\boldsymbol{x}_{i\backslash j}),$$

where we use the fact that $p(\boldsymbol{x}_i|\mathbb{B}_i) = p(\boldsymbol{x}_i)/\mathbb{P}(\mathcal{B}_i)$ and that $\mathbb{P}(\mathcal{B}_i)$ is a constant. Recall that

$$p(x_{ij}|\boldsymbol{X}_{i\backslash j}) = \sqrt{\boldsymbol{\Theta}_{jj}/(2\pi)}\exp\big[-\boldsymbol{\Theta}_{jj}/2(x_{ij} - \boldsymbol{X}_{i\backslash j}^T\boldsymbol{\beta}_j^*)^2\big] \quad \text{where } \boldsymbol{\beta}_j^* = \boldsymbol{\Theta}_{j\backslash j}.$$

Thus the conditional expectation of $(X_{ij} - X_{i'j})^2 I(\mathcal{A}_{ij})$ given $\mathcal{I}_i$ and $\mathcal{I}_{i'}$ is

$$\begin{aligned}&\mathbb{E}\Big[(X_{ij} - X_{i'j})^2 I(\mathcal{A}_{ii'})\Big|\mathcal{I}_i \cap \mathcal{I}_{i'}\Big]\\ =&\boldsymbol{\Theta}_{jj}/(2\pi)\int_{-\tau}^{\tau}\int_{-\tau}^{\tau}(x_{ij} - x_{ij})^2\exp\Big\{-\boldsymbol{\Theta}_{jj}/2\big[(x_{ij} - \boldsymbol{\beta}_j^T\boldsymbol{x}_{i\backslash j})^2 + (x_{i'j} - \boldsymbol{\beta}_j^T\boldsymbol{x}_{i'\backslash j})^2\big]\Big\}\mathrm{d}x_{ij}\mathrm{d}x_{i'j}.\end{aligned}$$

Note that on event $\mathcal{I}_i$, $|\boldsymbol{\beta}_j^T\boldsymbol{X}_{i\backslash j}| \leq R$, hence the expression above can be lower-bounded by

$$\begin{aligned}&\mathbb{E}\Big[(X_{ij} - X_{i'j})^2 I(\mathcal{A}_{ii'})\Big|\mathcal{I}_i \cap \mathcal{I}_{i'}\Big]\\ \geq&\boldsymbol{\Theta}_{jj}/(2\pi)\int_{-\tau}^{\tau}\int_{-\tau}^{\tau}(x_{ij} - x_{i'j})^2\exp\Big\{-\boldsymbol{\Theta}_{jj}/2\big[x_{ij}^2 + x_{i'j}^2 + 2R^2 + 2R(|x_{ij}| + |x_{i'j}|)\big]\Big\}\mathrm{d}x_{ij}\mathrm{d}x_{i'j}.\end{aligned}$$

The last expression is positive and we denote it as $C_2(R,\tau)$ for simplicity. Thus by the law of total expectation we obtain

$$\mathbf{v}^T\mathbb{E}(\boldsymbol{\Delta})\mathbf{v} = \mathbf{v}^T\mathbb{E}\big[\mathbb{E}(\boldsymbol{\Delta}\big|\cap_{i=1}^n\mathcal{I}_i)\big]\mathbf{v} \geq C_2(R,\tau)\mathbb{E}\Big\{\big[(\boldsymbol{X}_{i\backslash j} - \boldsymbol{X}_{i'\backslash j})^T\mathbf{v}\big]^2 I(\mathcal{B}_i)I(\mathcal{B}_j)\Big\}.$$

By Cauchy-Schwarz inequality we have

$$\mathbb{E}\Big\{\big[(\boldsymbol{X}_{i\backslash j} - \boldsymbol{X}_{i'\backslash j})^T\mathbf{v}\big]^2\big[1 - I(\mathcal{B}_i)I(\mathcal{B}_{i'})\big]\Big\} \leq \sqrt{\mathbb{E}[(\boldsymbol{X}_{i\backslash j} - \boldsymbol{X}_{i'\backslash j})^T\mathbf{v}\big]^4}\sqrt{\mathbb{P}\big(\mathcal{B}_i^c \cup \mathcal{B}_{i'}^c\big)}. \tag{B.2}$$

Note that for Gaussian graphical model, the marginal distribution of $\boldsymbol{X}_{\backslash j}$ is $N(\mathbf{0}, \boldsymbol{\Sigma}_{\backslash j\backslash j})$. If we denote $\boldsymbol{\Sigma}_{\backslash j\backslash j}$ as $\boldsymbol{\Sigma}_1$, we have $(\boldsymbol{X}_{i\backslash j}–\boldsymbol{X}_{i'\backslash j})^T\mathbf{v} \sim N(\mathbf{0}, \sigma_v^2)$, $\boldsymbol{X}_{i\backslash j}^T\boldsymbol{\beta}_j^* \sim N(\mathbf{0}, \sigma_1^2)$ and $\boldsymbol{X}_{i\backslash j}^T\boldsymbol{\beta} \sim N(\mathbf{0}, \sigma_2^2)$ where $\sigma_v^2 = 2\mathbf{v}^T\mathbf{\Sigma}_1\mathbf{v}$, $\sigma_1^2 = \boldsymbol{\beta}_j^{*T}\mathbf{\Sigma}_1\boldsymbol{\beta}_j^*$ and $\sigma_2^2 = \boldsymbol{\beta}_j^T\mathbf{\Sigma}_1\boldsymbol{\beta}_j$. Hence we have $\mathbb{E}\big[(\boldsymbol{X}_{i\setminus j} - \boldsymbol{X}_{i'\setminus j})^T\mathbf{v}\big]^4 = 3\sigma_v^4$. Because the maximum eigenvalue of $\mathbf{\Sigma}_1$ is upper bounded by $C_\lambda$, we have $\sigma_1^2 \le C_\lambda C_\beta^2$ and $\sigma_v^2 \le 2C_\lambda$. Note that $\sigma_2^2 - \sigma_1^2 = \boldsymbol{\beta}_j^T\mathbf{\Sigma}_1\boldsymbol{\beta}_j - \boldsymbol{\beta}_j^{*T}\mathbf{\Sigma}_1\boldsymbol{\beta}_j^*$, the following lemma in linear algebra bounds this type of error:

**Lemma B.2.** Let $\mathbf{M} \in \mathbb{R}^{d\times d}$ be a symmetric matrix and vectors $\mathbf{v}_1$ and $\mathbf{v}_2 \in \mathbb{R}^d$, then

$$\big|\mathbf{v}_1^T\mathbf{M}\mathbf{v}_1 - \mathbf{v}_2^T\mathbf{M}\mathbf{v}_2\big| \le \|\mathbf{M}\|_\infty\|\mathbf{v}_1 - \mathbf{v}_2\|_1^2 + 2\|\mathbf{M}\mathbf{v}_2\|_\infty\|\mathbf{v}_1 - \mathbf{v}_2\|_1.$$

*Proof of Lemma B.2.* Note that $\mathbf{v}_1^T\mathbf{M}\mathbf{v}_1 - \mathbf{v}_2^T\mathbf{M}\mathbf{v}_2 = (\mathbf{v}_1 - \mathbf{v}_2)^T\mathbf{M}(\mathbf{v}_1 - \mathbf{v}_2) + 2\mathbf{v}_2^T\mathbf{M}(\mathbf{v}_1 - \mathbf{v}_2)$, Hölder's inequality implies

$$\begin{aligned}\big|\mathbf{v}_1^T\mathbf{M}\mathbf{v}_1 - \mathbf{v}_2^T\mathbf{M}\mathbf{v}_2\big| &\le \big|(\mathbf{v}_1 - \mathbf{v}_2)^T\mathbf{M}(\mathbf{v}_1 - \mathbf{v}_2)\big| + 2\big|\mathbf{v}_2^T\mathbf{M}(\mathbf{v}_1 - \mathbf{v}_2)\big| \\ &\le \|\mathbf{M}\|_\infty\|\mathbf{v}_1 - \mathbf{v}_2\|_1^2 + 2\|\mathbf{M}\mathbf{v}_2\|_\infty\|\mathbf{v}_1 - \mathbf{v}_2\|_1.\end{aligned}$$

□

By Lemma B.2, we have

$$\sigma_2^2 - \sigma_1^2 \le \|\mathbf{\Sigma}_1\|_\infty\|\boldsymbol{\beta}_j - \boldsymbol{\beta}_j^*\|_1^2 + 2\|\mathbf{\Sigma}_1\boldsymbol{\beta}_j^*\|_\infty\|\boldsymbol{\beta}_j - \boldsymbol{\beta}_j^*\|_1. \tag{B.3}$$

By Hölder's inequality and the relation between $\ell_1$-norm and $\ell_2$-norm of a vector, we have $\|\mathbf{\Sigma}_1\boldsymbol{\beta}_j^*\|_\infty \le \|\mathbf{\Sigma}_1\|_\infty\|\boldsymbol{\beta}_j^*\|_1 \le \sqrt{s^*}C_\beta D$. Therefore the right-hand side of (B.3) can be bounded by

$$\sigma_2^2 - \sigma_1^2 \le r^2 D + 2\sqrt{s^*}rC_\beta D,$$

which shows that $\sigma_2^2$ is also bounded because $r = \mathcal{O}(1/\sqrt{s^*})$. In addition, by the bound $1 - \Phi(x) \le \exp(-x^2/2)/(x\sqrt{2\pi})$ for the standard normal distribution function, we obtain that

$$\mathbb{P}\big(\mathcal{B}_i^c\big) \le \mathbb{P}\big(\boldsymbol{X}_{i\setminus j}^T\boldsymbol{\beta}_j^* > R\big) + \mathbb{P}\big(\boldsymbol{X}_{i\setminus j}^T\boldsymbol{\beta}_j > R\big) \le c\sigma_1\exp\big[-R^2/(2\sigma_1^2)\big]/R + c\sigma_2\exp\big[-R^2/(2\sigma_2^2)\big]/R,$$

where the constant $c = 1/\sqrt{2\pi}$. We denote the last expression as $C_3(R)$, then the right-hand side of (B.2) can be upper-bounded by $\sqrt{3\sigma_v^4}\sqrt{2C_3(R)} \le 2\sqrt{6C_3(R)}C_\lambda$. Hence we can choose a sufficiently large $R$ such that $2\sqrt{6C_3(R)}C_\lambda = \lambda_{\min}(\mathbf{\Sigma})$ and we denote this particular choice of $R$ as $R_0$.

Now we have

$$\mathbb{E}\Big\{\big[(\boldsymbol{X}_{i\setminus j} - \boldsymbol{X}_{i'\setminus j})^T\mathbf{v}\big]^2\big[1 - I(\mathcal{B}_i)I(\mathcal{B}_{i'})\big]\Big\} \le \lambda_{\min}(\mathbf{\Sigma})$$

Note that $\mathbb{E}\big\{[(\boldsymbol{X}_{i\setminus j} - \boldsymbol{X}_{i'\setminus j})^T\mathbf{v}]^2\big\} = \sigma_v^2 \ge 2\lambda_{\min}(\mathbf{\Sigma})$, we obtain that

$$\mathbf{v}^T\mathbb{E}\big[\nabla^2 L_j(\boldsymbol{\beta}_j)\big]\mathbf{v} \ge C_1(R_0,\tau)C_2(R_0,\tau)\lambda_{\min}(\mathbf{\Sigma}) \quad \text{for all} \quad \tau \in \mathbb{R}.$$

Therefore we conclude that for all $\boldsymbol{\beta}_j \in \mathbb{R}^{d-1}$ such that $\|\boldsymbol{\beta}_j - \boldsymbol{\beta}_j^*\|_1 \le r$,

$$\mathbf{v}^T\mathbb{E}\big[\nabla^2 L_j(\boldsymbol{\beta}_j)\big]\mathbf{v} \ge \max_{\tau\in\mathbb{R}}\big\{C_1(R_0,\tau)C_2(R_0,\tau)\big\}\lambda_{\min}(\mathbf{\Sigma}). \tag{B.4}$$

**Step (ii): Upper Bound of $\mathbf{v}^T\mathbb{E}\big[\nabla^2 L_j(\boldsymbol{\beta}_j)\big]\mathbf{v}$.** For any $\boldsymbol{\beta}_j \in \mathbb{R}^{d-1}$ such that $\|\boldsymbol{\beta}_j - \boldsymbol{\beta}_j^*\|_1 \le r$ and for any $\mathbf{v} \in \mathbb{R}^{d-1}$ with $\|\mathbf{v}\|_2 = 1$, by the definition of $\nabla^2 L_j(\boldsymbol{\beta}_j)$ we have

$$\mathbf{v}^T\nabla^2 L_j(\boldsymbol{\beta}_j)\mathbf{v} \le (X_{ij} - X_{i'j})^2\big[(\boldsymbol{X}_{i\setminus j} - \boldsymbol{X}_{i'\setminus j})^T\mathbf{v}\big]^2. \tag{B.5}$$

Notice that conditioning on $\boldsymbol{X}_{i\backslash j}$, $X_{ij} \sim N\big(\boldsymbol{X}_{i\backslash j}^T \boldsymbol{\beta}_j^*, \boldsymbol{\Theta}_{jj}^{-1}\big)$, hence

$$\mathbb{E}\big[(X_{ij} - X_{i'j})^2 \big| \boldsymbol{X}_{i\backslash j}, \boldsymbol{X}_{i'\backslash j}\big] = \big[(\boldsymbol{X}_{i\backslash j} - \boldsymbol{X}_{i'\backslash j})^T \boldsymbol{\beta}_j^*\big]^2 + 2\boldsymbol{\Theta}_{jj}^{-1}. \tag{B.6}$$

Combining (B.5) and (B.6) we obtain

$$\begin{aligned}\mathbb{E}\big[\mathbf{v}^T \nabla^2 L_j(\boldsymbol{\beta}_j)\mathbf{v}\big] &\le \mathbb{E}\Big\{\mathbb{E}\big[(X_{ij} - X_{i'j})^2 \big| \boldsymbol{X}_{i\backslash j}, \boldsymbol{X}_{i'\backslash j}\big] \cdot \big[(\boldsymbol{X}_{i\backslash j} - \boldsymbol{X}_{i'\backslash j})^T \mathbf{v}\big]^2\Big\} \\ &\le 2\boldsymbol{\Theta}_{jj}^{-1}\mathbb{E}\big((\boldsymbol{X}_{i\backslash j} - \boldsymbol{X}_{i'\backslash j})^T \mathbf{v}\big)^2 + \mathbb{E}\Big\{\big[(\boldsymbol{X}_{i\backslash j} - \boldsymbol{X}_{i'\backslash j})^T \boldsymbol{\beta}_j^*\big]^2 \big[(\boldsymbol{X}_{i\backslash j} - \boldsymbol{X}_{i'\backslash j})^T \mathbf{v}\big]^2\Big\}.\end{aligned} \tag{B.7}$$

Because $\boldsymbol{X}_{i\backslash j} \sim N(\mathbf{0}, \boldsymbol{\Sigma}_1)$ where $\boldsymbol{\Sigma}_1 := \boldsymbol{\Sigma}_{\backslash j, \backslash j}$, and also note that the maximum eigenvalue of $\boldsymbol{\Sigma}_1$ is upper bounded by $C_\lambda$, we have

$$\mathbb{E}\big[(\boldsymbol{X}_{i\backslash j} - \boldsymbol{X}_{i'\backslash j})^T \mathbf{v}\big]^2 = 2\mathbf{v}^T \boldsymbol{\Sigma}_1 \mathbf{v} \le 2C_\lambda.$$

Moreover, by inequality $2ab \le a^2 + b^2$ we obtain

$$2\mathbb{E}\Big\{\big[(\boldsymbol{X}_{i\backslash j} - \boldsymbol{X}_{i'\backslash j})^T \boldsymbol{\beta}_j^*\big]^2 \big[(\boldsymbol{X}_{i\backslash j} - \boldsymbol{X}_{i'\backslash j})^T \mathbf{v}\big]^2\Big\} \le \mathbb{E}\big[(\boldsymbol{X}_{i\backslash j} - \boldsymbol{X}_{i'\backslash j})^T \boldsymbol{\beta}_j^*\big]^4 + \mathbb{E}\big[(\boldsymbol{X}_{i\backslash j} - \boldsymbol{X}_{i'\backslash j})^T \mathbf{v}\big]^4.$$

Since $(\boldsymbol{X}_{i\backslash j} - \boldsymbol{X}_{i'\backslash j})^T \mathbf{v} \sim N(0, \sigma_v^2)$ and $(\boldsymbol{X}_{i\backslash j} - \boldsymbol{X}_{i'\backslash j})^T \boldsymbol{\beta}_j^* \sim N(0, 2\sigma_1^2)$ where $\sigma_v^2$ and $\sigma_1^2$ are defined as $2\mathbf{v}^T \boldsymbol{\Sigma}_1 \mathbf{v}$ and $\boldsymbol{\beta}_j^{*T} \boldsymbol{\Sigma}_1 \boldsymbol{\beta}_j^*$ respectively, we obtain

$$\mathbb{E}\big[(\boldsymbol{X}_{i\backslash j} - \boldsymbol{X}_{i'\backslash j})^T \boldsymbol{\beta}_j^*\big]^4 = 3\sigma_v^4 \le 12C_\lambda^2 \text{ and } \mathbb{E}\big[(\boldsymbol{X}_{i\backslash j} - \boldsymbol{X}_{i'\backslash j})^T \mathbf{v}\big]^4 = 12\sigma_1^4 \le 12C_\lambda C_\beta^2.$$

Therefore we can bound the right-hand side of (B.7) by

$$\mathbb{E}\big[\mathbf{v}^T \nabla^2 L_j(\boldsymbol{\beta}_j)\mathbf{v}\big] \le 4\boldsymbol{\Theta}_{jj}^{-1} C_\lambda + 6C_\lambda^2 + 6C_\lambda C_\beta^2. \tag{B.8}$$

Combining (B.4) and (B.8) we conclude that Proposition B.1 holds with

$$\rho_* = \max_{\tau \in \mathbb{R}} \big\{C_1(R_0, \tau) C_2(R_0, \tau)\big\} \lambda_{\min}(\boldsymbol{\Sigma}) \quad \text{and} \quad \rho^* = 4\boldsymbol{\Theta}_{jj}^{-1} C_\lambda + 6C_\lambda^2 + 6C_\lambda C_\beta^2.$$

□

## B.2 Refined Statistical Rates of Parameter Estimation

In this subsection we show more refined statistical rates of convergence for the proposed estimators. In specific, we consider the case where $\boldsymbol{\beta}_j^*$ contains nonzero elements with both strong and week magnitudes.

**Theorem B.3** (Refined statistical rates of convergence)**.** Under Assumptions 4.1 and 4.3, we let $K_1$ and $K_2$ be the constants defined in Theorem A.1 and also let $\rho_* > 0$ and $r > 0$ be defined in Assumption 4.3. For all $j \in [d]$, we define the support of $\boldsymbol{\beta}_j^*$ as $S_j := \{(j,k) \colon \beta_{jk}^* \ne 0, k \in [d]\}$ and let $s^* = \max_{j \in [d]} \|\boldsymbol{\beta}_j^*\|_0$. The penalty function $p_\lambda(u) \colon [0, +\infty) \to [0, +\infty)$ in (3.2) satisfies regularity conditions (C.1), (C.2) and (C.3) listed in §3.2 with $c_1 = 0.91$ and $c_2 \ge 24/\rho_*$ for condition (C.3). We set the regularity parameter $\lambda = C\sqrt{\log d/n}$ such that $C \ge 25K_1$. Moreover, we assume that the penalty function $p_\lambda(u)$ satisfies an extra condition (C.4): there exists a constant $c_3 > 0$ such that $p'_\lambda(u) = 0$ for $u \in [c_3\lambda, +\infty)$. Suppose that the support of $\boldsymbol{\beta}^*_j$ can be partitioned into $S_j = S_{1j} \cup S_{2j}$ where $S_{1j} = \{(j,k) \colon |\beta^*_{jk}| \geq (c_2 + c_3)\lambda\}$ and $S_{2j} = S_j - S_{1j}$. We denote constants $A_1 = 22\varrho$, $A_2 = 2.2c_2$, $B_1 = 32\varrho$, $B_2 = 3.2c_2$, $\varrho = c_2(c_2\rho_* - 11)^{-1}$, $\gamma = 11c_2^{-1}\rho_*^{-1} < 1$ and $a = 1.04$; we let $s^*_{1j} = |S_{1j}|$ and $s^*_{2j} = |S_{2j}|$. With probability at least $1 - d^{-1}$, we have the following more refined rates of convergence:

$$\big\|\widehat{\boldsymbol{\beta}}_j^{(\ell)} - \boldsymbol{\beta}^*_j\big\|_2 \leq A_1\Big\{\big\|\nabla_{S_{1j}} L_j(\boldsymbol{\beta}^*_j)\big\|_2 + a\sqrt{s^*_{2j}}\lambda\Big\} + A_2\sqrt{s^*}\lambda\gamma^\ell \quad \text{and} \tag{B.9}$$

$$\big\|\widehat{\boldsymbol{\beta}}_j^{(\ell)} - \boldsymbol{\beta}^*_j\big\|_1 \leq B_1\Big\{\big\|\nabla_{S_{1j}} L_j(\boldsymbol{\beta}^*_j)\big\|_2 + a\sqrt{s^*_{2j}}\lambda\Big\} + B_2 s^*\lambda\gamma^\ell, \forall j \in [d]. \tag{B.10}$$

*Proof of Theorem B.3.* Let $S_j = \{(j,k) \colon \beta^*_{jk} \neq 0, k \in [d]\}$ be the support of $\boldsymbol{\beta}^*_j$ and let index set $G^\ell_j$, $J^\ell_j$ and $I^\ell_j$ be the same as defined in the proof of Theorem 4.4. For notational simplicity, we omit the subscript $j$ in these index sets which stands for the $j$-th node of the graph; we simply write them as $G^\ell$, $J^\ell$ and $I^\ell$. Moreover, we let $\boldsymbol{\delta}^{(\ell)} = \widehat{\boldsymbol{\beta}}_j^{(\ell)} - \boldsymbol{\beta}^*_j$, it is shown in Lemma A.3 that

$$\big\|\boldsymbol{\delta}^{(\ell)}_{I^\ell}\big\|_2 \leq 10\rho_*^{-1}\Big(\big\|\nabla_{\widetilde{G}^\ell} L_j(\boldsymbol{\beta}^*_j)\big\|_2 + \big\|\boldsymbol{\lambda}^{(\ell-1)}_{S_j}\big\|_2\Big); \quad \widetilde{G}^\ell = (G^\ell)^c. \tag{B.11}$$

In the proof of Theorem 4.4, we show that $|\widetilde{G}^\ell| \leq 2s^*$ for all $j \in [d]$ and $\ell \geq 1$. Because $S_j = S_{1j} \cup S_{2j}$ where $S_{1j} = \{(j,k) \colon |\beta^*_{jk}| \geq (c_2 + c_3)\lambda\}$ and $S_{2j} = S_j - S_{1j}$, then by triangle inequality we have

$$\big\|\nabla_{S_j} L_j(\boldsymbol{\beta}^*_j)\big\|_2 \leq \big\|\nabla_{S_{1j}} L_j(\boldsymbol{\beta}^*_j)\big\|_2 + \sqrt{s^*_{2j}}\big\|\nabla_{S_{2j}} L_j(\boldsymbol{\beta}^*_j)\big\|_\infty.$$

Since $\lambda \geq 25\big\|\nabla L_j(\boldsymbol{\beta}^*_j)\big\|_\infty$ with high probability, by (A.7), we further have

$$\big\|\nabla_{\widetilde{G}^\ell} L_j(\boldsymbol{\beta}^*_j)\big\|_2 \leq \big\|\nabla_{S_{1j}} L_j(\boldsymbol{\beta}^*_j)\big\|_2 + \sqrt{s^*_{2j}}\lambda/25 + \big\|\boldsymbol{\delta}^{(\ell-1)}_{I^{\ell-1}}\big\|_2\Big/(25c_2). \tag{B.12}$$

Note that by the definition of $S_{1j}$, for any $(j,k) \in S_{1j}$, $p'_\lambda(|\beta_{jk}| - c_2\lambda) \leq p'_\lambda(c_3\lambda) = 0$, then we have

$$\Upsilon_j := \lambda\Big[\sum_{(j,k)\in S_j} p'_\lambda\big(|\beta^*_{jk}| - c_2\lambda\big)^2\Big]^{1/2} = \lambda\Big[\sum_{(j,k)\in S_{2j}} p'_\lambda\big(|\beta^*_{jk}| - c_2\lambda\big)^2\Big]^{1/2} \leq \sqrt{s^*_{2j}}\lambda.$$

Therefore (A.9) is reduced to

$$\big\|\boldsymbol{\lambda}^{(\ell-1)}_{S_j}\big\|_2 \leq \Upsilon_j + \big\|\boldsymbol{\delta}^{(\ell-1)}_{I^{\ell-1}}\big\|_2/c_2 \leq \sqrt{s^*_{2j}}\lambda + \big\|\boldsymbol{\delta}^{(\ell-1)}_{I^{\ell-1}}\big\|_2\Big/c_2. \tag{B.13}$$

Combining (B.11), (B.12) and (B.13) we obtain

$$\big\|\boldsymbol{\delta}^{(\ell)}_{I^\ell}\big\|_2 \leq 10\rho_*^{-1}\Big\{\big\|\nabla_{S_{1j}} L_j(\boldsymbol{\beta}^*_j)\big\|_2 + 1.04\sqrt{s^*_{2j}}\lambda + 1.04\big\|\boldsymbol{\delta}^{(\ell-1)}_{I^{\ell-1}}\big\|_2/c_2\Big\}.$$

Then by recursion, we obtain the following estimation error:

$$\big\|\boldsymbol{\delta}^{(\ell)}_{I^\ell}\big\|_2 \leq 10\varrho\Big\{\big\|\nabla_{S_{1j}} L_j(\boldsymbol{\beta}^*_j)\big\|_2 + 1.04\sqrt{s^*_{2j}}\lambda\Big\} + \gamma^{\ell-1}\big\|\boldsymbol{\delta}^{(1)}_{I^1}\big\|_2,$$

where $\gamma := 11c_2^{-1}\rho_*^{-1}$ and $\varrho := c_2(c_2\rho_* - 11)^{-1}$. Note that we assume $c_2 \geq 24\rho_*^{-1}$, for $k = 1$ by (A.11) we have

$$2.2\big\|\boldsymbol{\delta}^{(1)}_{I_1}\big\|_2 \leq 2.2c_2\gamma\sqrt{s^*}\lambda \quad \text{and} \quad 2.2\sqrt{2s^*}\big\|\boldsymbol{\delta}^{(1)}_{I_1}\big\|_2 \leq 3.2c_2\gamma s^*\lambda.$$

Therefore using the original notation, we can obtain the refined rates of convergence by (A.2)

$$\big\|\widehat{\boldsymbol{\beta}}_j^{(\ell)} - \boldsymbol{\beta}_j^*\big\|_2 \le 22\varrho\Big\{\big\|\nabla_{S_{1j}} L_j(\boldsymbol{\beta}_j^*)\big\|_2 + 1.04\sqrt{s_{2j}^*}\lambda\Big\} + 2.2c_2\gamma^\ell\sqrt{s^*}\lambda \ \text{ and}$$
$$\big\|\widehat{\boldsymbol{\beta}}_j^{(\ell)} - \boldsymbol{\beta}_j^*\big\|_1 \le 32\varrho\sqrt{s^*}\Big\{\big\|\nabla_{S_{1j}} L_j(\boldsymbol{\beta}_j^*)\big\|_2 + 1.04\sqrt{s_{2j}^*}\lambda\Big\} + 3.2c_2\gamma^\ell s^*\lambda,$$

where $s_{2j}^* = |S_{2j}|$. Moreover, it is easy to see that, with probability at least $1–d^{-1}$, these convergence rates hold for all $j \in [d]$. □

# C Proof of the Auxiliary Results for Estimation

In this appendix, we prove the main results for estimation results presented in §4.1. In this appendix, we prove the auxiliary results for estimation. In specific, we give detailed proofs of Lemmas A.1, A.2, and A.3, which are pivotal for the proof of Theorem4.4. We first prove Lemmas A.1, which gives an upper bound for $\big\|\nabla L_j(\boldsymbol{\beta}_j^*)\big\|_\infty$.

## C.1 Proof of Lemma A.1

*Proof of Lemma A.1.* By definition, $\nabla L_j(\boldsymbol{\beta}_j^*)$ is a centered second-order $U$-statistic with kernel function $\mathbf{h}_{ii'}^j(\boldsymbol{\beta}_j^*)\in\mathbb{R}^{d-1}$, whose entries are given by

$$\big[\mathbf{h}_{ii'}^j(\boldsymbol{\beta}_j^*)\big]_{jk} = \frac{R_{ii'}^j(\boldsymbol{\beta}_j^*)(X_{ij} - X_{i'j})(X_{ik} - X_{i'k})}{1 + R_{ii'}^j(\boldsymbol{\beta}_j^*)}.$$

By the tail probability bound in (4.1), for any $i \in [n]$ and $j \in [d]$, we have

$$\begin{aligned}\mathbb{P}\big(|X_{ij}| > x, \forall i \in [n], \forall j \in [d]\big) &\le \sum_{i\in[n],j\in[d]} \mathbb{P}(|X_{ij}| > x)\\ &\le 2\exp(\kappa_m + \kappa_h/2)\exp(-x + \log d + \log n).\end{aligned} \tag{C.1}$$

By setting $x = C\log d$ for some constant $C$, we conclude that event $\mathcal{E} := \{|X_{ij}| \le C\log d, \forall i \in [n], \forall j \in [d]\}$ holds with probability at least $1 - (4d)^{-1}$. Following from the same argument as in Ning and Liu (2014), it is easy to show that, conditioning on $\mathcal{E}$, $\mathbf{h}_{ii'}^j(\boldsymbol{\beta}_j^*)$ is also centered. Note that conditioning on event $\mathcal{E}$, $\big\|\mathbf{h}_{ii'}^j(\boldsymbol{\beta}_j^*)\big\|_\infty \le C\log^2 d$ for some generic constant $C$ and for all $i, i' \in [d]$ and $j \in [d]$. The following Bernstein's inequality for $U$-statistics, presented in Arcones (1995), gives an upper bound for the tail probability of $\nabla L_j(\boldsymbol{\beta}_j^*)$.

**Lemma C.1** (Bernstein's inequality for $U$-statistics)**.** Given i.i.d. random variables $Z_1, \dots Z_n$ taking values in a measurable space $(\mathbb{S}, \mathcal{B})$ and a symmetric and measurable kernel function $h\colon \mathbb{S}^m \to R$, we define the $U$-statistics with kernel $h$ as $U := \binom{n}{m}^{-1}\sum_{i_1<\dots<i_m} h(Z_{i_1}, \dots, Z_{i_m})$. Suppose that $\mathbb{E}h(Z_{i_1}, \dots, Z_{i_m}) = 0$, $\mathbb{E}\big\{\mathbb{E}[h(Z_{i_1}, \dots, Z_{i_m}) \mid Z_{i_1}]\big\}^2 = \sigma^2$ and $\|h\|_\infty \le b$. There exists an absolute constants $K(m) > 0$ depending on $m$ such that

$$\mathbb{P}(|U| > t) \le 4\exp\big\{-nt^2/[2m^2\sigma^2 + K(m)bt]\big\},\ \forall t > 0. \tag{C.2}$$

Note that by (4.1), the fourth moment of $\boldsymbol{X}$ is bounded, which implies that $\mathbb{E}[\mathbf{h}_{ii'}^{j}(\boldsymbol{\beta}_j^*)]^2$ is uniformly bounded by an absolute constant for all $j \in [d]$. By Lemma C.1, setting $b = C \log^2 d$ in (C.2) yields that

$$\mathbb{P}\big(\big|\nabla_{jk} L_j(\boldsymbol{\beta}_j^*)\big| > t \big| \mathcal{E}\big) \leq 4 \exp\Big[-nt^2/(C_1 + C_2 \log^2 d \cdot t)\Big] \tag{C.3}$$

for some generic constants $C_1$ and $C_2$. Taking a union bound over $\{(j,k)\colon j,k \in [d], k \neq j\}$ we obtain

$$\max_{j\in[d]} \Big\{ \mathbb{P}\big(\big\|\nabla L_j(\boldsymbol{\beta}_j^*)\big\|_\infty > t \big| \mathcal{E}\big) \Big\} \lesssim d^2 \cdot \exp\big[-nt^2/(C_1 + C_2 \log^2 d \cdot t)\big]. \tag{C.4}$$

Under Assumption 4.3 and conditioning on $\mathcal{E}$, by setting $t = K_1 \sqrt{\log d/n}$ for a sufficiently large $K_1 > 0$, it holds probability greater than $1 - (4d)^{-1}$ that

$$\big\|\nabla L_j(\boldsymbol{\beta}_j^*)\big\|_\infty \leq K_1 \sqrt{\log d/n} \quad \forall j \in [d].$$

Note that event $\mathcal{E}$ holds with probability at least $1 - (4d)^{-1}$, we conclude the proof of Lemma A.1. □

## C.2 Proof of Lemma A.2

*Proof of Lemma A.2.* In what follows, for notational simplicity and readability, we omit $j$ in the subscript and $\ell$ in the superscript by simply writing $S_j$, $G_j^\ell$, $J_j^\ell$ and $I_j^\ell$ as $S, G, J$ an $I$ respectively. By the definition of $G$, $\big\|\boldsymbol{\lambda}_{G^\ell}^{(\ell-1)}\big\|_{\min} \geq p'_\lambda(\theta) \geq 0.91\lambda > 22.75 \|\nabla L_j(\boldsymbol{\beta}_j^*)\|_\infty$. We prove this lemma in two steps. In the **first step** we show that $\big\|\widehat{\boldsymbol{\beta}}_j^{(\ell)} - \boldsymbol{\beta}_j^*\big\|_1 \leq 2.2 \big\|\widehat{\boldsymbol{\beta}}_{G^c}^\ell - \boldsymbol{\beta}_{G^c}^*\big\|_1$. Suppose that $\widehat{\boldsymbol{\beta}}_j^{(\ell)}$ is the solution in the $\ell$-th iteration and we denote $\nabla_{jk} L_j(\boldsymbol{\beta}_j) = \partial L_j(\boldsymbol{\beta}_j)/\partial \beta_{jk}$, the Karush-Kuhn-Tucker condition implies that

$$\begin{aligned}
&\nabla_{jk} L_j(\widehat{\boldsymbol{\beta}}_j^{(\ell)}) + \lambda_{jk}^{(\ell-1)} \mathrm{sign}(\widehat{\beta}_{jk}^{(\ell)}) = 0 \quad if \quad \widehat{\beta}_{jk}^{(\ell)} \neq 0; \\
&\nabla_{jk} L_j(\widehat{\boldsymbol{\beta}}_j^{(\ell)}) + \lambda_{jk}^{(\ell-1)} \xi_{jk}^{(\ell)} = 0, \quad \xi_{jk}^{(\ell)} \in [-1, 1] \quad if \quad \widehat{\beta}_{jk}^{(\ell)} = 0.
\end{aligned}$$

The above Karush-Kuhn-Tuker condition can be written in a compact form as

$$\nabla L_j(\widehat{\boldsymbol{\beta}}_j^{(\ell)}) + \boldsymbol{\lambda}_j^{(\ell-1)} \circ \boldsymbol{\xi}_j^{(\ell)} = 0, \tag{C.5}$$

where $\boldsymbol{\xi}_j^{(\ell)} \in \partial \big\|\widehat{\boldsymbol{\beta}}_j^{(\ell)}\big\|_1$and $\boldsymbol{\lambda}_j^{(\ell-1)} = \big(\lambda_{j1}^{(\ell-1)}, \ldots, \lambda_{jj-1}^{(\ell-1)}, \lambda_{jj+1}^{(\ell-1)}, \ldots, \lambda_{jd}^{(\ell-1)}\big)^T \in \mathbb{R}^{d-1}$.

For notational simplicity, we let $\boldsymbol{\delta} = \widehat{\boldsymbol{\beta}}_j^{(\ell)} - \boldsymbol{\beta}_j^* \in \mathbb{R}^{d-1}$ and omit the superscript $\ell$ and subscript $j$ in both $\boldsymbol{\lambda}_j^{(\ell-1)}$ and $\boldsymbol{\xi}_j^{(\ell)}$ by writing them as $\boldsymbol{\lambda}$ and $\boldsymbol{\xi}$. By definition, $I = G^c \cup J$. Note that we denote the support of $\boldsymbol{\beta}_j^*$ as $S$; we define $H := G^c - S$, then $S, H$ and $G$ is a partition of $\big\{(j,k)\colon k \in [d], k \neq j\big\}$.

By the Mean-Value theorem, there exists an $\alpha \in [0,1]$ such that $\widetilde{\boldsymbol{\beta}}_j := \alpha \boldsymbol{\beta}_j^* + (1-\alpha)\widehat{\boldsymbol{\beta}}_j^{(\ell)} \in \mathbb{R}^{d-1}$ satisfies

$$\nabla L_j(\widehat{\boldsymbol{\beta}}_j) - \nabla L_j(\boldsymbol{\beta}_j^*) = \nabla^2 L_j(\widetilde{\boldsymbol{\beta}}_j)\boldsymbol{\delta}.$$

Then (C.5) implies that

$$0 \le \boldsymbol{\delta}^T \nabla^2 L_j(\widetilde{\boldsymbol{\beta}}_j)\boldsymbol{\delta} = -\underbrace{\big\langle \boldsymbol{\delta}, \boldsymbol{\lambda} \circ \boldsymbol{\xi} \big\rangle}_{\text{(i)}} - \underbrace{\big\langle \nabla L_j(\boldsymbol{\beta}_j^*), \boldsymbol{\delta} \big\rangle}_{\text{(ii)}}. \tag{C.6}$$

For term (ii) in (C.6), Hölder's inequality implies that

$$\text{(ii)} \ge -\big\|\nabla L_j(\boldsymbol{\beta}_j^*)\big\|_\infty \|\boldsymbol{\delta}\|_1. \tag{C.7}$$

For term (i) in (C.6), recall that we denote $|\mathbf{v}|$ as the vector that takes entrywise absolute value for $\mathbf{v}$. By the fact that $\xi_{jk}^{(\ell)} \widehat{\beta}_{jk}^{(\ell)} = |\widehat{\beta}_{jk}^{(\ell)}|$, we have $\boldsymbol{\xi}_G \circ \boldsymbol{\delta}_G = |\boldsymbol{\delta}_G|$ and $\boldsymbol{\xi}_H \circ \boldsymbol{\delta}_H = |\boldsymbol{\delta}_H|$. Since $\boldsymbol{\delta}_{S^c} = \widehat{\boldsymbol{\beta}}_{S^c}^{(\ell)}$. Hölder's inequality implies that

$$\begin{aligned} \big\langle \boldsymbol{\delta}, \boldsymbol{\lambda} \circ \boldsymbol{\xi} \big\rangle &= \big\langle \boldsymbol{\delta}_S, (\boldsymbol{\lambda} \circ \boldsymbol{\xi})_S \big\rangle + \big\langle |\boldsymbol{\delta}_H|, \boldsymbol{\lambda}_H \big\rangle + \big\langle |\boldsymbol{\delta}_G|, \boldsymbol{\lambda}_G \big\rangle \\ &\ge -\|\boldsymbol{\delta}_S\|_1 \|\boldsymbol{\lambda}_S\|_\infty + \|\boldsymbol{\delta}_G\|_1 \|\boldsymbol{\lambda}_G\|_{\min} + \|\boldsymbol{\delta}_H\|_1 \|\boldsymbol{\lambda}_H\|_{\min}. \end{aligned} \tag{C.8}$$

Combining (C.6), (C.7) and (C.8) we have

$$-\|\boldsymbol{\delta}_S\|_1 \|\boldsymbol{\lambda}_S\|_\infty + \|\boldsymbol{\delta}_G\|_1 \|\boldsymbol{\lambda}_G\|_{\min} + \|\boldsymbol{\delta}_H\|_1 \|\boldsymbol{\lambda}_H\|_{\min} - \big\|\nabla L_j(\boldsymbol{\beta}_j^*)\big\|_\infty \|\boldsymbol{\delta}\|_1 \le 0. \tag{C.9}$$

By the definition of $G$, we have $\|\boldsymbol{\lambda}_G\|_{\min} \ge p'_\lambda(c_2\lambda) \ge 0.91\lambda$. Rearranging terms in (C.9) we have

$$p'_\lambda(c_2\lambda)\|\boldsymbol{\delta}_G\|_1 \le \|\boldsymbol{\delta}_G\|_1 \|\boldsymbol{\lambda}_G\|_{\min} \le \big\|\nabla L_j(\boldsymbol{\beta}_j^*)\big\|_\infty \|\boldsymbol{\delta}\|_1 + \|\boldsymbol{\delta}_S\|_1 \|\boldsymbol{\lambda}_S\|_\infty.$$

Using the decomposability of the $\ell_1$-norm, we have

$$\Big[p'_\lambda(c_2\lambda) - \big\|\nabla L_j(\boldsymbol{\beta}_j^*)\big\|_\infty\Big] \|\boldsymbol{\delta}_G\|_1 \le \Big[\|\boldsymbol{\lambda}_S\|_\infty + \|\nabla L_j(\boldsymbol{\beta}_j^*)\|_\infty\Big] \|\boldsymbol{\delta}_{G^c}\|_1 \tag{C.10}$$

Recall that $\lambda > 25\big\|\nabla L_j(\boldsymbol{\beta}_j^*)\big\|_\infty$ and $p'_\lambda(\theta) \ge 0.91\lambda$, (C.10) implies

$$\big\|\boldsymbol{\delta}_G\big\|_1 \le \frac{\lambda + \big\|\nabla L_j(\boldsymbol{\beta}_j^*)\big\|_\infty}{p'_\lambda(c_2\lambda) - \big\|\nabla L_j(\boldsymbol{\beta}_j^*)\big\|_\infty} \|\boldsymbol{\delta}_{G^c}\|_1 \le 1.2 \|\boldsymbol{\delta}_{G^c}\|_1, \tag{C.11}$$

where we use the fact that

$$\frac{\lambda + \big\|\nabla L_j(\boldsymbol{\beta}_j^*)\big\|_\infty}{p'_\lambda(c_2\lambda) - \big\|\nabla L_j(\boldsymbol{\beta}_j^*)\big\|_\infty} \le \frac{\lambda + 0.04\lambda}{0.91\lambda - 0.04\lambda} \le 1.2.$$

Going back to the original notation, (C.11) is equivalent to

$$\big\|\widehat{\boldsymbol{\beta}}_j^{(\ell)} - \boldsymbol{\beta}_j^*\big\|_1 \le 2.2 \big\|\widehat{\boldsymbol{\beta}}_{\widetilde{G}_j^\ell}^{(\ell)} - \boldsymbol{\beta}_{\widetilde{G}_j^\ell}^*\big\|_1.$$

Now we show in the **second step** that $\big\|\widehat{\boldsymbol{\beta}}_j^{(\ell)} - \boldsymbol{\beta}_j^*\big\|_2 \le 2.2 \big\|\widehat{\boldsymbol{\beta}}_{I_j^\ell}^{(\ell)} - \boldsymbol{\beta}_{I_j^\ell}^*\big\|_2$. Recall that $J$ is the largest $k^*$ components of $\widehat{\boldsymbol{\beta}}_G^{(\ell)}$ in absolute value where we omit the subscript $j$ and superscript $\ell$ in the sets $G_j^\ell, J_j^\ell$ and $I_j^\ell$. By the definition of $J$ we obtain that

$$\|\boldsymbol{\delta}_{I^c}\|_\infty \le \|\boldsymbol{\delta}_J\|_1 / k^* \le \|\boldsymbol{\delta}_G\|_1 / k^*, \quad \text{where} \;\; \boldsymbol{\delta} = \widehat{\boldsymbol{\beta}}_j^{(\ell)} - \boldsymbol{\beta}_j^*.$$

By inequality (C.11) and the fact that $G^c \subset I$, we further have

$$\|\boldsymbol{\delta}_{I^c}\|_\infty \leq 1.2/k^* \cdot \|\boldsymbol{\delta}_{G^c}\|_1 \leq 1.2/k^* \cdot \|\boldsymbol{\delta}_I\|_1. \tag{C.12}$$

Then by Hölder' inequality and (C.12) we obtain that

$$\|\boldsymbol{\delta}_{I^c}\|_2 \leq \big(\|\boldsymbol{\delta}_{I^c}\|_1 \|\boldsymbol{\delta}_{I^c}\|_\infty\big)^{1/2} \leq (1.2/k^*)^{1/2}\big(\|\boldsymbol{\delta}_I\|_1\|\boldsymbol{\delta}_{I^c}\|_1\big)^{1/2}. \tag{C.13}$$

By the definition of index sets $G$ and $I$, we have $I^c \subset G$ and $G^c \subset I$. Then by (C.11) and (C.13) we obtain

$$\|\boldsymbol{\delta}_{I^c}\|_2 \leq (1.2/k^*)^{1/2}\big(\|\boldsymbol{\delta}_{G^c}\|_1\|\boldsymbol{\delta}_G\|_1\big)^{1/2} \leq 1.2\|\boldsymbol{\delta}_{G^c}\|_1/\sqrt{k^*}.$$

By the norm inequality between $\ell_1$-norm and $\ell_2$-norm, we have

$$\|\boldsymbol{\delta}_{I^c}\|_2 \leq 1.2\|\boldsymbol{\delta}_{G^c}\|_1/\sqrt{k^*} \leq 1.2\sqrt{2s^*/k^*}\|\boldsymbol{\delta}_{G^c}\|_2 \leq 1.2\|\boldsymbol{\delta}_I\|_2,$$

where we use $k^* \geq 2s^*$ and the induction assumption that $|G| \leq 2s^*$. Then triangle inequality for $\ell_2$-norm yields that

$$\|\boldsymbol{\delta}\|_2 \leq \|\boldsymbol{\delta}_{I^c}\|_2 + \|\boldsymbol{\delta}_I\|_2 \leq 2.2\|\boldsymbol{\delta}_I\|_2. \tag{C.14}$$

Note that (C.11) and (C.14) are equivalent to

$$\big\|\widehat{\boldsymbol{\beta}}_j^{(\ell)} - \boldsymbol{\beta}_j^*\big\|_2 \leq 2.2\big\|\widehat{\boldsymbol{\beta}}_{I_j^\ell}^{(\ell)} - \boldsymbol{\beta}_{I_j^\ell}^*\big\|_2 \quad \text{and} \quad \big\|\widehat{\boldsymbol{\beta}}_j^{(\ell)} - \boldsymbol{\beta}_j^*\big\|_1 \leq 2.2\big\|\widehat{\boldsymbol{\beta}}_{\widetilde{G}_j^\ell}^{(\ell)} - \boldsymbol{\beta}_{\widetilde{G}_j^\ell}^*\big\|_1,$$

where $\widetilde{G}_j^\ell = \big(G_j^\ell\big)^c$, which concludes the proof. □

## C.3 Proof of Lemma A.3

*Proof of Lemma A.3.* We first show that $\widehat{\boldsymbol{\beta}}_j^{(\ell)}$ stays in the $\ell_1$-ball centered at $\boldsymbol{\beta}_j^*$ with radius $r = C_\rho s^*\sqrt{\log d/n}$, where $C_\rho \geq 33\rho_*^{-1}$. For notational simplicity, we denote $\boldsymbol{\delta} = \widehat{\boldsymbol{\beta}}_j^{(\ell)} - \widehat{\boldsymbol{\beta}}_j$ and write $S_j$, $G_j^\ell$, $J_j^\ell$ and $I_j^\ell$ as $S, G, J$ an $I$ respectively. We prove by contradiction. Suppose that $\|\boldsymbol{\delta}\|_1 > r$, then we define $\widetilde{\boldsymbol{\beta}}_j = \boldsymbol{\beta}_j^* + t(\widehat{\boldsymbol{\beta}}_j^{(\ell)} - \boldsymbol{\beta}_j^*) \in \mathbb{R}^{d-1}$ with $t \in [0,1]$ such that $\big\|\widetilde{\boldsymbol{\beta}}_j - \boldsymbol{\beta}_j^*\big\|_1 \leq r$. Letting $\widetilde{\boldsymbol{\delta}} := \widetilde{\boldsymbol{\beta}}_j - \boldsymbol{\beta}_j^*$, by (C.14) we obtain

$$\|\widetilde{\boldsymbol{\delta}}\|_2 = t\|\boldsymbol{\delta}\|_2 \leq 2.2t\|\boldsymbol{\delta}_I\|_2 = 2.2\|\widetilde{\boldsymbol{\delta}}_I\|_2. \tag{C.15}$$

Moreover, by Lemma (A.2) and the relation between $\ell_1$- and $\ell_2$-norms we have

$$\|\widetilde{\boldsymbol{\delta}}\|_1 = t\|\boldsymbol{\delta}\|_1 \leq 2.2t\|\boldsymbol{\delta}_{G^c}\|_1 \leq 2.2\sqrt{2s^*}\|\widetilde{\boldsymbol{\delta}}_I\|_2, \tag{C.16}$$

where we use the fact that $G^c \subset I$ and the induction assumption that $|G^c| \leq 2s^*$. By Mean-Value theorem, there exist a $\gamma \in [0,1]$ such that $\nabla L_j(\widetilde{\boldsymbol{\beta}}_j) - \nabla L_j(\boldsymbol{\beta}_j^*) = \nabla^2 L_j(\boldsymbol{\beta}_1)\widetilde{\boldsymbol{\delta}}$, where $\boldsymbol{\beta}_1 := \gamma\boldsymbol{\beta}_j^* + (1-\gamma)\widetilde{\boldsymbol{\beta}}_j \in \mathbb{R}^{d-1}$. In what follows we will derive an upper bound for $\|\widetilde{\boldsymbol{\delta}}_I\|_2$ from $\widetilde{\boldsymbol{\delta}}^T\nabla^2 L_j(\boldsymbol{\beta}_1)\widetilde{\boldsymbol{\delta}}$. Before doing that, we present two lemmas. The first one shows that the restricted correlation coefficients defined as follows are closely related to the sparse eigenvalues. This lemma also appear in Zhang (2010) and Zhang et al. (2013) for $\ell_2$-loss.

**Lemma C.2.** [Local sparse eigenvalues and restricted correlation coefficients] Let $m$ be a positive integer and $\mathbf{M}(\cdot)\colon \mathbb{R}^m \to \mathbb{S}^m$ be a mapping from $\mathbb{R}^m$ to the space of $m \times m$ symmetric matrices. We define the $s$-sparse eigenvalues of $\mathbf{M}(\cdot)$ over the $\ell_1$-ball centered at $\mathbf{u}_0 \in \mathbb{R}^m$ with radius $r$ as

$$\rho_+\big(\mathbf{M}, \mathbf{u}_0; s, r\big) = \sup_{\mathbf{v},\mathbf{u}\in\mathbb{R}^m} \big\{\mathbf{v}^T\mathbf{M}(\mathbf{u})\mathbf{v}\colon \|\mathbf{v}\|_0 \le s, \|\mathbf{v}\|_2 = 1, \|\mathbf{u}-\mathbf{u}_0\|_1 \le r\big\};$$
$$\rho_-\big(\mathbf{M}, \mathbf{u}_0; s, r\big) = \inf_{\mathbf{v},\mathbf{u}\in\mathbb{R}^m} \big\{\mathbf{v}^T\mathbf{M}(\mathbf{u})\mathbf{v}\colon \|\mathbf{v}\|_0 \le s, \|\mathbf{v}\|_2 = 1, \|\mathbf{u}-\mathbf{u}_0\|_1 \le r\big\}.$$

In addition, we define the restricted correlation coefficients of $\mathbf{M}$ over the $\ell_1$-ball centered at $\mathbf{u}_0$ with radius $r$ as

$$\pi\big(\mathbf{M}, \mathbf{u}_0; s, k, r\big) := \sup_{\mathbf{v},\mathbf{w},\mathbf{u}\in\mathbb{R}^m} \left\{ \frac{\mathbf{v}_I^T\mathbf{M}(\mathbf{u})\mathbf{w}_J\big\|\mathbf{v}_I\big\|_2}{\mathbf{v}_I^T\mathbf{M}(\mathbf{u})\mathbf{v}_I\big\|\mathbf{w}_J\big\|_\infty} : I\cap J=\emptyset, |I|\le s, |J|\le k, \big\|\mathbf{u}-\mathbf{u}_0\big\|_1 \le r \right\}.$$

Suppose that the local sparse eigenvalue $\rho_-\big(\mathbf{M}, \mathbf{u}_0; s+k, r\big) > 0$, then we have the following upper bound on the restricted correlation coefficient $\pi(\mathbf{M}, \mathbf{u}_0; s, k)$:

$$\pi\big(\mathbf{M}, \mathbf{u}_0; s, k, r\big) \le \frac{\sqrt{k}}{2}\sqrt{\rho_+\big(\mathbf{M}, \mathbf{u}_0; k, r\big)\big/\rho_-\big(\mathbf{M}, \mathbf{u}_0; s+k, r\big) - 1}.$$

*Proof of Lemma C.2.* See §E.1.1 for a detailed proof. □

We denote the restricted correlation coefficients of $\nabla^2 L_j(\cdot)$ over the $\ell_1$-ball centered at $\boldsymbol{\beta}_j^*$ with radius $r$ as $\pi_j(s_1, s_2) := \pi\big(\nabla^2 L_j, \boldsymbol{\beta}_j^*; s_1, s_2, r\big)$ and denote the $s$-sparse eigenvalues $\rho_-\big(\nabla^2 L_j, \boldsymbol{\beta}_j^*; s, r\big)$ and $\rho_+\big(\nabla^2 L_j, \boldsymbol{\beta}_j^*; s, r\big)$ as $\rho_{j-}(s)$ and $\rho_{j+}(s)$ respectively. Applying Lemma C.2 to $\pi_j(2s^*+k^*, k^*)$ we obtain

$$\pi_j(2s^*+k^*, k^*) \le k^{*1/2}/2 \cdot \sqrt{\rho_{j+}(k^*)/\rho_{j-}(2s^*+2k^*) - 1}. \tag{C.17}$$

By the law of large numbers, if the sample size $n$ is sufficiently large such that $\nabla^2 L_j$ is close to its expectation $\mathbb{E}\big[\nabla^2 L_j\big]$. When $\boldsymbol{\beta}_j$ is close to $\boldsymbol{\beta}_j^*$, by Assumption 4.3, we expect that the sparse eigenvalue condition also holds for $\nabla^2 L_j(\boldsymbol{\beta}_j)$ with high probability. The following lemma justifies this intuition.

**Lemma C.3.** Recall that we define the sparse eigenvalues of $\mathbb{E}\big[\nabla^2 L_j(\boldsymbol{\beta}_j^*)\big]$ in Definition 4.2. Under Assumptions 4.1 and 4.3, if $n$ is sufficiently large such that $\rho_* \gtrsim k^*\lambda\log^2 d$, with probability at least $1-(2d)^{-1}$, for all $j\in[d]$, there exists a constant $C_\rho \ge 33\rho_*^{-1}$ such that

$$\rho_{j-}^*(2s^*+2k^*) - 0.05\rho_* \le \rho_{j-}(2s^*+2k^*) < \rho_{j+}(k^*) \le \rho_{j+}^*(k^*) + 0.05\rho_*, \quad \text{and}$$
$$\rho_{j+}(k^*)\big/\rho_{j-}(2s^*+2k^*) \le 1 + 0.27k^*/s^*,$$

where we denote the local sparse eigenvalues $\rho_-\big(\nabla^2 L_j, \boldsymbol{\beta}_j^*; s, r\big)$ and $\rho_+\big(\nabla^2 L_j, \boldsymbol{\beta}_j^*; s, r\big)$ with $r = C_\rho\sqrt{\log d/n}$ as $\rho_{j-}(s)$ and $\rho_{j+}(s)$, respectively.

*Proof.* See §E.1.2 for a detailed proof. □

Thus by Lemma C.3 we have

$$\pi_j(2s^*{+}k^*, k^*) \le 0.5\sqrt{0.27{k^*}^2/s^*}. \tag{C.18}$$

By (C.11), (C.18) and $G^c \subset I$ we obtain

$$1 - 2\pi_j(2s^*{+}k^*, k^*){k^*}^{-1}\|\widetilde{\boldsymbol{\delta}}_G\|_1/\|\widetilde{\boldsymbol{\delta}}_I\|_2 \ge 1 - 1.2\sqrt{0.54} := \kappa_1, \tag{C.19}$$

where we denote $\kappa_1 := 1 - 1.2\sqrt{0.54} \ge 0.11$. Now we use the second lemma to get an lower bound of $\widetilde{\boldsymbol{\delta}}^T\nabla^2 L_j(\boldsymbol{\beta}_1)\widetilde{\boldsymbol{\delta}}$, which implies an upper bound for $\|\widetilde{\boldsymbol{\delta}}_I\|_2$.

**Lemma C.4.** Let $\mathbf{M}\colon \mathbb{R}^m \to \mathbb{S}^m$ be a mapping from $\mathbb{R}^m$ to the space of $m\times m$-symmetric matrices. Suppose that the sparse eigenvalue $\rho_-\big(\mathbf{M}, \mathbf{u}_0; s{+}k, r\big) > 0$, let the restricted correlation coefficients of $\mathbf{M}(\cdot)$ be defined in Lemma C.2. We denote the restricted correlation coefficients $\pi\big(\mathbf{M}, \mathbf{u}_0; s, k, r\big)$ and $s$-sparse eigenvalue $\rho_-\big(\mathbf{M}, \mathbf{u}_0; s, r\big)$ as $\pi(s,k)$ and $\rho_-(s)$ respectively for notational simplicity. For any $\mathbf{v} \in \mathbb{R}^d$, let $F$ be any index set such that $|F^c| \le s$, let $J$ be the set of indices of the largest $k$ entries of $\mathbf{v}_F$ in absolute value and let $I = F^c \cup J$. For any $\mathbf{u} \in \mathbb{R}^d$ such that $\|\mathbf{u} - \mathbf{u}_0\|_2 \le r$ and any $\mathbf{v} \in \mathbb{R}^d$ satisfying $1 - 2\pi(s{+}k, k)\|\mathbf{v}_F\|_1/\|\mathbf{v}_I\|_2 > 0$ we have

$$\mathbf{v}^T\mathbf{M}(\mathbf{u})\mathbf{v} \ge \rho_-(s{+}k)\big[\|\mathbf{v}_I\|_2 - 2\pi(s{+}k, k)\|\mathbf{v}_F\|_1/k\big]\|\mathbf{v}_I\|_2.$$

*Proof of Lemma C.4.* See §E.1.3 for a detailed proof. □

Now applying Lemma C.4 to $\nabla^2 L_j(\cdot)$ with $F{=}G$, $s{=}2s^*$ and $k{=}k^*$ we obtain

$$\widetilde{\boldsymbol{\delta}}^T\nabla^2 L_j(\boldsymbol{\beta}_1)\widetilde{\boldsymbol{\delta}} \ge \rho_{j-}(2s^*{+}k^*)\|\widetilde{\boldsymbol{\delta}}_I\|_2\big[\|\widetilde{\boldsymbol{\delta}}_I\|_2 - 2\pi_j(2s^*{+}k^*, k^*)/k^*\|\widetilde{\boldsymbol{\delta}}_G\|_1\big]. \tag{C.20}$$

Then by (C.19), the right-hand side of (C.20) can be lower bounded by

$$\widetilde{\boldsymbol{\delta}}^T\nabla^2 \ell(\boldsymbol{\beta}_1)\widetilde{\boldsymbol{\delta}} \ge \kappa_1\rho_{j-}(2s^*{+}k^*)\|\widetilde{\boldsymbol{\delta}}_I\|_2^2 \ge 0.95\kappa_1\rho_*\|\widetilde{\boldsymbol{\delta}}_I\|_2^2 = \kappa_2\rho_*\|\widetilde{\boldsymbol{\delta}}_I\|_2^2, \tag{C.21}$$

where we let $\kappa_2 := 0.95\kappa_1 \ge 0.1$. Now we derive an upper bound for $\widetilde{\boldsymbol{\delta}}^T\nabla^2 L_j(\boldsymbol{\beta}_1)\widetilde{\boldsymbol{\delta}}$. We define the symmetric Bregman divergence of $L_j(\boldsymbol{\beta}_j)$ as $D_j(\boldsymbol{\beta}_1, \boldsymbol{\beta}_2) := \big\langle \boldsymbol{\beta}_1 - \boldsymbol{\beta}_2, \nabla L_j(\boldsymbol{\beta}_1) - \nabla L_j(\boldsymbol{\beta}_2)\big\rangle$, where $\boldsymbol{\beta}_1, \boldsymbol{\beta}_2 \in \mathbb{R}^{d-1}$. Then by definition, $\widetilde{\boldsymbol{\delta}}^T\nabla^2 \ell(\boldsymbol{\beta}_1)\widetilde{\boldsymbol{\delta}} = D_j(\widetilde{\boldsymbol{\beta}}_j, \boldsymbol{\beta}_j^*)$. The following lemma relates $D_j(\widetilde{\boldsymbol{\beta}}_j, \boldsymbol{\beta}_j^*)$ with $D_j(\widehat{\boldsymbol{\beta}}_j, \boldsymbol{\beta}_j^*)$.

**Lemma C.5.** Let $D_j(\boldsymbol{\beta}_1, \boldsymbol{\beta}_2) := \big\langle \boldsymbol{\beta}_1 - \boldsymbol{\beta}_2, \nabla L_j(\boldsymbol{\beta}_1) - \nabla L(\boldsymbol{\beta}_2)\big\rangle$, $\boldsymbol{\beta}(t) = \boldsymbol{\beta}_1 + t(\boldsymbol{\beta}_2 - \boldsymbol{\beta}_1)$, $t \in (0,1)$ be any point on the line segment between $\boldsymbol{\beta}_1$ and $\boldsymbol{\beta}_2$. Then we have

$$D_j(\boldsymbol{\beta}(t), \boldsymbol{\beta}_1) \le tD_j(\boldsymbol{\beta}_2, \boldsymbol{\beta}_1)$$

*Proof of Lemma C.5.* See §E.1.4 for a detailed proof. □

By Lemma C.5 and (C.6),

$$D_j(\widetilde{\boldsymbol{\beta}}_j, \boldsymbol{\beta}_j^*) \le tD_j(\widehat{\boldsymbol{\beta}}_j, \boldsymbol{\beta}_j^*) \le \underbrace{-t\big\langle \nabla L_j(\boldsymbol{\beta}_j^*), \boldsymbol{\delta}\big\rangle}_{(i)}\underbrace{-t\big\langle \boldsymbol{\delta}, \boldsymbol{\lambda}_j \circ \boldsymbol{\xi}_j\big\rangle}_{(ii)}. \tag{C.22}$$

For term $(i)$ in (C.22), by Hölder's inequality we have

$$\begin{aligned}-t\big\langle\nabla L_j(\boldsymbol{\beta}_j^*),\boldsymbol{\delta}\big\rangle &\le t\big\|\nabla_{G^c}L_j(\boldsymbol{\beta}_j^*)\big\|_2\|\boldsymbol{\delta}_{G^c}\|_2+t\big\|\nabla_G L_j(\boldsymbol{\beta}_j^*)\big\|_\infty\|\boldsymbol{\delta}_G\|_1\\ &\le \big\|\nabla_{G^c}L_j(\boldsymbol{\beta}_j^*)\big\|_2\|\widetilde{\boldsymbol{\delta}}_I\|_2+\big\|\nabla_G L_j(\boldsymbol{\beta}_j^*)\big\|_\infty\|\widetilde{\boldsymbol{\delta}}_G\|_1,\end{aligned}\tag{C.23}$$

where the inequality follows from $G^c\subset I$. For term $(ii)$ in (C.22), by (C.8) and Hölder's inequality we have

$$-t\big\langle\boldsymbol{\delta},\boldsymbol{\lambda}_j\circ\boldsymbol{\xi}_j\big\rangle\le-\big\langle\boldsymbol{\delta}_S,(\boldsymbol{\lambda}_j\circ\boldsymbol{\xi}_j)_S\big\rangle-\big\langle|\widetilde{\boldsymbol{\delta}}_G|,\boldsymbol{\lambda}_G\big\rangle\le\|\boldsymbol{\lambda}_S\|_2\|\widetilde{\boldsymbol{\delta}}_I\|_2-p'_\lambda(c_2\lambda)\|\widetilde{\boldsymbol{\delta}}_G\|_1,\tag{C.24}$$

where we use the Hölder's inequality and the definition of $G$. Combining (C.21),(C.23) and (C.24) we obtain that

$$\begin{aligned}\kappa_2\rho_*\big\|\widetilde{\boldsymbol{\delta}}_I\big\|_2^2&\le\big(\big\|\nabla_{G^c}L_j(\boldsymbol{\beta}_j^*)\big\|_2+\|\boldsymbol{\lambda}_S\|_2\big)\|\widetilde{\boldsymbol{\delta}}_I\|_2+\big[\big\|\nabla L_j(\boldsymbol{\beta}_j^*)\big\|_\infty-p'_\lambda(c_2\lambda)\big]\|\widetilde{\boldsymbol{\delta}}_G\|_1\\&\le\big(\big\|\nabla_{G^c}L_j(\boldsymbol{\beta}_j^*)\big\|_2+\|\boldsymbol{\lambda}_S\|_2\big)\big\|\widetilde{\boldsymbol{\delta}}_I\big\|_2,\end{aligned}$$

where the second inequality follows from $p'_\lambda(c_2\lambda)>\big\|\nabla L_j(\boldsymbol{\beta}_j^*)\big\|_\infty$. From the inequality above and the induction assumption $|G^c|\le 2s^*$ we obtain that

$$\big\|\widetilde{\boldsymbol{\delta}}_I\big\|_2\le 10\rho_*^{-1}\big(\big\|\nabla_{G^c}L_j(\boldsymbol{\beta}_j^*)\big\|_2+\|\boldsymbol{\lambda}_S\|_2\big)\le 10\rho_*^{-1}\sqrt{s^*}\big(\sqrt{2}\big\|\nabla L_j(\boldsymbol{\beta}_j^*)\big\|_\infty+\lambda\big).\tag{C.25}$$

Thus (C.16), (C.25) and the the fact that $25\big\|\nabla L_j(\boldsymbol{\beta}_j^*)\big\|_\infty\le\lambda$ imply that

$$\|\widetilde{\boldsymbol{\delta}}\|_1\le 22\sqrt{2}\rho_*^{-1}(1+\sqrt{2}/25)s^*\lambda<33\rho_*^{-1}s^*\lambda\le r,\tag{C.26}$$

where the last inequality follows from the definition of $\lambda$. Notice that (C.26) contradicts our assumption that $\|\widetilde{\boldsymbol{\delta}}\|_1=r$, the reason for this contradiction is because we assume that $\big\|\widehat{\boldsymbol{\beta}}_j^{(\ell)}-\boldsymbol{\beta}_j^*\big\|_1>r$, hence $\big\|\widehat{\boldsymbol{\beta}}_j^{(\ell)}-\boldsymbol{\beta}_j^*\big\|_1\le r$ and $\widetilde{\boldsymbol{\beta}}_j=\widehat{\boldsymbol{\beta}}_j^{(\ell)}$. This means that $\widehat{\boldsymbol{\beta}}_j^{(\ell)}$ stays in the $\ell_1$-ball centered at $\boldsymbol{\beta}_j^*$ with radius $r$ in each iteration.

Moreover, by (C.14) and (C.25), we obtain the following upper bound for $\|\boldsymbol{\delta}_I\|_2$:

$$\|\boldsymbol{\delta}\|_2\le 22\rho_*^{-1}\big(\big\|\nabla_{G^c}L_j(\boldsymbol{\beta}_j^*)\big\|_2+\|\boldsymbol{\lambda}_S\|_2\big)\le 24\rho_*^{-1}\sqrt{s^*}\lambda,$$

where we use the condition that $\lambda\ge 25\big\|\nabla L_j(\boldsymbol{\beta}_j^*)\big\|_\infty$. In addition, by(C.11) and (C.25) we obtain the following bound on $\|\boldsymbol{\delta}\|_1$

$$\|\boldsymbol{\delta}\|_1\le 2.2\|\boldsymbol{\delta}_{G^c}\|_1\le 22\sqrt{2s^*}\rho_*^{-1}\Big(\big\|\nabla_{G^c}L_j(\boldsymbol{\beta}_j^*)\big\|_2+\|\boldsymbol{\lambda}_S\|_2\Big)\le 33\rho_*^{-1}s^*\lambda,\tag{C.27}$$

Therefore going back to the original notations, note that $\kappa_2\ge 0.1$, we establish the following crude rates of convergence for $\ell\ge 1$:

$$\big\|\widehat{\boldsymbol{\beta}}_j^{(\ell)}-\boldsymbol{\beta}_j^*\big\|_2\le 24\rho_*^{-1}\sqrt{s^*}\lambda\ \text{ and }\ \big\|\widehat{\boldsymbol{\beta}}_j^{(\ell)}-\boldsymbol{\beta}_j^*\big\|_1\le 33\rho_*^{-1}s^*\lambda.\tag{C.28}$$

And (C.25) is equivalent to

$$\big\|\widehat{\boldsymbol{\beta}}_{I_j^\ell}^{(\ell)}-\boldsymbol{\beta}_{I_j^\ell}^*\big\|_2\le 10\rho_*^{-1}\Big(\big\|\nabla_{\widetilde{G}_j^\ell}L_j(\boldsymbol{\beta}_j^*)\big\|_2+\big\|\boldsymbol{\lambda}_{S_j}^{(\ell-1)}\big\|_2\Big),\quad \widetilde{G}_j^\ell:=(G_j^\ell)^c.\tag{C.29}$$

Note that we use Lemmas A.1 and C.3, hence (C.29) and (C.29) hold with probability at least $1-d^{-1}$ for all $j\in[d]$. □

# D Proof of Auxiliary Results for Asymptotic Inference

We prove the auxiliary results for asymptotic inference. More specifically, we first prove Lemma A.5, which is pivotal for deriving the limiting distribution of the pairwise score statistic. Then we prove the lemmas presented in the proof of Theorems 4.7 and ??.

## D.1 Proof of Lemma A.5

*Proof of Lemma A.5.* Before proving this lemma, we first let $\nabla^2 L_{jk}(\boldsymbol{\beta}_{j\vee k})$ be the Hessian of $L_{jk}(\boldsymbol{\beta}_{j\vee k})$ and define $\mathbf{H}^{jk} := \mathbb{E}\big[\nabla^2 L_{jk}(\boldsymbol{\beta}^*_{j\vee k})\big]$. We also define

$$\boldsymbol{\Sigma}^{jk} := \mathbb{E}\big[\mathbf{g}_{jk}(\boldsymbol{X}_i)\mathbf{g}_{jk}(\boldsymbol{X}_i)^T\big] \quad \text{and} \quad \boldsymbol{\Theta}^{jk} := \mathbb{E}\big[\mathbf{h}^{jk}_{ii'}(\boldsymbol{\beta}^*)\mathbf{h}^{jk}_{ii'}(\boldsymbol{\beta}^*)^T\big].$$

Under assumption4.1, we first show that there exist a positive constant $D$ such that for any $j,k \in d$, $j \neq k$, $\max\Big\{\big\|\boldsymbol{\Sigma}^{jk}\big\|_\infty, \big\|\mathbf{H}^{jk}\big\|_\infty, \big\|\boldsymbol{\Theta}^{jk}\big\|_\infty\Big\} \leq D$. The reason is as follows.

Note that Hölder's inequality imply

$$\big\|\mathbf{H}^{jk}\big\|_\infty \lesssim \max_{j\in[d]} \mathbb{E}|X_{ij} - X_{i'j}|^4 \lesssim \max_{j\in[d]} \mathbb{E}|X_j|^4 \quad \text{for any } j,k \in [d], j \neq k.$$

Similarly, for $\boldsymbol{\Theta}^{jk}$, we also have $\big\|\boldsymbol{\Theta}^{jk}\big\|_\infty \lesssim \max_{j\in[d]} \mathbb{E}|X_j|^4$. By (4.1) we have

$$\mathbb{E}|X_j|^4 = \int_0^\infty \mathbb{P}(|X_4|^4 > t)dt \leq \int_0^\infty c\exp(-t^{1/4})dt = 24c, \quad c = 2\exp(\kappa_m + \kappa_h/2).$$

Moreover, note that by the law of total variance, the diagonal elements of $\boldsymbol{\Sigma}^{jk}$ are no larger than the corresponding diagonal elements of $\boldsymbol{\Theta}^{jk}$; then by Cauchy-Schwarz inequality, $\|\boldsymbol{\Sigma}^{jk}\|_\infty \leq \|\boldsymbol{\Theta}^{jk}\|_\infty$. Therefore there exist a constant $D$ that does not depend on $(s^*, n, d)$ such that

$$\max\big\{\|\mathbf{H}^{jk}\|_\infty, \|\boldsymbol{\Sigma}^{jk}\|_\infty, \|\boldsymbol{\Theta}^{jk}\|_\infty\big\} \leq D, \quad 1 \leq j < k \leq d. \tag{D.1}$$

Now we are ready to prove the lemma. Recall that $\nabla L_{jk}(\boldsymbol{\beta}_{j\vee k})$ is a $U$-statistic with kernel function $\mathbf{h}^{jk}_{ii'}(\boldsymbol{\beta}_{j\vee k})$. Because $\mathbf{h}^{jk}_{ii'}(\boldsymbol{\beta}^*_{j\vee k})$ is centered, the law of total expectation implies that $\mathbb{E}\big[\mathbf{g}_{jk}(\boldsymbol{X}_i)\big] = \mathbf{0}$. Note that the left-hand side of (A.19) can be written as

$$\begin{aligned}\frac{\sqrt{n}}{2}\mathbf{b}^T\nabla L_{jk}(\boldsymbol{\beta}^*_{j\vee k}) &= \frac{\sqrt{n}}{2}\mathbf{b}^T\mathbf{U}_{jk} + \frac{\sqrt{n}}{2}\mathbf{b}^T\left[\nabla L_{jk}(\boldsymbol{\beta}^*_{j\vee k}) - \mathbf{U}_{jk}\right] \\ &= \underbrace{\frac{1}{\sqrt{n}}\sum_{i=1}^n \boldsymbol{b}^T\mathbf{g}_{jk}(\boldsymbol{X}_i)}_{I_1} + \underbrace{\frac{\sqrt{n}}{2}\mathbf{b}^T\left[\nabla L_{jk}(\boldsymbol{\beta}^*_{j\vee k}) - \mathbf{U}_{jk}\right]}_{I_2}.\end{aligned}$$

Notice that $I_1$ is a weighted sum of i.i.d. random variables with the mean and variance given by

$$\mathbb{E}\big[\mathbf{b}^T\mathbf{g}_{jk}(\boldsymbol{X}_i)\big] = \mathbf{0} \quad \text{and} \quad \mathrm{Var}\big[\mathbf{b}^T\mathbf{g}_{jk}(\boldsymbol{X}_i)\big] = \mathbf{b}^T\boldsymbol{\Sigma}^{jk}\mathbf{b}.$$

Central limit theorem implies that $I_1 \rightsquigarrow N(0, \mathbf{b}^T\boldsymbol{\Sigma}^{jk}\mathbf{b})$. In what follows we use $\mathbf{h}_{ii'}$ and $\mathbf{h}_{ii'|i}$ to denote $\mathbf{h}^{jk}_{ii'}(\boldsymbol{\beta}^*_{j\vee k})$ and $\mathbb{E}\big[\mathbf{h}^{jk}_{ii'}(\boldsymbol{\beta}^*_{j\vee k})\big|\boldsymbol{X}_i\big] = \mathbf{g}_{jk}(\boldsymbol{X}_i)$. Thus we can write $I_2$ as

$$I_2 = \frac{1}{\sqrt{n}(n-1)}\sum_{i<i'}\mathbf{b}^T\boldsymbol{\chi}_{ii'}, \quad \text{where } \boldsymbol{\chi}_{ii'} = (\mathbf{h}_{ii'} - \mathbf{h}_{ii'|i} - \mathbf{h}_{ii'|i'}).$$

Then $\mathbb{E}(I_2^2)$ can be expanded as

$$\mathbb{E}(I_2^2) = \frac{1}{n(n-1)^2} \sum_{i<i',s<s'} \mathbf{b}^T \mathbb{E}(\boldsymbol{\chi}_{ii'} \boldsymbol{\chi}_{ss'}^T)\mathbf{b}. \tag{D.2}$$

By the definition of $\boldsymbol{\chi}_{ii'}$, we have

$$\begin{aligned} &\mathbb{E}(\boldsymbol{\chi}_{ii'}\boldsymbol{\chi}_{ss'}^T) = \mathbb{E}(\mathbf{h}_{ii'}\mathbf{h}_{ss'}^T) - \mathbb{E}(\mathbf{h}_{ii'}\mathbf{h}_{ss'|s}^T) - \mathbb{E}(\mathbf{h}_{ii'}\mathbf{h}_{ss'|s'}^T) - \mathbb{E}(\mathbf{h}_{ii'|i}\mathbf{h}_{ss'}^T) \\ &+ \mathbb{E}(\mathbf{h}_{ii'|i}\mathbf{h}_{ss'|s}^T) + \mathbb{E}(\mathbf{h}_{ii'|i}\mathbf{h}_{ss'|s'}^T) - \mathbb{E}(\mathbf{h}_{ii'|i'}\mathbf{h}_{ss'}^T) + \mathbb{E}(\mathbf{h}_{ii'|i'}\mathbf{h}_{ss'|s}^T) + \mathbb{E}(\mathbf{h}_{ii'|i'}\mathbf{h}_{ss'|s'}^T). \end{aligned} \tag{D.3}$$

Therefore, for $i \neq s, s'$ and $i' \neq s, s'$, law of total expectation implies that $\mathbb{E}(\boldsymbol{\chi}_{ii'}\boldsymbol{\chi}_{ss'}^T) = \mathbf{0}$. Similarly, if exactly one of $i, i'$ is identical to one of $s, s'$, say $i = s$, then (D.3) becomes

$$\mathbb{E}(\boldsymbol{\chi}_{ii'}\boldsymbol{\chi}_{ii''}^T) = \mathbb{E}(\mathbf{h}_{ii'}\mathbf{h}_{ii''}^T) - \mathbb{E}(\mathbf{h}_{ii'}\mathbf{h}_{ii''|i}^T) - \mathbb{E}(\mathbf{h}_{ii'|i}\mathbf{h}_{ii''}^T) + \mathbb{E}(\mathbf{h}_{ii'|i}\mathbf{h}_{ii''|i}^T), \;\; i \neq i' \neq i''.$$

Note that by the law of total expectation, for each term in (D.3) we have

$$\mathbb{E}(\mathbf{h}_{ii'}\mathbf{h}_{ii''}^T) = \mathbb{E}(\mathbf{h}_{ii'}\mathbf{h}_{ii''|i}^T) = \mathbb{E}(\mathbf{h}_{ii'|i}\mathbf{h}_{ii''}^T) = \mathbb{E}(\mathbf{h}_{ii'|i}\mathbf{h}_{ii''|i}^T).$$

Therefore, $\mathbb{E}(\boldsymbol{\chi}_{ii'}\boldsymbol{\chi}_{ii''}^T) = \mathbf{0}$. Finally, if $i = s$ and $i' = s'$, by the law of total expectation, (D.3) can be further reduced to $\mathbb{E}(\boldsymbol{\chi}_{ii'}\boldsymbol{\chi}_{ii'}^T) = \mathbb{E}(\mathbf{h}_{ii'}\mathbf{h}_{ii'}^T) - \mathbb{E}(\mathbf{h}_{ii'|i}\mathbf{h}_{ii'|i}^T) - \mathbb{E}(\mathbf{h}_{ii'|i'}\mathbf{h}_{ii'|i'}^T) = \boldsymbol{\Theta}^{jk} - 2\boldsymbol{\Sigma}^{jk}$. Thus by triangle inequality we have

$$\big\|\mathbb{E}(\boldsymbol{\chi}_{ii'}\boldsymbol{\chi}_{ii'}^T)\big\|_\infty \leq \big\|\mathbb{E}(\mathbf{h}_{ii'}\mathbf{h}_{ii'}^T)\big\|_\infty + \big\|\mathbb{E}(\mathbf{h}_{ii'|i}\mathbf{h}_{ii'|i}^T)\big\|_\infty + \big\|\mathbb{E}(\mathbf{h}_{ii'|j}\mathbf{h}_{ii'|j}^T)\big\|_\infty \leq 3D,$$

where the last inequality follows from Assumption 4.5. Then equation (D.2) can be reduced to

$$\mathbb{E}(I_2^2) = \frac{1}{n(n-1)^2} \sum_{i<i',s<s'} \mathbf{b}^T \mathbb{E}(\boldsymbol{\chi}_{ii'}\boldsymbol{\chi}_{ss}^T)\mathbf{b} = \frac{1}{n(n-1)^2} \sum_{i<i'} \mathbf{b}^T \mathbb{E}\big(\boldsymbol{\chi}_{ii'}\boldsymbol{\chi}_{ii'}^T\big)\mathbf{b}.$$

By Hölder's inequality we obtain

$$\mathbb{E}(I_2^2) \leq \frac{1}{2(n-1)} \|\mathbf{b}\|_1 \big\|\mathbb{E}(\boldsymbol{\chi}_{ii'}\boldsymbol{\chi}_{ii'}^T)\mathbf{b}\big\|_\infty \leq \frac{1}{2(n-1)} \|\mathbf{b}\|_1^2 \big\|\mathbb{E}(\boldsymbol{\chi}_{ii'}\boldsymbol{\chi}_{ii'}^T)\big\|_\infty \leq \frac{3D}{2(n-1)} \|\mathbf{b}\|_1^2. \tag{D.4}$$

Since $\|\mathbf{b}\|_0 \leq \widetilde{s}$, by the relationship between $\ell_1$-norm and $\ell_2$-norm, we can further bound the right-hand side of (D.4) by $\mathbb{E}(I_2^2) \leq 1.5\widetilde{s}D/(n-1) \to 0$, where we use the condition that $\lim_{n\to\infty} \widetilde{s}/n = 0$. □

## D.2 Proof of Lemma A.4

*Proof of Lemma A.4.* By the definition of $\mathbf{w}_{j,k}^*$ we have $\mathbf{H}_{jk,j\backslash k}^j = \mathbf{w}_{j,k}^{*T}\mathbf{H}_{j\backslash k,j\backslash k}^j$. We let $\widehat{\boldsymbol{\beta}}_j' = (0, \widehat{\boldsymbol{\beta}}_{j\backslash k})$ and denote $\nabla^2 L_j(\widehat{\boldsymbol{\beta}}_j')$ and $\nabla^2 L_j(\boldsymbol{\beta}_j^*)$ as $\boldsymbol{\Lambda}$ and $\boldsymbol{\Lambda}^*$ respectively. In addition, we write $\mathbf{H}^j, \mathbf{w}_{j,k}^*$ and $\widehat{\mathbf{w}}_{j,k}$ as $\mathbf{H}, \mathbf{w}^*$ and $\widehat{\mathbf{w}}$ respectively for notational simplicity. Triangle inequality implies that

$$\|\boldsymbol{\Lambda}_{jk,j\backslash k} - \mathbf{w}^{*T}\boldsymbol{\Lambda}_{j\backslash k,j\backslash k}\|_\infty \leq \|\mathbf{H}_{jk,j\backslash k} - \boldsymbol{\Lambda}_{jk,j\backslash k}\|_\infty + \|\mathbf{w}^{*T}(\mathbf{H}_{j\backslash k,j\backslash k} - \boldsymbol{\Lambda}_{j\backslash k,j\backslash k})\|_\infty.$$

Hölder's inequality implies that

$$\|\mathbf{\Lambda}_{jk,j\backslash k} - \mathbf{w}^{*T}\mathbf{\Lambda}_{j\backslash k,j\backslash k}\|_\infty \le \|\mathbf{\Lambda} - \mathbf{H}\|_\infty(1 + \|\mathbf{w}^*\|_1). \tag{D.5}$$

Under null hypothesis, $\beta^*_{jk} = 0$. By Lemma E.1, we have $\|\mathbf{\Lambda} - \mathbf{H}\|_\infty \lesssim s^*\lambda \log^2 d$. Then the right-hand side of (D.5) is bounded by

$$\|\mathbf{\Lambda}_{jk,j\backslash k} - \mathbf{w}^{*T}\mathbf{\Lambda}_{j\backslash k,j\backslash k}\|_\infty \lesssim (w_0 + 1)s^*\lambda \log^2 d.$$

Therefore, by the assumption that $\lambda_D \gtrsim \max\{1, w_0\}s^*\lambda \log^2 d$ we can ensure that $\mathbf{w}^*$ is in the feasible region of the Dantzig selector problem (3.5), hence we have $\|\widehat{\mathbf{w}}\|_1 \le \|\mathbf{w}^*\|_1 \le w_0$ by the optimality of $\widehat{\mathbf{w}}$. Let $J$ be the support set of $\mathbf{w}^*$, that is, $J := \{(j,\ell)\colon [\mathbf{w}^*_{j,k}]_{j\ell} \neq 0, \ell \in [d], \ell \neq j\}$; the optimality of $\mathbf{w}^*$ is equivalent to $\|\widehat{\mathbf{w}}_{J^c}\|_1 + \|\widehat{\mathbf{w}}_J\|_1 \le \|\mathbf{w}^*_J\|_1$. By triangle inequality, we have

$$\|\widehat{\mathbf{w}}_{J^c} - \mathbf{w}^*_{J^c}\|_1 = \|\widehat{\mathbf{w}}_{J^c}\|_1 \le \|\mathbf{w}^*_J\|_1 - \|\widehat{\mathbf{w}}_J\|_1 \le \|\widehat{\mathbf{w}}_J - \mathbf{w}^*_J\|_1, \tag{D.6}$$

where $J^c := \{(j,\ell)\colon (j,\ell) \notin J, j \text{ fixed}\}$. Letting $\widehat{\boldsymbol{\omega}} = \widehat{\mathbf{w}} - \mathbf{w}^*$, inequality (D.6) is equivalent to $\|\widehat{\boldsymbol{\omega}}_{J^c}\|_1 \le \|\widehat{\boldsymbol{\omega}}_J\|_1$. Moreover, triangle inequality yields that

$$\|\mathbf{\Lambda}_{j\backslash k,j\backslash k}\widehat{\boldsymbol{\omega}}\|_\infty \le \|\mathbf{\Lambda}_{jk,j\backslash k} - \mathbf{\Lambda}_{j\backslash k,j\backslash k}\widehat{\mathbf{w}}\|_\infty + \|\mathbf{\Lambda}_{jk,j\backslash k} - \mathbf{\Lambda}_{j\backslash k,j\backslash k}\mathbf{w}^*\|_\infty \le 2\lambda_D,$$

where the last inequality follows from that both $\mathbf{w}^*$ and $\widehat{\mathbf{w}}$ are feasible for the Dantzig selector problem (3.5). Then triangle inequality implies that

$$|\widehat{\boldsymbol{\omega}}^T\mathbf{\Lambda}_{j\backslash k,j\backslash k}\widehat{\boldsymbol{\omega}}| \le \underbrace{|\widehat{\boldsymbol{\omega}}_J^T\mathbf{\Lambda}_{J,j\backslash k}\widehat{\boldsymbol{\omega}}|}_{A_1} + \underbrace{|\boldsymbol{\omega}_{J^c}^T\mathbf{\Lambda}_{J^c,j\backslash k}\widehat{\boldsymbol{\omega}}|}_{A_2}.$$

By Hölder's inequality and inequality between $\ell_1$-norm and $\ell_2$-norms, we obtain that

$$A_1 \le 2\lambda_D\|\widehat{\boldsymbol{\omega}}_J\|_1 \le 2\sqrt{s_0^\star}\lambda_D\|\widehat{\boldsymbol{\omega}}_J\|_2 \quad \text{and} \quad A_2 \le 2\lambda_D\|\widehat{\boldsymbol{\omega}}_{J^c}\|_1 \le 2\lambda_D\|\widehat{\boldsymbol{\omega}}_J\|_1 \le 2\sqrt{s_0^\star}\lambda_D\|\widehat{\boldsymbol{\omega}}_J\|_2.$$

Hence we conclude that $|\widehat{\boldsymbol{\omega}}^T\mathbf{\Lambda}_{j\backslash k,j\backslash k}\widehat{\boldsymbol{\omega}}| \le 4\sqrt{s_0^\star}\lambda_D\|\widehat{\boldsymbol{\omega}}_J\|_2$.

We let $J_1$ be the set of indices of the largest $k_0^\star$ component of $\widehat{\boldsymbol{\omega}}_{J^c}$ in absolute value and let $I = J_1 \cup J$, then $|I| \le s_0^\star + k_0^\star$. Under the null hypothesis, $\|\widehat{\boldsymbol{\beta}}'_j - \boldsymbol{\beta}^*_j\|_1 = \|\widehat{\boldsymbol{\beta}}_{j\backslash k} - \boldsymbol{\beta}^*_{j\backslash k}\|_1 \le 33\rho_*^{-1}s^*\lambda$. We denote the $s$-sparse eigenvalue of $\nabla^2_{j\backslash k,j\backslash k}L_j(\boldsymbol{\beta}_j)$ over the $\ell_1$-ball centered at $\boldsymbol{\beta}^*_j$ with radius $r$ as $\rho'_{j+}(s)$ and $\rho'_{j-}(s)$ respectively and denote the corresponding restricted correlation coefficients as $\pi'_j(s_1, s_2)$. And we denote these quantities of $\nabla^2 L_j(\boldsymbol{\beta}^*_j)$ as $\rho_{j-}(s), \rho_{j+}(s)$ and $\pi_j(s_1, s_2)$. By definition, we immediately have $\rho_{j-}(s) \le \rho'_{j-}(s) \le \rho'_{j+}(s) \le \rho_{j+}(s)$.

By Lemma C.4 we have

$$|\widehat{\boldsymbol{\omega}}^T\mathbf{\Lambda}_{j\backslash k,j\backslash k}\widehat{\boldsymbol{\omega}}| \ge \rho'_{j-}\big(k^\star + s^\star\big)\big[\|\widehat{\boldsymbol{\omega}}_I\|_2 - 2\pi'_j(s^\star + k_0^\star, s_0^\star)\|\widehat{\boldsymbol{\omega}}_{J^c}\|_1/k^\star\big]\|\widehat{\boldsymbol{\omega}}_I\|_2. \tag{D.7}$$

The following lemma relates the sparse eigenvalues of $\nabla^2 L_j(\boldsymbol{\beta}_j)$ to those of $\mathbb{E}\nabla^2 L_j(\boldsymbol{\beta}^*_j)$.

**Lemma D.1.** Under Assumptions 4.1, 4.3 and 4.6, if $n$ is sufficiently large such that $\rho_* \gtrsim s^*\lambda \log^2 d$, with probability at least $1 - (2d)^{-1}$, for all $j \in [d]$, there exists a constant $C_\rho \ge 33\rho_*^{-1}$ such that

$$\rho^*_{j-}(2s_0^\star + 2k_0^\star) - 0.05\nu_* \le \rho_{j-}(2s_0^\star + 2k_0^\star) < \rho_{j+}(k_0^\star) \le \rho^*_{j+}(k_0^\star) + 0.05\nu_*, \quad \text{and}$$
$$\rho_{j+}(k_0^\star)/\rho_{j-}(2s_0^\star + 2k_0^\star) \le 1 + 0.58k_0^\star/s_0^\star,$$

where we denote the local sparse eigenvalues $\rho_-\big(\nabla^2 L_j, \boldsymbol{\beta}^*_j; s, r\big)$ and $\rho_+\big(\nabla^2 L_j, \boldsymbol{\beta}^*_j; s, r\big)$ with $r = C_\rho\sqrt{\log d/n}$ as $\rho_{j-}(s)$ and $\rho_{j+}(s)$, respectively.

*Proof.* The proof is similar to that of Lemma 4.2, hence is omitted here. $\square$

By $\|\widehat{\boldsymbol{\omega}}_{J^c}\|_1 \le \|\widehat{\boldsymbol{\omega}}_J\|_1 \le \sqrt{s_0^\star}\|\widehat{\boldsymbol{\omega}}_J\|_2$ and Lemma D.1, the right-hand side of (D.7) can be reduced to

$$|\widehat{\boldsymbol{\omega}}^T \boldsymbol{\Lambda}_{j\backslash k, j\backslash k}\widehat{\boldsymbol{\omega}}| \ge 0.95\nu_*\big(\|\widehat{\boldsymbol{\omega}}_I\|_2 - 2\pi_j'(s_0^\star + k_0^\star, s^\star)\|\widehat{\boldsymbol{\omega}}_J\|_2\sqrt{s_0^\star/k_0^\star}\big)\|\widehat{\boldsymbol{\omega}}_I\|_2. \tag{D.8}$$

Using Lemma C.2 we obtain

$$\begin{aligned} 2\pi_j'(s_0^\star + k_0^\star, k_0^\star)\sqrt{s_0^\star/k_0^\star} &\le \sqrt{s_0^\star/k_0^\star}\sqrt{\rho_{j+}'(k_0^\star)/\rho_{j-}'(s_0^\star + 2k_0^\star) - 1} \\ \le &\sqrt{s_0^\star/k_0^\star}\sqrt{\rho_{j+}(k_0^\star)/\rho_{j-}(s_0^\star + 2k_0^\star) - 1} \le \sqrt{s_0^\star/k_0^\star}\sqrt{0.58k_0^\star/s_0^\star} \le 0.76. \end{aligned}$$

Thus the right-hand side of (D.8) can be reduced to

$$|\widehat{\boldsymbol{\omega}}^T \boldsymbol{\Lambda}_{j\backslash k, j\backslash k}\widehat{\boldsymbol{\omega}}| \ge 0.95\nu_*(1 - 0.76\|\widehat{\boldsymbol{\omega}}_J\|_2/\|\widehat{\boldsymbol{\omega}}_I\|_2)\|\widehat{\boldsymbol{\omega}}_I\|_2^2 \ge \nu_*\kappa\|\widehat{\boldsymbol{\omega}}_I\|_2^2, \tag{D.9}$$

where $\kappa = 0.22$. This inequality holds because $J \subset I$. By (D.9) we have

$$\nu_*\kappa\|\widehat{\boldsymbol{\omega}}_I\|_2^2 \le 4\sqrt{s_0^\star}\lambda_d\|\widehat{\boldsymbol{\omega}}_J\|_2 \le 4\sqrt{s_0^\star}\lambda_d\|\widehat{\boldsymbol{\omega}}_I\|_2, \quad \text{which implies } \|\widehat{\boldsymbol{\omega}}_I\|_2 \le 4\nu_*^{-1}\kappa^{-1}\sqrt{s_0^\star}\lambda_D.$$

Therefore the estimation error of $\widehat{\mathbf{w}}_{j,k}$ can be bounded by

$$\|\widehat{\boldsymbol{\omega}}\|_1 \le 2\|\widehat{\boldsymbol{\omega}}_J\|_1 \le 2\sqrt{s^\star}\|\widehat{\boldsymbol{\omega}}_J\|_2 \le 8\nu_*^{-1}\kappa^{-1}s_0^\star\lambda_D \le 37\nu_*^{-1}s_0^\star\lambda_D.$$

Returning to the original notations, we conclude that $\|\widehat{\mathbf{w}}_{j,k} - \mathbf{w}_{j,k}^*\|_1 \le 37\nu_*^{-1}s_0^\star\lambda_D$ for all $(j,k)$ such that $j,k \in [d]$, $j \ne k$. $\square$

### D.3 Proof of Lemma A.6

*Proof of Corollary A.6.* We only need to show that $\widehat{\sigma}_{jk}^2$ is a consistent estimator of $\sigma_{jk}^2$, which is equivalent to showing that $\lim_{n\to\infty}|\widehat{\sigma}_{jk}^2 - \sigma_{jk}^2| = 0$. To begin with, triangle inequality implies that

$$\begin{aligned} |\widehat{\sigma}_{jk}^2 - \sigma_{jk}^2| \le& \underbrace{\big|\widehat{\boldsymbol{\Sigma}}_{jk,jk}^{jk} - \boldsymbol{\Sigma}_{jk,jk}^{jk}\big|}_{I_1} + 2\underbrace{\big|\widehat{\mathbf{w}}_{j,k}^T\widehat{\boldsymbol{\Sigma}}_{j\backslash k,jk}^{jk} - \mathbf{w}_{j,k}^{*T}\boldsymbol{\Sigma}_{j\backslash k,jk}^{jk}\big|}_{I_{2j}} + \underbrace{\big|\widehat{\mathbf{w}}_{j,k}^T\widehat{\boldsymbol{\Sigma}}_{j\backslash k,j\backslash k}^{jk}\widehat{\mathbf{w}}_{j,k} - \mathbf{w}_{j,k}^{*T}\boldsymbol{\Sigma}_{j\backslash k,j\backslash k}^{jk}\mathbf{w}_{j,k}^*\big|}_{I_{3j}} \\ &+ 2\underbrace{\big|\widehat{\mathbf{w}}_{k,j}^T\widehat{\boldsymbol{\Sigma}}_{k\backslash j,jk}^{jk} - \mathbf{w}_{k,j}^{*T}\boldsymbol{\Sigma}_{k\backslash j,jk}^{jk}\big|}_{I_{2k}} + \underbrace{\big|\widehat{\mathbf{w}}_{k,j}^T\widehat{\boldsymbol{\Sigma}}_{k\backslash j,k\backslash j}^{jk}\widehat{\mathbf{w}}_{k,j} - \mathbf{w}_{k,j}^{*T}\boldsymbol{\Sigma}_{k\backslash j,k\backslash j}^{jk}\mathbf{w}_{k,j}^*\big|}_{I_{3k}}, \end{aligned}$$

where $\widehat{\boldsymbol{\Sigma}}^{jk} = \widehat{\boldsymbol{\Sigma}}^{jk}\big(\widehat{\boldsymbol{\beta}}_{j\vee k}'\big)$ and $\widehat{\boldsymbol{\Sigma}}^{jk}\big(\boldsymbol{\beta}_{j\vee k}\big)$ is defined as

$$\widehat{\boldsymbol{\Sigma}}^{jk}(\boldsymbol{\beta}_{j\vee k}) = \frac{1}{n}\sum_{i=1}^n\Big\{\frac{1}{n-1}\sum_{i'\ne i}\mathbf{h}_{ii'}^{jk}(\boldsymbol{\beta}_{j\vee k})\Big\}^{\otimes 2}. \tag{D.10}$$

To prove the consistency of $\widehat{\sigma}_{jk}^2$, we need the following theorem to show that $\widehat{\boldsymbol{\Sigma}}^{jk}$ is a consistent estimator of $\boldsymbol{\Sigma}^{jk}$ in the sense that $\big\|\widehat{\boldsymbol{\Sigma}}^{jk} - \boldsymbol{\Sigma}^{jk}\big\|_\infty$ is negligible.

**Lemma D.2.** For $1 \leq j < k \leq d$, let $\widehat{\boldsymbol{\Sigma}}^{jk}\big(\boldsymbol{\beta}_{j\vee k}\big)$ be defined as (D.10). Suppose $\widehat{\boldsymbol{\beta}}_j$ and $\widehat{\boldsymbol{\beta}}_k$ are the estimators of $\boldsymbol{\beta}_j^*$ and $\boldsymbol{\beta}_k^*$ obtained from Algorithm 1 and we denote $\widehat{\boldsymbol{\beta}}_{j\vee k} = (\widehat{\beta}_{jk}, \widehat{\boldsymbol{\beta}}_{j\backslash k}^T, \widehat{\boldsymbol{\beta}}_{k\backslash j}^T)^T$. Then $\widehat{\boldsymbol{\Sigma}}^{jk}(\widehat{\boldsymbol{\beta}}_{j\vee k})$ is a consistent estimator of $\boldsymbol{\Sigma}^{jk}$. There exist a constant $C_\Sigma$ that does not depend on $(j,k)$ such that, with probability tending to one,

$$\big\|\widehat{\boldsymbol{\Sigma}}^{jk}(\widehat{\boldsymbol{\beta}}_{j\vee k}) - \boldsymbol{\Sigma}^{jk}\big\|_\infty \leq C_\Sigma s^* \lambda \log^2 d \quad \text{for } 1 \leq j < k \leq d.$$

*Proof of Lemma D.2.* See §E.2.1 for a detailed proof. □

In the rest of the proof, we will omit the superscripts in both $\widehat{\boldsymbol{\Sigma}}^{jk}$ and $\boldsymbol{\Sigma}^{jk}$ for notational simplicity. By Lemma D.2,

$$I_1 \leq \|\widehat{\boldsymbol{\Sigma}} - \boldsymbol{\Sigma}\|_\infty \leq \mathcal{O}_{\mathbb{P}}\big(s^* \lambda \log^2 d\big). \tag{D.11}$$

By triangle inequality, we have the following inequality for $I_2$ :

$$I_{2j} \leq \underbrace{\big|(\widehat{\mathbf{w}}_{j,k} - \mathbf{w}_{j,k}^*)^T\big(\widehat{\boldsymbol{\Sigma}}_{j\backslash k,jk} - \boldsymbol{\Sigma}_{j\backslash k,jk}\big)\big|}_{I_{21}} + \underbrace{\big|(\widehat{\mathbf{w}}_{j,k} - \mathbf{w}_{j,k}^*)^T\boldsymbol{\Sigma}_{j\backslash k,jk}\big|}_{I_{22}} + \underbrace{\big|\mathbf{w}_{j,k}^{*T}\big(\widehat{\boldsymbol{\Sigma}}_{j\backslash k,jk} - \boldsymbol{\Sigma}_{j\backslash k,jk}\big)\big|}_{I_{23}}.$$

By Hölder's inequality, Lemma D.2 and the estimation error of $\widehat{\mathbf{w}}_{j,k}$, we obtain an upper-bound for $I_{21}$ as follows:

$$I_{21} \leq \|\widehat{\mathbf{w}}_{j,k} - \mathbf{w}_{j,k}^*\|_1 \|\widehat{\boldsymbol{\Sigma}} - \boldsymbol{\Sigma}\|_\infty = \mathcal{O}_{\mathbb{P}}\big(s^* s_0^\star \lambda_D \lambda \log^2 d\big). \tag{D.12}$$

Similarly, for $I_{22}$, Hölder's inequality implies that

$$I_{22} \leq \|\widehat{\mathbf{w}}_{j,k} - \mathbf{w}_{j,k}^*\|_1 \|\boldsymbol{\Sigma}\|_\infty = \mathcal{O}_{\mathbb{P}}\big(s_0^\star \lambda_D D\big), \tag{D.13}$$

where the constant $D$ appears in (D.1). For $I_{23}$, by Hölder's inequality and D.2 we obtain

$$I_{23} \leq \|\mathbf{w}_{j,k}^*\|_1 \|\widehat{\boldsymbol{\Sigma}} - \boldsymbol{\Sigma}\|_\infty = \mathcal{O}_{\mathbb{P}}\big(w_0 s^* \lambda \log^2 d\big). \tag{D.14}$$

Combining (D.12), (D.13) and (D.14) we have

$$I_{2j} \lesssim (w_0 + s_0^\star \lambda_D) s^* \lambda \log^2 d + s_0^\star \lambda_D. \tag{D.15}$$

For $I_{3j}$, by triangle inequality we have

$$I_{3j} \leq \underbrace{\big|\widehat{\mathbf{w}}_{j,k}^T\big(\widehat{\boldsymbol{\Sigma}}_{j\backslash k,j\backslash k} - \boldsymbol{\Sigma}_{j\backslash k,j\backslash k}\big)\widehat{\mathbf{w}}_{j,k}\big|}_{I_{31}} + \underbrace{\big|\widehat{\mathbf{w}}_{j,k}^T\boldsymbol{\Sigma}_{j\backslash k,j\backslash k}\widehat{\mathbf{w}}_{j,k} - \mathbf{w}_{j,k}^{*T}\boldsymbol{\Sigma}_{j\backslash k,j\backslash k}\mathbf{w}_{j,k}^*\big|}_{I_{32}}.$$

For term $I_{31}$, Hölder's inequality and the optimality of $\widehat{\mathbf{w}}$ implies that

$$I_{31} \leq \|\widehat{\mathbf{w}}_{j,k}\|_1^2 \|\widehat{\boldsymbol{\Sigma}}_{j\backslash k,j\backslash k} - \boldsymbol{\Sigma}_{j\backslash k,j\backslash k}\|_\infty \leq C_\Sigma w_0^2 s^* \lambda \log^2 d. \tag{D.16}$$

For term $I_{32}$, Lemma B.2 implies that

$$I_{32} \leq \|\boldsymbol{\Sigma}_{j\backslash k,j\backslash k}\|_\infty \|\widehat{\mathbf{w}}_{j,k} - \mathbf{w}_{j,k}^*\|_1^2 + \|\boldsymbol{\Sigma}_{j\backslash k,j\backslash k}\mathbf{w}_{j,k}^*\|_\infty \|\widehat{\mathbf{w}}_{j,k} - \mathbf{w}_{j,k}^*\|_1 \leq \big(D\omega_0 s_0^\star \lambda_D + D s_0^{\star 2} \lambda_D^2\big), \tag{D.17}$$

where we use Hölder's inequality $\|\boldsymbol{\Sigma}_{j\backslash k,j\backslash k}\mathbf{w}^*_{j,k}\|_\infty \le \|\mathbf{w}^*_{j,k}\|_1\|\boldsymbol{\Sigma}\|_\infty \le Dw_0$. By (D.16), (D.17) and $\lambda_D \gtrsim w_0 s^* \lambda \log^2 d$, we obtain

$$I_{3j} \lesssim w_0^2 s^* \lambda \log^2 d + \big(D\omega_0 s_0^\star \lambda_D + D s_0^{\star 2} \lambda_D^2\big). \tag{D.18}$$

Therefore combining (D.11), (D.15) and (D.18) we obtain $I_1 + I_{2j} + I_{3j} = o_{\mathbb{P}}(1)$. We can show similarly that $I_{2k} + I_{3k} = o_{\mathbb{P}}(1)$. Thus $\lim_{n\to\infty} \max_{j<k} \big|\widehat{\sigma}_{jk}^2 - \sigma_{jk}^2\big| = 0$ with probability converging to one. □

# E Proof of Technical Lemmas

Finally, we prove the technical lemmas in this appendix. Specifically, we prove the lemmas introduced to derive the auxiliary results.

## E.1 Proof of Technical Lemmas in §C

In this subsection we prove the technical lemmas we use to prove the auxiliary results of estimation. These lemmas are standard for high-dimensional linear regression, but proving them for our logistic-type loss function needs nontrivial extensions.

### E.1.1 Proof of Lemma C.2

*Proof of Lemma C.2.* Let $I$ and $J$ be two index sets with $I \cap J = \emptyset, |I| \le s, |J| \le k$, for any $\mathbf{u} \in \mathbb{R}^d$ with $\|\mathbf{u} - \mathbf{u}_0\|_2 \le r$ and any $\mathbf{v}, \mathbf{w} \in \mathbb{R}^d$, let $\boldsymbol{\theta} = \mathbf{v}_I + \alpha \mathbf{w}_J$ with some $\alpha \in \mathbb{R}$, then by definition, $\|\boldsymbol{\theta}\|_0 \le s + k$. For notational simplicity, we denote $s$-sparse eigenvalues $\rho_+\big(\mathbf{M}, \mathbf{u}_0; s, r\big)$ and $\rho_-\big(\mathbf{M}, \mathbf{u}_0; s, r\big)$ as $\rho_-(s)$ and $\rho_+(s)$ respectively. By definition, we have

$$\rho_-(s+k)\|\boldsymbol{\theta}\|_2^2 \le \boldsymbol{\theta}^T\mathbf{M}(\mathbf{u})\boldsymbol{\theta} = \underbrace{\mathbf{v}_I^T\mathbf{M}(\mathbf{u})\mathbf{v}_I}_{A1} + 2\alpha \underbrace{\mathbf{v}_I^T\mathbf{M}(\mathbf{u})\mathbf{w}_J}_{A2} + \alpha^2 \underbrace{\mathbf{w}_J^T\mathbf{M}(\mathbf{u})\mathbf{w}_J}_{A3}. \tag{E.1}$$

Since $\|\boldsymbol{\theta}\|_2^2 = \|\mathbf{v}_I\|_2^2 + \alpha^2\|\mathbf{w}_J\|_2^2$. Rearranging the terms in (E.1) we have

$$\big[A_3 - \rho_-(s+k)\|\mathbf{w}_J\|_2^2\big]\alpha^2 + 2A_2\alpha + \big[A_1 - \rho_-(s+k)\|\mathbf{v}_I\|_2^2\big] \ge 0 \quad \text{for all } \alpha \in \mathbb{R}. \tag{E.2}$$

Note that the left-hand side (E.2) is a univariate quadratic function in $\alpha$, thus (E.2) implies that

$$\big[A_1 - \rho_-(s+k)\|\mathbf{v}_I\|_2^2\big]\big[A_3 - \rho_-(s+k)\|\mathbf{w}_J\|_2^2\big] \ge A_2^2. \tag{E.3}$$

Therefore by multiplying $4\|\mathbf{v}_I\|_2^\infty/(A_1^2\|\mathbf{w}_J\|_2^2)$ to both sides of (E.3) we have

$$\frac{4A_2^2\|\mathbf{v}_I\|_2^2}{A_1^2\|\mathbf{w}_J\|_2^2} \le \frac{4\|\mathbf{v}_I\|_2^2}{A_1\|\mathbf{w}_J\|_2^2}\left[\frac{A_1 - \rho_-(s+k)\|\mathbf{v}_I\|_2^2}{A_1}\right]\big[A_3 - \rho_-(s+k)\|\mathbf{w}_J\|_2^2\big]. \tag{E.4}$$

By the inequality of arithmetic and geometric means, we have

$$\frac{\rho_-(s+k)\|\mathbf{v}_I\|_2^2}{A_1}\left[\frac{A_1 - \rho_-(s+k)\|\mathbf{v}_I\|_2^2}{A_1}\right] \le \frac{1}{4}.$$

Then the right-hand side of (E.3) can be bounded by

$$\frac{4A_2^2\|\mathbf{v}_I\|_2^2}{A_1^2\|\mathbf{w}_J\|_2^2} \le \frac{A_3 - \rho_-(s+k)\|\mathbf{w}_J\|_2^2}{\rho_-(s+k)\|\mathbf{w}_J\|_2^2} \le \frac{\rho_+(k)}{\rho_-(s+k)} - 1,$$

where the last inequality follows from $A_3 \le \rho_+(k)\|\mathbf{w}_J\|_2^2$. Note that by the relationship between $\ell_2$- and $\ell_\infty$ norm, we have $\|\mathbf{w}_J\|_2 \le \sqrt{k}\|\mathbf{w}_J\|_\infty$, which further implies that

$$\frac{\mathbf{v}_I^T\mathbf{M}(\mathbf{u})\mathbf{w}_J\|\mathbf{v}_I\|_2}{\mathbf{v}_I^T\mathbf{M}(\mathbf{u})\mathbf{v}_I\|\mathbf{w}_J\|_\infty} \le \frac{\sqrt{k}\mathbf{v}_I^T\mathbf{M}(\mathbf{u})\mathbf{w}_J\|\mathbf{v}_I\|_2}{\mathbf{v}_I^T\mathbf{M}(\mathbf{u})\mathbf{v}_I\|\mathbf{w}_J\|_2} = \frac{\sqrt{k}A_2\|\mathbf{v}_I\|_2}{A_1\|\mathbf{w}_J\|_2} \le \frac{\sqrt{k}}{2}\sqrt{\rho_+(k)/\rho_-(s+k) - 1}.$$

Taking a supremum over $\mathbf{v}, \mathbf{w} \in \mathbb{R}^d$ finally yields Lemma C.2. ☐

### E.1.2 Proof of lemma C.3

*Proof of lemma C.3.* Under Assumption 4.3, for any $\boldsymbol{\beta}_j \in \mathbb{R}^{d-1}$ such that $\|\boldsymbol{\beta}_j - \boldsymbol{\beta}_j^*\|_2 \le r$ and any $\mathbf{v} \in \mathbb{R}^{d-1}$ with $\|\mathbf{v}\|_0 \le 2s^* + 2k^*$, we denote $\nabla^2 L_j(\boldsymbol{\beta}_j) - \nabla^2 L_j(\boldsymbol{\beta}_j^*)$ and $\nabla^2 L_j(\boldsymbol{\beta}_j) - \mathbb{E}\big[\nabla^2 L_j(\boldsymbol{\beta}_j^*)\big]$ as $\boldsymbol{\Lambda}_1$ and $\boldsymbol{\Lambda}_2$ respectively. Our goal is to show that both $|\mathbf{v}^T\boldsymbol{\Lambda}_1\mathbf{v}|$ and $|\mathbf{v}^T\boldsymbol{\Lambda}_2\mathbf{v}|$ are negligible. Hölder's inequality implies that $\big|\mathbf{v}^T\boldsymbol{\Lambda}_2\mathbf{v}\big| \le \|\mathbf{v}\|_1\|\boldsymbol{\Lambda}_2\mathbf{v}\|_\infty \le \|\mathbf{v}\|_1^2\|\boldsymbol{\Lambda}_2\|_\infty$. We use the following lemma to control $|\mathbf{v}^\top\boldsymbol{\Lambda}_1\mathbf{v}|$ and $\|\boldsymbol{\Lambda}_2\|_\infty$.

**Lemma E.1.** We denote $s^* = \max_{j\in[d]} \|\boldsymbol{\beta}_j^*\|_0$. For $r_1(s^*, n, d) > 0$ satisfying $\lim_{n\to\infty} r_1(s^*, n, d)\log^2 d = 0$, we denote $\mathbb{B}_j(r_1) := \big\{\boldsymbol{\beta}_j \in \mathbb{R}^{d-1} : \big\|\boldsymbol{\beta}_j - \boldsymbol{\beta}_j^*\big\|_1 \le r_1(s^*, n, d)\big\}$ as the $\ell_1$-ball centered at $\boldsymbol{\beta}_j^*$ with radius $r_1(s^*, n, d)$. Under Assumptions 4.1 and 4.3, there exist absolute constants $C_h, C_r > 0$ such that, with probability at least $1 - (2d)^{-1}$, for all $j \in [d]$, $\boldsymbol{\beta}_j \in \mathbb{B}_j(r_1)$ and $\mathbf{v} \in \mathbb{R}^d$, it holds that,

$$\big\|\nabla^2 L_j(\boldsymbol{\beta}_j^*) - \mathbb{E}\big[\nabla^2 L_j(\boldsymbol{\beta}_j^*)\big]\big\|_\infty \le C_h\sqrt{\log d/n}, \tag{E.5}$$

$$\big\|\nabla^2 L_j(\boldsymbol{\beta}_j) - \nabla^2 L_j(\boldsymbol{\beta}_j^*)\big\|_\infty \le C_r r_1(s^*, n, d)\log^2 d \quad \text{and} \tag{E.6}$$

$$\big|\mathbf{v}^T\big[\nabla^2 L_j(\boldsymbol{\beta}_j) - \nabla^2 L_j(\boldsymbol{\beta}_j^*)\mathbf{v}\big| \le C_r r_1(s^*, n, d)\|\mathbf{v}\|_2^2. \tag{E.7}$$

*Proof of Lemma E.1.* See §E.3 for a detailed proof. ☐

Lemma E.1 implies that $\|\boldsymbol{\Lambda}_2\|_\infty \le C_h\sqrt{\log d/n}$ with probability at least $1 - (2d)^{-1}$. By the relation between $\ell_1$- and $\ell_2$-norms, we have

$$\big|\mathbf{v}^T\boldsymbol{\Lambda}_2\mathbf{v}\big| \le (2s^* + 2k^*)\|\mathbf{v}\|_2^2\|\boldsymbol{\Lambda}\|_\infty \le (2s^* + 2k^*)C_h\sqrt{\log d/n}.$$

Moreover, setting $r = C_\rho s^*\sqrt{\log d/n}$ with $C_\rho \ge 33\rho_*^{-1}$ , we have

$$|\mathbf{v}^T\boldsymbol{\Lambda}_1\mathbf{v}| \le C_r C_\rho\|\mathbf{v}\|_1^2 \le C_r C_\rho(2s^* + 2k^*)\sqrt{\log d/n}$$

. By Assumption 4.3, if $n$ is large enough such that $(2s^* + 2k^*)(C_r C_\rho + C_h)\sqrt{\log d/n}\big] \le 0.05\rho_*$, then we have

$$0.95\rho_* \le \rho_{j-}^*(2s^* + 2k^*) - 0.05\rho_* \le \rho_{j-}(2s^* + 2k^*) < \rho_{j+}(k^*) \le \rho_{j+}^*(k^*) + 0.05\rho_*,$$

where we denote the $s$-sparse eigenvalues $\rho_-\big(\nabla^2 L_j, \boldsymbol{\beta}_j^*; s, r\big)$ and $\rho_+\big(\nabla^2 L_j, \boldsymbol{\beta}_j^*; s, r\big)$ as $\rho_{j-}(s)$ and $\rho_{j+}(s)$ respectively. Under Assumption 4.3, $\rho_{j+}^*(k^*)/\rho_{j-}^*(2s^*+2k^*) \le 1 + 0.2k^*/s^*$ and $k^* \ge 2s^*$, simple computation yields that

$$\frac{\rho_{j+}(k^*)}{\rho_{j-}(2s^*+2k^*)} \le \frac{\rho_{j+}^*(k^*) + 0.05\rho_*}{\rho_{j-}^*(2s^*+2k^*) - 0.05\rho_*} \le \frac{\rho_{j+}^*(k^*) + 0.05\rho_{j-}^*(2s^*+2k^*)}{0.95\rho_{j-}^*(2s^*+2k^*)} \le 1 + 0.27k^*/s^*.$$

□

### E.1.3 Proof of Lemma C.4

*Proof of Lemma C.4.* For $\mathbf{v} = (v_1, \ldots, v_d)^T \in \mathbb{R}^d$, without loss of generality, we assume that $F^c = [s_1]$ where $s_1 = |F^c| \le s$. In addition, we assume that when $j > s_1$, $v_j$ is arranged in descending order of $|v_j|$. That is, we rearrange the components of $\mathbf{v}$ such that $|v_j| \ge |v_{j+1}|$ for all $j \ge s_1$. Let $J_0 = [s_1]$ and $J_i = \{s_1 + (i-1)k + 1, \ldots, \min(s_1 + ik, d)\}$. By definition, we have $J = J_1$ and $I = J_0 \cup J_1$. Moreover, we have $\|\mathbf{v}_{J_i}\|_\infty \le \|\mathbf{v}_{J_{i-1}}\|_1/k$ when $i \ge 2$ because by the definition of $J_i$ we have $\sum_{i\ge 2} \|\mathbf{v}_{J_i}\|_\infty \le \|\mathbf{v}_F\|_1/k$. Note that by the definition of index sets $I$ and $J_i$, $|J_i| \le k$ and $|I| = k + s_1 \le k + s$. We denote the restricted correlation coefficients $\pi(\mathbf{M}, \mathbf{u}_0; s, k, r)$ as $\pi(s, k)$, then by the definition of $\pi(s+k, k)$ we have

$$\big|\mathbf{v}_I^T \mathbf{M}(\mathbf{u}) \mathbf{v}_{J_i}\big| \le \pi(s+k, k)\big[\mathbf{v}_I^T \mathbf{M}(\mathbf{u}) \mathbf{v}_I\big]\|\mathbf{v}_{J_i}\|_\infty / \|\mathbf{v}_I\|_2.$$

Thus we have the following upper bound for $\big|\mathbf{v}_I^T \mathbf{M}(\mathbf{u}) \mathbf{v}_{I^c}\big|$:

$$\begin{aligned}
\big|\mathbf{v}_I^T \mathbf{M}(\mathbf{u}) \mathbf{v}_{I^c}\big| &\le \sum_{i\ge 2} \big|\mathbf{v}_I^T \mathbf{M}(\mathbf{u}) \mathbf{v}_{J_i}\big| \le \pi(s+k, k)\|\mathbf{v}_I\|_2^{-1}\big[\mathbf{v}_I^T \mathbf{M}(\mathbf{u}) \mathbf{v}_I\big] \sum_{i\ge 2} \|\mathbf{v}_{J_i}\|_\infty \\
&\le \pi(s+k, k)\|\mathbf{v}_I\|_2^{-1}\big[\mathbf{v}_I^T \mathbf{M}(\mathbf{u}) \mathbf{v}_I\big]\|\mathbf{v}_F\|_1/k. \qquad \text{(E.8)}
\end{aligned}$$

Because $\mathbf{v}^T \mathbf{M}(\mathbf{u}) \mathbf{v} \ge \mathbf{v}_I^T \mathbf{M}(\mathbf{u}) \mathbf{v}_I + 2\mathbf{v}_I^T \mathbf{M}(\mathbf{u}) \mathbf{v}_{I^c}$, by (E.8) we have

$$\begin{aligned}
\mathbf{v}^T \mathbf{M}(\mathbf{u}) \mathbf{v} &\ge \mathbf{v}_I^T \mathbf{M}(\mathbf{u}) \mathbf{v}_I - 2\pi(s+k, k)\|\mathbf{v}_I\|_2^{-1}\big[\mathbf{v}_I^T \mathbf{M}(\mathbf{u}) \mathbf{v}_I\big]\|\mathbf{v}_F\|_1/k \\
&= \big[\mathbf{v}_I^T \mathbf{M}(\mathbf{u}) \mathbf{v}_I\big]\big[1 - 2\pi(s+k, k)\|\mathbf{v}_I\|_2^{-1}\|\mathbf{v}_F\|_1/k\big].
\end{aligned}$$

Thus we can bound the right-hand side of the last formula using the sparse eigenvalue condition

$$\mathbf{v}^T \mathbf{M}(\mathbf{u}) \mathbf{v} \ge \rho_-(s+k)\big[1 - 2\pi(s+k, k)k^{-1}\|\mathbf{v}_I\|_2^{-1}\|\mathbf{v}_F\|_1\big]\|\mathbf{v}_I\|_2^2, \qquad \text{(E.9)}$$

where we denote $s$-sparse eigenvalue $\rho_-(\mathbf{M}, \mathbf{u}_0; s, r)$ as $\rho_-(s+k)$ for the simplicity of notations. Inequality (E.9) concludes the proof of Lemma C.4. □

### E.1.4 Proof of Lemma C.5

*Proof of Lemma C.5.* Let $F(t) = L_j\big(\boldsymbol{\beta}(t)\big) - L_j(\boldsymbol{\beta}_1) - \big\langle \nabla L_j(\boldsymbol{\beta}_1), \boldsymbol{\beta}(t) - \boldsymbol{\beta}_1\big\rangle$, because the derivative of $L_j\big(\boldsymbol{\beta}(t)\big)$ with respect to $t$ is $\big\langle \nabla L_j\big(\boldsymbol{\beta}(t)\big), \boldsymbol{\beta}_2 - \boldsymbol{\beta}_1\big\rangle$ then the derivative of $F$ is

$$F'(t) = \big\langle \nabla L_j\big(\boldsymbol{\beta}(t)\big) - \nabla L_j(\boldsymbol{\beta}_1), \boldsymbol{\beta}_2 - \boldsymbol{\beta}_1\big\rangle.$$

Therefore the Bregman divergence $D_j\big(\boldsymbol{\beta}(t),\boldsymbol{\beta}_1\big)$ can be written as

$$D_j\big(\boldsymbol{\beta}(t),\boldsymbol{\beta}_1\big)=\big\langle \nabla L_j[\boldsymbol{\beta}(t)]-\nabla L_j(\boldsymbol{\beta}_1),t(\boldsymbol{\beta}_2-\boldsymbol{\beta}_1)\big\rangle = tF'(t).$$

By definition, it is easy to see that $F'(1)=D_j(\boldsymbol{\beta}_2,\boldsymbol{\beta}_1)$. To derive Lemma C.5, it suffices to show that $F(t)$ is convex, which implies that $F'(t)$ is non-decreasing and $D_j\big(\boldsymbol{\beta}(t),\boldsymbol{\beta}_1\big)=tF'(t)\le tF'(1)=tD_j(\boldsymbol{\beta}_2,\boldsymbol{\beta}_1)$.

For $\forall t_1,t_2\in\mathbb{R}_+, t_1+t_2=1, x,y\in(0,1)$, by the linearity of $\boldsymbol{\beta}(t)$, $\boldsymbol{\beta}(t_1x+t_2y)=t_1\boldsymbol{\beta}(x)+t_2\boldsymbol{\beta}(y)$. Then we have

$$\big\langle \nabla L_j(\boldsymbol{\beta}_1),\boldsymbol{\beta}(t_1x+t_2y)-\boldsymbol{\beta}_1\big\rangle = t_1\big\langle \nabla L_j(\boldsymbol{\beta}_1),\boldsymbol{\beta}(x)-\boldsymbol{\beta}_1\big\rangle + t_2\big\langle \nabla L_j(\boldsymbol{\beta}_1),\boldsymbol{\beta}(y)-\boldsymbol{\beta}_1\big\rangle. \tag{E.10}$$

In addition, by convexity of function $L_j(\cdot)$, we obtain

$$L_j\big(\boldsymbol{\beta}(t_1x+t_2y)\big)\le t_1L_j\big(\boldsymbol{\beta}(x)\big)+t_2L_j\big(\boldsymbol{\beta}(y)\big). \tag{E.11}$$

Adding (E.10) and (E.11) we obtain

$$F\big(t_1x+t_2y\big)\le t_1F(x)+t_2F(y).$$

Therefore $F(t)$ is convex, thus we have $D_j(\boldsymbol{\beta}(t),\boldsymbol{\beta}_1)\le tD_j(\boldsymbol{\beta}_2,\boldsymbol{\beta}_1)$. □

## E.2 Proof of Technical Lemmas in §D

Now we prove the lemmas that supports the auxiliary inferential results. We first prove Lemma D.2, which implies that the $\widehat{\sigma}_{jk}^2$ is a consistent estimator of the asymptotic variance of $\sigma_{jk}$.

### E.2.1 Proof of lemma D.2

*Proof of Lemma D.2.* Recall that we denote $\boldsymbol{\beta}_{j\vee k}=(\beta_{jk},\boldsymbol{\beta}_{j\setminus k},\boldsymbol{\beta}_{k\setminus j})$ and $L_{jk}\big(\boldsymbol{\beta}_{j\vee k}\big)=L_j(\boldsymbol{\beta}_j)+L_k(\boldsymbol{\beta}_k)$. We denote the kernel function of the second-order $U$-statistic $\nabla L_{jk}\big(\boldsymbol{\beta}_{j\vee k}\big)$ as $\mathbf{h}_{ii'}^{jk}\big(\boldsymbol{\beta}_{j\vee k}\big)$ where the subscripts $i,i'$ indicate that $\mathbf{h}_{ii'}^{jk}(\cdot)$ depends on $\boldsymbol{X}_i$ and $\boldsymbol{X}_{i'}$. We define $\mathbf{V}_{ii'i''}^{jk}\big(\boldsymbol{\beta}_{j\vee k}\big):=\mathbf{h}_{ii'}^{jk}\big(\boldsymbol{\beta}_{j\vee k}\big)\mathbf{h}_{ii'}^{jk}\big(\boldsymbol{\beta}_{j\vee k}\big)^T$. Then by definition, $\widehat{\boldsymbol{\Sigma}}^{jk}\big(\boldsymbol{\beta}_{j\setminus k}\big)$can be written as

$$\widehat{\boldsymbol{\Sigma}}^{jk}\big(\boldsymbol{\beta}_{j\vee k}\big)=\frac{1}{n(n-1)^2}\sum_{i=1}^n\sum_{i'\neq i,i''\neq i}\mathbf{V}_{ii'i''}^{jk}\big(\boldsymbol{\beta}_{j\vee k}\big).$$

Note that $\widehat{\boldsymbol{\Sigma}}^{jk}\big(\boldsymbol{\beta}_{j\vee k}\big)-\boldsymbol{\Sigma}^{jk}=\underbrace{\widehat{\boldsymbol{\Sigma}}^{jk}\big(\boldsymbol{\beta}_{j\vee k}\big)-\widehat{\boldsymbol{\Sigma}}^{jk}\big(\boldsymbol{\beta}_{j\vee k}^*\big)}_{I_1}+\underbrace{\widehat{\boldsymbol{\Sigma}}^{jk}\big(\boldsymbol{\beta}_{j\vee k}^*\big)-\boldsymbol{\Sigma}^{jk}}_{I_2}.$

We first consider $I_2$. For notational simplicity, we use $\mathbf{h}_{ii'}$ and $\mathbf{h}_{ii'|i}$ to denote $\mathbf{h}_{ij}^{jk}\big(\boldsymbol{\beta}_{j\vee k}^*\big)$ and $\mathbf{h}_{ii'|i}^{jk}\big(\boldsymbol{\beta}_{j\vee k}^*\big):=\mathbb{E}\big[\mathbf{h}_{ij}^{jk}\big(\boldsymbol{\beta}_{j\vee k}^*\big)\big|\boldsymbol{X}_i\big]$ respectively. As shown in §D.1, for $i\neq i'\neq i''$,

$$\mathbb{E}\big(\mathbf{h}_{ii'}\mathbf{h}_{ii''}^T\big)=\mathbb{E}\big(\mathbf{h}_{ii'}\mathbf{h}_{ii''}^T\big|\boldsymbol{X}_i\big)=\mathbb{E}\big(\mathbf{h}_{ii'|i}\mathbf{h}_{ii''|i}^T\big)=\boldsymbol{\Sigma}^{jk}\;\text{ and }\;\mathbb{E}\big(\mathbf{h}_{ij}\mathbf{h}_{ij}^T\big)=\boldsymbol{\Theta}^{jk},$$

we can write $I_2$ as

$$I_2 = \frac{n-2}{n-1} \underbrace{\left\{ \binom{n}{3}^{-1} \sum_{i<i'<i''} \big[ \mathbf{V}_{ii'i''} - \mathbb{E}(\mathbf{V}_{ii'i''}) \big] \right\}}_{I_{21}} + \frac{1}{n-1} \underbrace{\left\{ \binom{n}{2}^{-1} \sum_{i<i'} \big[ \mathbf{V}_{ii'} - \mathbb{E}(\mathbf{V}_{ii'}) \big] \right\}}_{I_{22}} + \frac{1}{n-1} (\mathbf{\Theta}^{jk} - \mathbf{\Sigma}^{jk}),$$

where we use $\mathbf{V}_{ii'i''}$ to denote $\mathbf{V}^{jk}_{ii'i''}(\boldsymbol{\beta}^*_{j\vee k})$. Observing that $I_{21}$ is a centered third order $U$-statistic, for $x$ large enough such that $x^4 \geq \big\| \mathbb{E}\big[\mathbf{V}_{ijk}(\boldsymbol{\beta}^*_{j\vee k})\big] \big\|_\infty$ and for any $(a,b),(c,d) \in \big\{(p,q) \colon p,q \in \{j,k\}\big\}$ we have

$$\begin{aligned} \mathbb{P}\big( \big[ \mathbf{V}^{jk}_{ii'i''}(\boldsymbol{\beta}_{j\vee k}) \big]_{ab,cd} > 2x^4 \big) &\leq \mathbb{P}\big[ (X_{ia} - X_{i'a})(X_{ib} - X_{i'b})(X_{ic} - X_{i''c})(X_{id} - X_{i''d}) > x^4 \big] \\ &\leq 8 \exp(2\kappa_m + \kappa_h) \exp(-x). \end{aligned}$$

Thus there exist constants $c_1$ and $C_1$ that does not depend on $n$ or $d$ or $(j,k)$ such that for any $x \in \mathbb{R}$, any $i,i',i'' \in [n]$ and any $j,k \in [d]$,

$$\mathbb{P}\big( [\mathbf{V}^{jk}_{ii'i''}(\boldsymbol{\beta}^*_{j\vee k})]_{ab,cd} > x \big) \leq C_1 \exp(c_1 x^{1/4}). \tag{E.12}$$

This implies that there exists some generic constant $C$ such that $\|\mathbf{V}^{jk}_{ii'i''}(\boldsymbol{\beta}^*_{j\vee k})\|_\infty \leq C \log^4 d$ for all $j,k \in [d]$ and $i,i' \in [n]$ with probability tending to one. Similar to the method we use in §E.3, we define $\mathcal{E} \coloneqq \big\{ \|\mathbf{V}^{jk}_{ii'i''}(\boldsymbol{\beta}^*_{j\vee k})\|_\infty \leq C \log^4 d, \forall i,i',i'' \in [n], j,k \in [d] \big\}$. By Bernstein's inequality for $U$-statistics (Lemma C.1) with $b = C\log^4 d$ in (C.2), for some generic constants $C$, it holds with high probability that

$$\binom{n}{2}^{-1} \sum_{i<i'} \big[ \mathbf{V}_{ii'} - \mathbb{E}(\mathbf{V}_{ii'} | \mathcal{E}) \big] \leq C\sqrt{\log d/n}, \quad \forall j,k \in [d], i,i',i'' \in [n]. \tag{E.13}$$

Moreover, by (E.12), we have

$$\begin{aligned} &\mathbb{E}\big\{ [\mathbf{V}_{ii'i'}(\boldsymbol{\beta}^*_{j\vee k})]_{ab,cd} | \mathcal{E} \big\} - \mathbb{E}\big\{ [V_{ii'i''}(\boldsymbol{\beta}^*_{j\vee k})]_{ab,cd} \big\} \\ &\leq \int_{C\log^4 d}^{\infty} \mathbb{P}\big\{ \big| [\mathbf{V}^{jk}_{ii'i''}(\boldsymbol{\beta}^*_{j\vee k})]_{ab,cd} \big| > x \big\} \leq c_1 \log^3 d \cdot \exp(-c_2 \log d) \end{aligned} \tag{E.14}$$

for some absolute constant $c_1$ and $c_2$. Since (E.14) holds uniformly, we have

$$\binom{n}{2}^{-1} \sum_{i<i'} \big[ \mathbb{E}(\mathbf{V}_{ii'} | \mathcal{E}) - \mathbb{E}(\mathbf{V}_{ii'i''}) \big] \leq \log^3 d \cdot \exp(-c_2 \log d) \lesssim \sqrt{\log d/n}. \tag{E.15}$$

Combining (E.13) and (E.15) we obtain that

$$\|I_{21}\|_\infty = \mathcal{O}_{\mathbb{P}}\big(\sqrt{\log d/n}\big) \quad \text{uniformly for } 1 \leq j < k \leq n. \tag{E.16}$$

For the second part $I_{22}$, noting that it is a $U$-statistic of order 2, because (E.12) also holds for $\mathbf{V}_{ii'i''}(\boldsymbol{\beta}^*_{j\vee k})$, applying the same technique, we have $\|I_{21}\|_\infty = \mathcal{O}_{\mathbb{P}}\big(\sqrt{\log d/n}\big)$ uniformly for $1 \leq j < k \leq n$. Combining with (E.16), we conclude that, for some absolute constant $C$, we have

$$\big\| \widehat{\mathbf{\Sigma}}^{jk}(\boldsymbol{\beta}^*_{j\vee k}) - \mathbf{\Sigma}^{jk} \big\|_\infty \leq C\sqrt{\log d/n}, \quad \forall 1 \leq j < k \leq n. \tag{E.17}$$

Now we turn to $I_1$. For any $\boldsymbol{\beta}_j, \boldsymbol{\beta}_k \in \mathbb{R}^{d-1}$ such that $\|\boldsymbol{\beta}_j - \boldsymbol{\beta}_j^*\|_1 \le r(s^*, n, d)$ and $\|\boldsymbol{\beta}_k - \boldsymbol{\beta}_k^*\|_1 \le r(s^*, n, d)$, we denote $\omega_{ii'}^j := \exp\big[-(X_{ij} - X_{i'j})(\boldsymbol{\beta}_j - \boldsymbol{\beta}_j^*)^T(\boldsymbol{X}_{i\backslash j} - \boldsymbol{X}_{i'\backslash j})\big]$ and denote $\omega_{ii'}^k$ similarly. Recall that we denote $R_{ii'}^j(\boldsymbol{\beta}_j) = \exp\big[-(x_{ij} - x_{i'j})\boldsymbol{\beta}_j{}^T(\boldsymbol{x}_{i\backslash j} - \boldsymbol{x}_{i'\backslash j})\big]$. Hence by definition we have $R_{ii'}^j(\boldsymbol{\beta}_j) = \omega_{ii'}^j R_{ii'}^j(\boldsymbol{\beta}_j^*)$. As shown in §E.3, we have

$$\min\{1, \omega_{ii'}^j, \omega_{ii'}^k\}\mathbf{h}_{ii'}^{jk}(\boldsymbol{\beta}_{j\vee k}^*) \le \mathbf{h}_{ii'}^{jk}(\boldsymbol{\beta}_{j\vee k}) \le \max\{1, \omega_{ii'}^j, \omega_{ii'}^k\}\mathbf{h}_{ii'}^{jk}(\boldsymbol{\beta}_{j\vee k}^*), \tag{E.18}$$

where the inequality is taken elementwisely. We denote $b := \max_{i,i'\in[n];j\in[d]} r(s^*, n, d)\big\|(X_{ij} - X_{i'j})(\boldsymbol{X}_{i\backslash j} - \boldsymbol{X}_{i'\backslash j})\big\|_\infty$. Note that when $\|\boldsymbol{\beta}_j - \boldsymbol{\beta}_j^*\|_1 \le r(s^*, n, d)$ and $\|\boldsymbol{\beta}_k - \boldsymbol{\beta}_k^*\|_1 \le r(s^*, n, d)$, we have $\omega_{ii'}^j, \omega_{ii'}^k \in [\exp(-b), \exp(b)]$. Therefore by (E.18) and the definition of $V_{ii'i''}^{jk}(\boldsymbol{\beta}_{j\backslash k})$, we obtain the following elementwise inequality

$$\exp(-2b)\mathbf{V}_{ii'i''}^{jk}(\boldsymbol{\beta}_{j\backslash k}^*) \le \mathbf{V}_{ii'i''}^{jk}(\boldsymbol{\beta}_{j\backslash k}) \le \exp(2b)\mathbf{V}_{ii'i''}^{jk}(\boldsymbol{\beta}_{j\backslash k}^*),$$

which implies that

$$\big\|\widehat{\boldsymbol{\Sigma}}^{jk}(\boldsymbol{\beta}_{j\vee k}) - \widehat{\boldsymbol{\Sigma}}^{jk}(\boldsymbol{\beta}_{j\vee k}^*)\big\|_\infty \le \max\big\{1 - \exp(-2b), \exp(2b) - 1\big\}\big\|\widehat{\boldsymbol{\Sigma}}^{jk}(\boldsymbol{\beta}_{j\vee k}^*)\big\|_\infty. \tag{E.19}$$

As we show in §E.3, $b \le Cr(s^*, n, d)\log^2 d$ with high probability for some absolute constant $C > 0$. Since $\lim_{n\to\infty} r(s^*, n, d)\log^2 d = 0$, by (E.19) we have

$$\big\|\widehat{\boldsymbol{\Sigma}}^{jk}(\boldsymbol{\beta}_{j\vee k}) - \widehat{\boldsymbol{\Sigma}}^{jk}(\boldsymbol{\beta}_{j\vee k}^*)\big\|_\infty \lesssim b\big\|\widehat{\boldsymbol{\Sigma}}^{jk}(\boldsymbol{\beta}_{j\vee k}^*)\big\|_\infty\big\|_\infty \le b\big\|\widehat{\boldsymbol{\Sigma}}^{jk}(\boldsymbol{\beta}_{j\vee k}^*) - \boldsymbol{\Sigma}^{jk}\big\|_\infty + b\|\boldsymbol{\Sigma}^{jk}\|_\infty.$$

Note that we show $\|I_2\|_\infty = \big\|\widehat{\boldsymbol{\Sigma}}^{jk}(\boldsymbol{\beta}_{j\vee k}^*) - \boldsymbol{\Sigma}^{jk}\big\|_\infty = \mathcal{O}_{\mathbb{P}}\big(\sqrt{\log d/n}\big)$, which converges to zero asymptotically. Thus we conclude that

$$\big\|\widehat{\boldsymbol{\Sigma}}^{jk}(\boldsymbol{\beta}_{j\vee k}) - \widehat{\boldsymbol{\Sigma}}^{jk}(\boldsymbol{\beta}_{j\vee k}^*)\big\|_\infty = \mathcal{O}_{\mathbb{P}}\big(r(s^*, n, d)\log^2 d\big). \tag{E.20}$$

Combining (E.17) and (E.20), we have the following error bound for $\widehat{\boldsymbol{\Sigma}}^{jk}(\boldsymbol{\beta}_{j\vee k})$ :

$$\big\|\widehat{\boldsymbol{\Sigma}}^{jk}(\boldsymbol{\beta}_{j\vee k}) - \boldsymbol{\Sigma}^{jk}\big\|_\infty = \mathcal{O}_{\mathbb{P}}\Big(r(s^*, n, d)\log^2 d + \sqrt{\log d/n}\Big) \quad \text{for all } (j, k). \tag{E.21}$$

Finally, by the fact that $\max_{j\in[d]}\|\widehat{\boldsymbol{\beta}}_j - \boldsymbol{\beta}_j^*\|_1 \lesssim s^*\lambda$, Lemma D.2 follows from setting $r = Cs^*\lambda$. $\square$

### E.3 Proof of Lemma E.1

Now we turn to the last unproven result, namely Lemma E.1, which characterizes the perturbation of $\nabla^2 L_j(\boldsymbol{\beta}_j)$.

*Proof of Lemma E.1.* Note that $\nabla^2 L_j(\boldsymbol{\beta}_j)$ is a second-order $U$-statistic. Hence $\nabla^2 L_j(\boldsymbol{\beta}_j) - \mathbb{E}\big[\nabla^2 L_j(\boldsymbol{\beta}_j)\big]$ is a centered $U$-statistic. We denote its kernel as $\mathbf{T}_{ii'}(\boldsymbol{\beta}_j)$, then

$$\nabla^2 L_j(\boldsymbol{\beta}_j) - \mathbb{E}\big[\nabla^2 L_j(\boldsymbol{\beta}_j)\big] = \frac{2}{n(n-1)}\sum_{i<i'}\mathbf{T}_{ii'}(\boldsymbol{\beta}_j).$$

Note that $\big\|\mathbb{E}[\mathbf{T}_{ii'}(\boldsymbol{\beta}_j)]\big\|_\infty$ is bounded for all $\boldsymbol{\beta}_j \in \mathbb{R}^{d-1}$ because

$$\max_{\mathbf{u}\in\mathbb{R}^{d-1}}\big\|\mathbb{E}[T_{ii'}(\boldsymbol{\beta}_j)]\big\|_\infty \lesssim \max_{j\in[d]}\mathbb{E}|X_{ij} - X_{i'j}|^4 \lesssim \max_{j\in[d], i\in[n]}\mathbb{E}|X_{ij}|^4 \le \int_0^\infty c\exp(-t^{1/4})dt = 24c,$$

where $c = 2\exp(\kappa_m + \kappa_h/2)$. Here the last inequality follows from (4.1). Let $\nabla^2_{jk,j\ell} L_j(\boldsymbol{\beta}_j) = \partial^2 L_j(\boldsymbol{\beta}_j)/\big(\partial\beta_{jk}\partial\beta_{j\ell}\big)$ and let $\big[\boldsymbol{T}_{ii'}(\boldsymbol{\beta}_j)\big]_{k\ell}$ be the corresponding kernel function. That is, $\nabla^2_{jk,j\ell} L_j(\boldsymbol{\beta}_j) = \binom{n}{2}^{-1}\sum_{i<i'}\big[\boldsymbol{T}_{ii'}(\boldsymbol{\beta}_j)\big]_{k\ell}$. For $x > 0$ such that $x^4 > 24c$ and $k, \ell \neq j$, we have

$$\begin{aligned}
&\mathbb{P}\big\{\big|[\mathbf{T}_{ii'}(\boldsymbol{\beta}_j^*)]_{k\ell}\big| > 2x^4\big\} \leq \mathbb{P}\big[(X_{ij} - X_{i'j})^2(X_{ik} - X_{i'k})(X_{i\ell} - X_{i'\ell}) > x^4\big] \\
&\leq \mathbb{P}\big(|X_{ij} - X_{i'j}| > x\big) + \mathbb{P}\big(|X_{ik} - X_{i'k}| > x\big) + \mathbb{P}\big(|X_{i\ell} - X_{i'\ell}| > x\big).
\end{aligned} \tag{E.22}$$

As a direct implication of Assumption 4.1, we have $\mathbb{P}\big(|X_{ij} - X_{ij}| > x\big) \leq 2\exp(2\kappa_m + \kappa_k)\exp(-x)$ for all $j \in [d]$. Then we can bound the right-hand side of (E.22) by

$$\mathbb{P}\big\{\big|[\mathbf{T}_{ii'}(\boldsymbol{\beta}_j^*)]_{k\ell}\big| > 2x^4\big\} \leq 6\exp(2\kappa_m + \kappa_h)\exp(-x) \quad \text{when } x^4 > 48\exp(\kappa_m + \kappa_h/2).$$

Letting $C_T = \max\Big\{6\exp(2\kappa_m + \kappa_h), \exp\big\{[48\exp(\kappa_m + \kappa_h/2)]^{1/4}\big\}\Big\}$, it holds that

$$\mathbb{P}\big\{\big|[\mathbf{T}_{ii'}(\boldsymbol{\beta}_j^*)]_{k\ell}\big| > x\big\} \leq C_T\exp(-2^{-1/4}x^{1/4}) \quad \text{for all } x > 0. \tag{E.23}$$

Thus by a union bound, we conclude that there exists some generic constant $C$ such that $\|\mathbf{T}_{ii'}(\boldsymbol{\beta}_j^*)\|_\infty \leq C\log^4 d$ for all $j \in [d]$ and $i, i' \in [n]$ with probability at least $1 - (8d)^{-1}$. We define an event $\mathcal{E} := \big\{\|\mathbf{T}_{ii'}(\boldsymbol{\beta}_j^*)\|_\infty \leq C\log^4 d, \forall i, i' \in [n], j \in [d]\big\}$. By (E.23), it is easy to see that $\mathbf{T}_{ii'}(\boldsymbol{\beta}_j^*)$ is $\ell_2$-integrable. By Bernstein's inequality for $U$-statistics (Lemma C.1) with $b = C\log^4 d$ in (C.2), for some generic constants $C_1$ and $C_2$, we obtain that

$$\mathbb{P}\Big(\nabla^2 L_j(\boldsymbol{\beta}_j) - \mathbb{E}_1\big[\nabla^2 L_j(\boldsymbol{\beta}_j)\big] > t\big|\mathcal{E}\Big) \leq 4\exp\big[-nt^2/(C_1 + C_2\log^4 \cdot t)\big], \quad \forall j \in [d]. \tag{E.24}$$

Here we use $\mathbb{E}_1\big[\nabla^2 L_j(\boldsymbol{\beta}_j)\big]$ to denote $\mathbb{E}\big[\nabla^2 L_j(\boldsymbol{\beta}_j)\big|\mathcal{E}\big]$. Thus under Assumption 4.3 we obtain that, conditioning on event $\mathcal{E}$,

$$\big\|\nabla^2 L_j(\boldsymbol{\beta}_j) - \mathbb{E}_1\big[\nabla^2 L_j(\boldsymbol{\beta}_j)\big]\big\|_\infty \leq C\sqrt{\log d/n}, \quad \forall j \in [d] \tag{E.25}$$

with probability at least $1 - (8d)^{-1}$. Moreover, by (E.23) we obtain that

$$\mathbb{E}\big\{[\mathbf{T}_{ii'}(\boldsymbol{\beta}_j^*)]_{k\ell}\big|\mathcal{E}\big\} - \mathbb{E}\big\{[\mathbf{T}_{ii'}(\boldsymbol{\beta}_j^*)]_{k\ell}\big\} \leq \int_{C\log^4 d}^{\infty}\mathbb{P}\big\{\big|[\mathbf{T}_{ii'}(\boldsymbol{\beta}_j^*)]_{k\ell}\big| > x\big\} \leq c_1\log^3 d \cdot \exp(-c_2\log d)$$

for some absolute constant $c_1$ and $c_2$. Therefore we have

$$\big\|\mathbb{E}_1\big[\nabla^2 L_j(\boldsymbol{\beta}_j)\big] - \mathbb{E}\big[\nabla^2 L_j(\boldsymbol{\beta}_j)\big]\big\|_\infty \lesssim \log^3 d \cdot \exp(-c_2\log d) \lesssim \sqrt{\log d/n}. \tag{E.26}$$

Combining (E.25) and (E.26) we show that, with probability at least $1 - (4d)^{-1}$, $\big\|\nabla^2 L_j(\boldsymbol{\beta}_j^*) - \mathbb{E}[\nabla^2 L_j(\boldsymbol{\beta}_j^*)]\big\|_\infty \leq C_h\sqrt{\log d/n}$ for all $j \in [d]$.

For the second argument (E.6), let $\boldsymbol{\Delta} = \boldsymbol{\beta}_j - \boldsymbol{\beta}_j^*$ where $\boldsymbol{\beta}_j \in \mathbb{R}^{d-1}$ lies in the $\ell_1$-ball centered at $\boldsymbol{\beta}_j^*$ with radius $r_1(s^*, n, d)$, that is, $\big\|\boldsymbol{\beta}_j - \boldsymbol{\beta}_j^*\big\|_1 \leq r_1(s^*, n, d)$. By the independence between $\boldsymbol{X}_i$ and $\boldsymbol{X}_{i'}$, Assumption 4.1 implies that

$$\max\Big\{\log\mathbb{E}\big[\exp(X_{ij} - X_{i'j})\big], \log\mathbb{E}\big[\exp(X_{i'j} - X_{ij})\big]\Big\} \leq 2\kappa_m + \kappa_h,$$

which further implies that for any $x > 0$

$$\mathbb{P}\Big(\big|(X_{ij} - X_{i'j})\big| > x\Big) \le 2\exp(2\kappa_m + \kappa_h)\exp(-x), \quad \forall j \in [d].$$

Hence for any $x > 0$ and $j, k \in [d]$, a union bound implies that

$$\begin{aligned} \mathbb{P}\big[\big|(X_{ij} - X_{i'j})(X_{ik} - X_{i'k})\big| > x^2\big] &\le \mathbb{P}\big[\big|(X_{ij} - X_{i'j})\big| > x\big] + \mathbb{P}\big[\big|(X_{ik} - X_{i'k})\big| > x\big] \\ &\le 4\exp(2\kappa_m + \kappa_h)\exp(-x). \end{aligned} \tag{E.27}$$

Taking a union bound over $1 \le j < k \le d$ and $1 \le i < i' \le n$ we obtain that

$$\mathbb{P}\Big[\max_{i,i' \in [n]; j \in [d]} \big\|(X_{ij} - X_{i'j})(\boldsymbol{X}_{i\backslash j} - \boldsymbol{X}_{i'\backslash j})\big\|_\infty > x^2\Big] \lesssim n^2 d^2 \exp(-x).$$

If we denote $b := \max_{i,i' \in [n]; j \in [d]} r_1(s^*, n, d)\big\|(X_{ij} - X_{i'j})(\boldsymbol{X}_{i\backslash j} - \boldsymbol{X}_{i'\backslash j})\big\|_\infty$, then we obtain that $b \le C r_1(s^*, n, d)\log^2 d$ with probability at least $1 - (4d)^{-1}$ for some constant $C > 0$. Denoting $\omega_{ii'} := \exp\big\{-(X_{ij} - X_{i'j})\boldsymbol{\Delta}^T(\boldsymbol{X}_{i\backslash j} - \boldsymbol{X}_{i'\backslash j})\big\}$, by definition,

$$R^j_{ii'}(\boldsymbol{\beta}_j) = \exp\big\{-(X_{ij} - X_{i'j})(\boldsymbol{\Delta} + \boldsymbol{\beta}^*_j)^T(\boldsymbol{X}_{i\backslash j} - \boldsymbol{X}_{i'\backslash j})\big\} = \omega_{ii'} R^j_{ii'}(\boldsymbol{\beta}^*_j).$$

Thus we can write $\nabla^2 L_j(\boldsymbol{\beta}_j)$ as:

$$\nabla^2 L_j(\boldsymbol{\beta}_j) = \frac{2}{n(n-1)} \sum_{i<i'} \frac{R^j_{ii'}(\boldsymbol{\beta}^*)(X_{ij} - X_{i'j})^2(\boldsymbol{X}_{i\backslash j} - \boldsymbol{X}_{i'\backslash j})^{\otimes 2}}{\big(1 + R^j_{ii'}(\boldsymbol{\beta}^*)\big)^2} \frac{\omega_{ii'}\big(1 + R^j_{ii'}(\boldsymbol{\beta}^*)\big)^2}{\big(1 + \omega_{ii'} R^j_{ii'}(\boldsymbol{\beta}^*)\big)^2}. \tag{E.28}$$

If $\omega_{ii'} \ge 1$, then $(\omega_{ii'})^{-2} \le \big(1 + R^j_{ii'}(\boldsymbol{\beta}^*)\big)^2 / \big(1 + \omega_{ii'} R^j_{ii'}(\boldsymbol{\beta}^*)\big)^2 \le 1$; otherwise we have $1 \le \big(1 + R^j_{ii'}(\boldsymbol{\beta})\big)^2 / \big(1 + \omega_{ii'} R^j_{ii'}(\boldsymbol{\beta}^*)\big)^2 \le (\omega_{ii'})^{-2}$. This observation implies

$$\min\big\{\omega_{ii'}, 1/\omega_{ii'}\big\} \le \frac{\omega_{ii'}\big(1 + R^j_{ii'}(\boldsymbol{\beta})\big)^2}{\big(1 + \omega_{ii'} R^j_{ii'}(\boldsymbol{\beta}^*)\big)^2} \le \max\big\{\omega_{ii'}, 1/\omega_{ii'}\big\}. \tag{E.29}$$

By the definition of $\omega_{ii'}$, Hölder's inequality implies that $\big|(X_{ij} - X_{i'j})\boldsymbol{\Delta}^T(\boldsymbol{X}_{i\backslash j} - \boldsymbol{X}_{i'\backslash j})\big| \le b$, thus we have

$$\exp(-b) \le \min\big\{\omega_{ii'} 1/\omega_{ii'}\big\} \le \max\big\{\omega_{ii'}, 1/\omega_{ii'}\big\} \le \exp(b). \tag{E.30}$$

Combining (E.28),(E.29) and (E.30) we obtain

$$\exp(-b)\nabla^2 L_j(\boldsymbol{\beta}^*_j) \le \nabla^2 L_j(\boldsymbol{\beta}_j) \le \exp(b)\nabla^2 L_j(\boldsymbol{\beta}^*_j). \tag{E.31}$$

Then by (E.31), since $\lim_{n\to\infty} r_1(s^*, n, d)\log^2 d = 0$, we have

$$\big\|\nabla^2 L_j(\boldsymbol{\beta}_j) - \nabla^2 L_j(\boldsymbol{\beta}^*_j)\big\|_\infty \le \max\big\{1 - \exp(-b), \exp(b) - 1\big\}\big\|\nabla^2 L_j(\boldsymbol{\beta}^*_j)\big\|_\infty \lesssim b\big\|\nabla^2 L_j(\boldsymbol{\beta}^*_j)\big\|_\infty.$$

Notice that under Assumption 4.1, as shown in §D.1, we can assume that $\big\|\mathbb{E}\big[\nabla^2 L_j(\boldsymbol{\beta}^*_j)\big]\big\|_\infty \le D$ where $D$ appears in (D.1). By triangle inequality,

$$\big\|\nabla^2 L_j(\boldsymbol{\beta}^*_j)\big\|_\infty \le \big\|\nabla^2 L_j(\boldsymbol{\beta}^*_j) - \mathbb{E}\big[\nabla^2 L_j(\boldsymbol{\beta}^*_j)\big]\big\|_\infty + \big\|\mathbb{E}\big[\nabla^2 L_j(\boldsymbol{\beta}^*_j)\big]\big\|_\infty \le D + C_h\sqrt{\log d/n} \le 2D$$

with probability at least $1-(4d)^{-1}$, where the last inequality follows from the fact that $\big(\log^9 d/n\big)^{1/2}$ tends to zero as $n$ goes to infinity. Then we obtain that

$$\big\|\nabla^2 L_j(\boldsymbol{\beta}_j) - \nabla^2 L_j(\boldsymbol{\beta}_j^*)\big\|_\infty \le C_r r_1(s^*, n, d)\log^2 d$$

holds for some absolute constant $C_r > 0$ and uniformly for all $j \in [d]$ and $\boldsymbol{\beta}_j \in \mathbb{B}_j(r_1)$ with probability at least $1-(2d)^{-1}$.

Finally, for the last argument (E.7), for any $\mathbf{v} \in \mathbb{R}^{d-1}$, by (E.31) we have

$$\exp(-b)\mathbf{v}^T\nabla^2 L_j(\boldsymbol{\beta}_j^*)\mathbf{v} \le \mathbf{v}^T\nabla^2 L_j(\boldsymbol{\beta}_j)\mathbf{v} \le \exp(b)\mathbf{v}^T\nabla^2 L_j(\boldsymbol{\beta}_j^*)\mathbf{v}.$$

Thus we have

$$\big|\mathbf{v}^T\big[\nabla^2 L_j(\boldsymbol{\beta}_j) - \nabla^2 L_j(\boldsymbol{\beta}_j^*)\big]\mathbf{v}\big| \lesssim b\big|\mathbf{v}^T\nabla^2 L_j(\boldsymbol{\beta}_j^*)\mathbf{v}\big| \le b\|\mathbf{v}\|_1^2\big\|\nabla^2 L_j(\boldsymbol{\beta}_j^*)\big\|_\infty,$$

which implies (E.7). □